\documentclass[10pt,journal,compsoc]{IEEEtran}
\usepackage{comment}
\usepackage{amsmath,amssymb} % define this before the line numbering.
\usepackage{pifont}% http://ctan.org/pkg/pifont
\newcommand{\cmark}{\ding{51}}%
\newcommand{\xmark}{\ding{55}}%
\usepackage{color}
\definecolor{mycitecolor}{rgb}{0, 0.4, 0.7}
\usepackage[breaklinks=true,bookmarks=true,citecolor=mycitecolor,colorlinks,pagebackref=false]{hyperref}
\usepackage{booktabs}
\usepackage[numbers,sort]{natbib}
\usepackage{multirow}
\usepackage{multicol}
\usepackage{graphicx,nicefrac}
\usepackage{amsmath}
\usepackage{algorithmic}
\usepackage{array}
\usepackage{bm}
\usepackage{subcaption}
\usepackage{overpic}
\usepackage{makecell}
\usepackage[switch]{lineno}

\newcommand{\shorteq}{%
  \settowidth{\@tempdima}{-}% Width of hyphen
  \resizebox{\@tempdima}{\height}{=}%
}

\begin{document}
% \begin{comment}

% \end{comment}
% \linenumbers
\title{
Delving into the Devils of Bird's-eye-view Perception: A Review, Evaluation and Recipe
}

\author{Hongyang Li$^{*\dagger}$,~
        Chonghao Sima$^*$,~
        Jifeng Dai$^*$,~
        Wenhai Wang$^*$,~
        Lewei Lu$^{*\dagger}$,~
        Huijie Wang$^*$,~ 
        Jia Zeng$^*$,~\\
        Zhiqi Li$^*$,~
        Jiazhi Yang$^{*}$,~
        Hanming Deng$^{*}$,~
        Hao Tian$^{*}$,~
        Enze Xie$^*$,~
        Jiangwei Xie,~
        Li Chen,~
        Tianyu Li,~\\
        Yang Li,~ 
        Yulu Gao,~ 
        Xiaosong Jia,~
        % Bohan Yu,~
        Si Liu,~
        Jianping Shi,~
        % Ping Luo,~
        % Junchi Yan,~
        Dahua Lin~
        and Yu~Qiao
\IEEEcompsocitemizethanks{

\IEEEcompsocthanksitem $^*$indicates equal contribution. 
% $^{\dagger}$ denotes project leader for each track. 
For detailed role of each author, please refer to Author Contributions section. 
\IEEEcompsocthanksitem This is a joint work between Shanghai AI Lab and SenseTime.
$^{\dagger}$Primary contact to: Hongyang Li \text{\texttt{hy@opendrivelab.com}} or Lewei Lu
\text{\texttt{luotto@sensetime.com}}. 
% Author Contribution is listed in Appendix.

\vspace{.1cm}

\IEEEcompsocthanksitem 
H. Li, C. Sima, J. Dai, 
H. Wang, J. Zeng, Z. Li, J. Yang, L. Chen,  T. Li, Y. Li, X. Jia, D. Lin and Y. Qiao
are with Shanghai AI Lab.
L. Lu, H. Deng, H. Tian, J. Xie and J. Shi are with SenseTime.
C. Sima and L. Chen are also with The University of Hong Kong; W. Wang is with the Chinese University of Hong Kong. E. Xie 
 is with Huawei Inc. Y. Gao and S. Liu is with Beihang University.

}
\thanks{
Manuscript received September, 2022; first revision September, 2023.}
}

\markboth{Submitted to IEEE Transactions on Pattern Analysis and Machine Intelligence
% , August 2022
}%
{
% Li \MakeLowercase{\textit{et al.}}: 
% Towards Unified Perception and Decision Paradigm for 
%  End-to-end Autonomous Driving System: A Survey
}

\IEEEtitleabstractindextext{%
\begin{abstract}
Learning powerful representations in bird's-eye-view (BEV) for perception tasks is trending and drawing extensive attention both from industry and academia. Conventional approaches for most autonomous driving algorithms perform detection, segmentation, tracking, etc., in a front or perspective view. As sensor configurations get more complex, integrating multi-source information from different sensors and representing features in a unified view come of vital importance.
BEV perception inherits several advantages, as representing surrounding scenes in BEV is 
intuitive and fusion-friendly; and representing objects in BEV is most desirable for subsequent modules as in planning and/or control.
The core problems 
% to resolve 
for BEV perception %for camera sensor 
lie in (a) how to reconstruct the lost 3D information via view transformation from perspective view to BEV;   % core in BEV camera
(b) how to acquire ground truth annotations in BEV grid;  % dataset?
(c) how to formulate the pipeline to incorporate features from different sources and views;  % BEV LiDAR and fusion?
and (d) how to adapt and generalize algorithms as sensor configurations vary across different scenarios. % what's next?
In this survey, we review the most recent {}{works} on BEV perception and provide an in-depth analysis of different solutions.
Moreover, several systematic designs of BEV approach from the industry are depicted as well. 
Furthermore, we introduce a full suite of practical guidebook to improve the performance of BEV perception tasks, including camera, LiDAR and fusion inputs. 
At last, we point out the future research directions in this area. We hope this report will shed some light on the community and encourage more research 
effort on BEV perception.
We keep an active repository to collect the most recent work and provide a toolbox for bag of tricks at \texttt{\url{https://github.com/OpenDriveLab/Birds-eye-view-Perception}}.

\end{abstract}

}

\maketitle

\IEEEdisplaynontitleabstractindextext
\IEEEpeerreviewmaketitle

\IEEEraisesectionheading{
\section{Introduction}\label{sec:introduction}
}
\IEEEPARstart{P}{erception} recognition task in autonomous driving is essentially a 3D geometry reconstruction to the physical world.
As the diversity and number of sensors become more and more complicated in equipping the self-driving vehicle (SDV), representing features from different views in a unified perspective comes to vital importance. 
The well-known bird's-eye-view (BEV) is a natural and straightforward candidate view to serve as a unified representation. 
Compared to its front-view or perspective view counterpart, which is broadly investigated~\cite{ren2015faster,he2017mask} in 2D vision domains, BEV representation bears several inherent merits. 
First, it has no occlusion or scale problem as commonly existent in 2D tasks.
Recognizing vehicles with occlusion or cross-traffic could be better resolved.
Moreover, representing objects or road elements in such form would favorably make it convenient for subsequent modules (e.g. planning, control) to develop and deploy.

In this survey, we term \textbf{BEV Perception} to indicate all vision algorithms in sense of the BEV view representation for autonomous driving.
Note that we do not intend to exaggerate BEV perception as a new research concept; instead, how to formulate new pipeline or framework under BEV view for better feature fusion from multiple sensor inputs, deserves more attention from the community.

\subsection{Big Picture at a Glance}

\begin{figure*}
\centering
\includegraphics[width=0.9\textwidth]{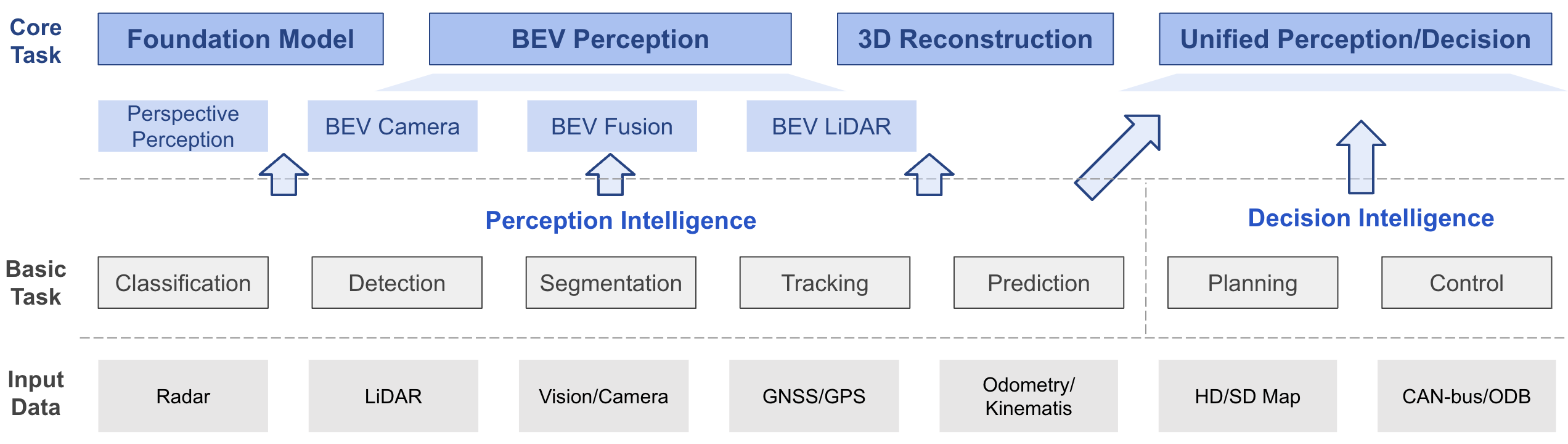}
\caption{The general picture of BEV perception at a glance, where consists of three sub-parts based on the input modality. BEV perception is a general task built on top of a series of fundamental tasks. For better completeness of the whole perception algorithms in autonomous driving, we list other topics (e.g., Foundation Model) as well.}
\label{fig:roadmap}
\vspace{-0.4cm}
\end{figure*}

Based on the input data, we divide BEV perception research into three parts mainly - BEV camera, BEV LiDAR and BEV fusion.
Fig.~\ref{fig:roadmap} depicts the general picture of BEV perception family. Specifically, \textbf{BEV camera} indicates vision-only or vision-centric algorithms for 3D object detection or segmentation from multiple surrounding cameras;
\textbf{BEV LiDAR} describes detection or segmentation task from point cloud input; 
\textbf{BEV fusion} depicts the fusion mechanisms from multiple sensor inputs, such as camera, LiDAR, GNSS, odometry, HD-Map, CAN-bus, etc.

As described in Fig.~\ref{fig:roadmap}, we group and classify fundamental perception algorithms (classification, detection, segmentation, tracking, etc.) with autonomous driving tasks into three levels, where the concept of BEV perception lies in the middle. Based on different combinations from the sensor input layer, fundamental task, and product scenario, a certain BEV perception algorithm could indicate accordingly. For instance, M$^2$BEV~\cite{xie2022m2bev} and BEVFormer~\cite{li2022bevformer} belonged to BEV camera track from multiple cameras to perform multiple tasks including 3D object detection and BEV map segmentation. 
BEVFusion~\cite{liu2022bevfusion} devised a fusion strategy in BEV space to perform 3D detection and tracking simultaneously from both camera and LiDAR inputs.
Tesla~\cite{tesla_ai_day} released its systematic pipeline to detect objects and lane lines in vector space (BEV) for L2 highway navigation and smart summon. 
In this report, we aim to summarize a general pipeline and key insights for recent advanced BEV perception research, apart from various input combinations and tasks.

\subsection{Motivation to BEV Perception Research}

% Three aspects are to be examined when it comes to the motivation of BEV perception research.
When it comes to the motivation of BEV perception research, three important aspects need to be examined.

\smallskip
\textbf{Significance.} \textit{Will BEV perception generate a real and meaningful impact on academia and/or society?} 
It is well-known that a tremendous performance gap exists between camera-only and LiDAR solutions. 
For example, as of submission in August 2022, the gap between first-ranking camera-only and LiDAR methods exceed 20\% on nuScenes dataset~\cite{caesar2020nuscenes} and 30\% on Waymo benchmark~\cite{sun2020scalability}.This naturally prompts us to investigate whether the camera-only solutions can beat or be on par with the LiDAR approaches.

From the academic perspective, the very essence of designing a camera-based pipeline such that it outperforms LiDAR, is better to understand the view transformation process from a 2D, appearance input to a 3D, geometry output. How to transfer camera features into geometry representations as does in point cloud leaves a meaningful impact for the academic society.
On the industrial consideration, the cost of a suite of LiDAR equipment into SDV is expensive; OEMs (Original Equipment Manufacturer, e.g., Ford, BMW, etc.) prefer a cheap as well as accurate deployment for software algorithms. Improving camera-only algorithms to LiDAR's naturally fall into this goal since the cost of a camera is usually 10x times cheaper than LiDAR. Moreover, camera based pipeline can recognize long-range distance objects and color-based road elements (e.g., traffic lights), both of which LiDAR approaches are incapable of.

Although there are several different solutions for camera and LiDAR based perception, BEV representation is one of the best candidates for LiDAR based methods in terms of superior performance and industry-friendly deployment. 
Moreover, recent trends show that BEV representation also has huge progress for multi-camera inputs. 
Since camera and LiDAR data can be projected to BEV space, another potential for BEV is that we can easily fuse the feature from different modalities under a unified representation.
%
% In summary, BEV feature representation is a promising solution for the multi-sensor multi-task perception and we believe future research on BEV perception will result in a great impact on both academia and industry.

\smallskip

\textbf{Space.} \textit{Are there open problems or caveats in BEV perception that require substantial innovation?} 
The gist behind BEV perception is to learn a robust and generalizable feature representation from both camera and LiDAR inputs. This is easy in LiDAR branch as the input (point cloud) bears such a 3D property. This is non-trivial otherwise in the camera branch, as learning 3D spatial information from monocular or multi-view settings is difficult. While we see there are some attempt to learn better 2D-3D correspondence by pose estimation~\cite{chen2022_epro-pnp} or temporal motion~\cite{wang2022dfm}, the core issue behind BEV perception requires a substantial innovation of depth estimation from raw sensor inputs, esp. for the camera branch. 

Another key problem is how to fuse features in early or middle stage of the pipeline. Most sensor fusion algorithms treat the problem as a simple object-level fusion or naive feature concatenation along the blob channel. This might explain why some fusion algorithms behave inferior to LiDAR-only solutions, due to the misalignment or inaccurate depth prediction between camera and LiDAR. How to align and integrate features from multi-modality input plays a vital role and thereby leaves extensive space to innovate.

\smallskip

\textbf{Readiness.} \textit{Are key conditions (e.g. dataset, benchmark) ready to conduct BEV perception research?}  
The short answer is yes. As BEV perception requires both camera and LiDAR, high-quality annotation and accurate alignment between 2D and 3D objects are two key evaluations for such benchmarks. While KITTI~\cite{geiger2012we} is comprehensive and draws much attention in early autonomous driving research, large-scale and diverse benchmarks such as Waymo~\cite{sun2020scalability}, nuScenes~\cite{caesar2020nuscenes}, Argoverse~\cite{Argoverse2} provide a solid playground for validating BEV perception ideas. These newly proposed benchmarks usually feature high-quality labels; the scenario diversity and data amount also scale up to great extent. Furthermore, open challenges~\cite{waymo2020cvprworkshop} on these leaderboards give a fair setting on the held-out test data, where all state-of-the-arts can be compared in an open and prompt sense.

As for the algorithm readiness, recent years have witnessed a great blossom for general vision, where Transformer~\cite{vaswani2017_transformer}, ViT~\cite{dosovitskiy2020_vit, han2022survey}, Masked Auto-encoders (MAE)~\cite{he2022_mae} and CLIP \cite{radford2021_clip}, etc., achieve impressive gain over conventional methods. We believe these work would benefit and inspire greatness for of BEV perception research.

\smallskip

Based on the discussions of the three aspects above, we conclude that BEV perception research has great potential impact, and deserves massive attention from both academia and industry to put great effort for a long period.
Compared to the recent surveys on 3D object detection~\cite{arnold2019survey, liang2021survey, qian20223d, ma2022vision, mao20223d}, our survey not only summarizes recent BEV perception algorithms at a more high level and formulates them into a general pipeline, but also provides useful recipe under such context, including solid data augmentation in both camera-based and LiDAR-based settings, efficient BEV encoder design, perception heads and loss function family, useful test-time augmentation (TTA) and ensemble policy, and so on.
We hope this survey can be a good starting point for new beginners and an insightful discussion for current researchers in this community.
% \smch{cite previous 3d object detection surveys and what's the difference}

\subsection{Contributions}
The main contributions in this survey are three-fold:
\begin{enumerate}
    \item We review the whole picture of BEV perception research in recent years, including high-level philosophy and in-depth detailed discussion.
    % on the 
    % significance, space, readiness, 
    % challenges and future work.
    \item We elaborate on a comprehensive analysis of BEV perception literature. The core issues such as depth estimation, view transformation, sensor fusion, domain adaptation, etc., are covered. Several important industrial system-level designs for BEV perception are introduced and discussed as well.  
    \item We provide a practical guidebook, besides theoretical contribution, for improving the performance in various BEV perception tasks. Such a release could facilitate the community to achieve better performance in a grab-and-go recipe sense.
\end{enumerate}

% \noindent The layout of this paper in the rest sections is organized as follows.
% %
% Sec.~\ref{related work} provide necessary background knowledge in this domain.
% % including a review of different 3D perception task settings and conventional approaches in Sec~\ref{sec: task_def_sol}, the common datasets and related metrics in Sec.~\ref{dataset} and a simple formulation of 3D geometry in Sec.~\ref{3dvision}.
% %
% % We then recap the basic knowledge for 3D vision in Sec.~\ref{3dvision}.
% %
% Then, the detailed literature review and discussions for BEV perception are provided in Sec.~\ref{bev_method}, where the last part in this section describes some milestone industrial designs for BEV algorithms on a system level. 
% %
% % Sec.~\ref{dataset} depicts the common datasets and evaluation metrics in this area.
% %
% Furthermore, we provide a practical guidebook in Sec.~\ref{recipe}, from the experience we participate in the Waymo Open Challenge 2022.
% %
% At last, in Sec.~\ref{challenge}, we list the current challenges in the summary and future research directions.

\section{Background in 3D perception }\label{related work}
% HY: 第二部分要至少写一页或以上
% written by Enze
% \smch{should be previous work in a nutshell, enough background for a common solution in this domain; aka background knowledge}
In this section, we present the essential background knowledge in 3D perception.
% including different task settings with ordinary solutions in Sec.~\ref{sec: task_def_sol}, common datasets and corresponding metrics in Sec.~\ref{dataset}, and preliminary knowledge in 3D geometry in Sec.~\ref{3dvision}. 
%
In Sec.~\ref{sec: task_def_sol}, we review conventional approaches to perform perception tasks, including monocular camera based 3D object detection, LiDAR based 3D object detection and segmentation, and sensor fusion strategies.
In Sec.~\ref{dataset}, we introduce predominant datasets in 3D perception, 
% especially in autonomous driving domain, 
such as KITTI dataset~\cite{geiger2012we}, nuScenes dataset~\cite{caesar2020nuscenes} and Waymo Open Dataset~\cite{sun2020scalability}.
%
% In Sec.~\ref{3dvision} in Appendix, we explain a set of mathematical formula on 3D-2D projection, 
% which is the base of view transformation in Sec.~\ref{bev-camera}
% , which is the core of BEV perception.

\subsection{Task Definition and Related Work}\label{sec: task_def_sol}

% In this section, we describe 3D perception tasks and their conventional solutions, including monocular camera-based 3D object detection, LiDAR-based 3D object detection and segmentation and sensor-fusion-based 3D object detection.

% \subsubsection{Monocular Camera-based Object Detection}
\noindent\textbf{Monocular Camera-based Object Detection.}
% \smch{should be re-written in an uniform way, say what's done in perspective view, consider structure in \cite{ma2022vision}}
% written by Enze
Monocular camera-based methods take an RGB image as input and attempt to predict the 3D location and category of each object. The main challenge of monocular 3D detection is that the RGB image lacks depth information, so these kinds of methods need to predict the depth. Due to estimating depth from a single image being an ill-posed problem, typically monocular camear-based methods have inferior performance than LiDAR-based methods.

\smallskip
% \subsubsection{LiDAR Detection and Segmentation}
\noindent\textbf{LiDAR Detection and Segmentation.}
% \smch{summerizes as point base, voxel base, range view base, consider the structure from \cite{mao20223d}}
% written by huijie
LiDAR describes surrounding environments with a set of points in the 3D space, which capture geometry information of objects. Despite the lack of color and texture information and the limited perception range, LiDAR-based methods outperform camera-based methods by a large margin 
% benefit from 
due to the depth prior.

% \subsubsection{Sensor Fusion}
% written by huijie / yulu / tianhao
\smallskip
\noindent\textbf{Sensor Fusion.}
% Perception is a vital task for autonomous driving, since the following tasks, namely prediction and planning, depend on the correct perception of the surrounding world. Multiple sensors are utilized to perform the perception task, such as camera, LiDAR, and Radar. 
Modern autonomous vehicles are equipped with different sensors such as cameras, LiDAR and Radar. Each sensor has advantages and disadvantages.
Camera data contains dense color and texture information but fails to capture depth information. 
LiDAR provides accurate depth and structure information but suffers from limited range and sparsity. 
Radar is more sparse than LiDAR, but has a longer sensing range and can capture information from moving objects. 
Ideally, sensor fusion would push the upper bound performance of the perception system, and yet how to fuse data from different modalities remains a challenging problem.

% % \clearpage
\begin{table*}[tb]
\centering
\renewcommand\arraystretch{1.2}
  \setlength\tabcolsep{0.1cm}
    \caption{ 
    BEV Perception datasets at a glance. \textbf{Scenes} indicates clips of dataset, and the length of scene is varying for diverse datasets. Under \textbf{Region}, ``AS'' for Asia, ``EU'' for Europe, ``NA'' for North America, ``Sim'' for simulation data. Under \textbf{Sensor Data}, ``Scans'' for point cloud. Under \textbf{Annotation}, \textbf{Frames} implies the number of 3D bbox/ 3D lane annotation frames, \textbf{3D bbox/ 3D lane} represents the number of 3D bbox/ 3D lane annotation instances, \textbf{3D seg.} means the number of segmentation annotation frames of point cloud. 
    ``\# Subm.'' denotes the popularity of a particular dataset by the number of submissions on Kaggle. 
    % \textbf{Methods} means a rough number of 
%   literature reporting performance on this benchmark and they are .
  $\dagger$ refers that the statistic is unavailable; $-$ means that the field is non-existent.
    }
\begin{center}
    \begin{tabular}{l|c|c|c|c|c|c|c|c|c|c|c|c|c}
    \toprule
    \multirow{2}{*}{\textbf{Dataset}}  & 
    \multirow{2}{*}{\textbf{Year}} &
    \multirow{2}{*}{\textbf{Region}} & 
    \multicolumn{4}{c|}{\textbf{Sensor Data}} &
    \multicolumn{4}{c|}{\textbf{Annotation}} &
    \multirow{2}{*}{\textbf{HD-Map}} &
    \multirow{2}{*}{\textbf{Other Data}} & \multirow{2}{*}{\textbf{\# Subm.}} \\
\cline{4-11}    &    &  & Scenes &   Hours  & Scans & Images & Frames & 3D bbox& 3D lane & 3D seg. & &   \\
    \midrule
    KITTI~\cite{geiger2012we} & 2012  & EU & 22 & 1.5  & 15k   & 15k    & 15k   & 80k    & -     & -     & \xmark & -   & 380+\\
    Waymo~\cite{sun2020scalability} & 2019  & NA & 1150 & 6.4 & 230k  & 12M    & 230k  & 12M   & -     & 50k      & \xmark & - & 200+\\
    nuScenes~\cite{caesar2020nuscenes} & 2019  & NA/AS & 1000   & 5.5 & 390k &  1.4M & 40k  & 1.4M  & -    & 40k      & \cmark & CAN-bus  & 350+\\
    Argo v1~\cite{Argoverse} & 2019  & NA  & 113 & 0.6 & 22k  & 490k  &   22k    & 993k  & -    & -     & \cmark & - & 100+\\
    Argo v2~\cite{Argoverse2} & 2022  & NA & 1000 & 4 & 150k   & 2.7M  & 150k   & $\dagger$     & -    & -         &  \cmark & -  & 10+\\
    \midrule
    ApolloScape~\cite{wang2019apolloscape} & 2018  &  AS  & 103 & 2.5 &  29k    & 144k  & 144k  & 70k   & $\dagger$    & -     &  \cmark & - & 200+ \\
    OpenLane~\cite{chen2022persformer} & 2022  & NA     & 1000 & 6.4 & -  & 200k &   200k& -   & 880k     & -      &   \xmark & -  & - \\
    ONCE-3DLanes~\cite{yan2022once} & 2022  & AS     & $\dagger$  & $\dagger$  & -  & 211k &   211k& -   & $\dagger$   & -      &   \xmark & -  & -\\
    \midrule
    Lyft L5~\cite{houston2020one} & 2019  & AF & 366 & 2.5 & 46k   & 240k   & 46k   & 1.3M  & -    & -         &   \xmark & - & 500+\\
    A* 3D~\cite{pham20203d} & 2019  & AS  & $\dagger$ & 55 &    39k   & 39k  &   39k    & 230k    & -     & -          &  \xmark & -  & -\\
    H3D~\cite{Patil2019TheHD} & 2019  & NA & 160 & 0.8  &   27k    & 83k   & 27k   &  1.1M   & -     & -          &  \xmark & - & -\\
    SemanticKITTI~\cite{behley2019iccv} & 2019  & EU & 22 & 1.2  & 43k & -     & -     & -     & -     & 43k &   \xmark & - & 30+\\
    A2D2~\cite{geyer2020a2d2} &   2020   &   EU    & $\dagger$ & $\dagger$ &    12.5k    & 41k   &   12.5k     &    43k     & -   &    41k       &   \xmark & - & -\\
    Cityscapes 3D~\cite{gahlert2020cityscapes} & 2020  & - & $\dagger$ & 2.5 & -    & 5k    & 5k    & 40k       & -     & -         &   \xmark & IMU/GPS & 400+\\
    PandaSet~\cite{xiao2021pandaset} & 2020  & NA & 179 & $\dagger$& 16k & 41k & 14k  &   $\dagger$   & -    & 60k     &  \xmark & -  & -\\
    KITTI-360~\cite{Liao2021ARXIV} & 2020  &  EU     & 11 & $\dagger$  & 80k   & 320k  & 80k   & 68k   & -    & 80k       &   \xmark & -& 30+\\
    Cirrus~\cite{wang2021cirrus} & 2020  & - & 12  & $\dagger$ & 6285  & 6285  & 6285  & $\dagger$  & -     & -     & \xmark & -  & -\\
    ONCE~\cite{mao2021one} & 2021  & AS &  $\dagger$  & 144 & 1M    & 7M & 15k   & 417k  & -     & -          &  \xmark & -   & -\\
    AIODrive~\cite{Weng2020_AIODrive} & 2021  & Sim     & 100  & 2.8& 100k  & 1M & 100k  & 26M   & -     & 100k      &  \cmark & -   & -\\
    DeepAccident~\cite{wangdeepaccident} & 2022  & Sim     & 464 & $\dagger$  & 131k  & 786k & 131k  & 1.8M   & -     & 131k      &  \cmark & - &  - \\

    \bottomrule
    \end{tabular}
\end{center}
\label{tab:dataset_comparison}
\vspace{-0.4cm}
\end{table*}

\subsection{Datasets and Metrics}\label{dataset}
% write by enze
We introduce some popular autonomous driving datasets and the common evaluation metrics.
Tab.~\ref{tab:dataset_comparison} summarizes the main statistics of prevailing benchmarks for BEV perception.
Normally, a dataset consists of various scenes, each of which is of different length in different datasets.
The total duration ranges from tens of minutes to hundreds of hours.
For BEV perception tasks, 3D bounding box annotation and 3D segmentation annotation are essential and HD-Map configuration has become a mainstream trend.
Most of them can be adopted in different tasks. 
A consensus reached that sensors with multiple modalities and various annotations are required.
%
% For more exploration, 
More types of data~\cite{caesar2020nuscenes, Argoverse,Argoverse2,wang2019apolloscape, wangdeepaccident, gahlert2020cityscapes} are released such as IMU/GPS and CAN-bus.
Similar to the Kaggle and EvalAI leaderboards, we reveal the total number of submissions on each dataset to indicate the popularity of a certain dataset. 

\subsubsection{Datasets}
% \smallskip
% \subsubsection{KITTI Benchmark}
\noindent
\textbf{KITTI Dataset.}
KITTI~\cite{geiger2012we} is a pioneering autonomous driving dataset proposed in 2012. It has 7481 training images and 7518 test images for 3D object detection tasks. It also has corresponding point clouds captured from the Velodyne laser scanner.
The test set is split into 3 parts: easy, moderate, and hard, mainly by the bounding box size and the occlusion level.
The evaluation of object detection is of two types: 3D object detection evaluation and bird's eye view evaluation. KITTI is the first comprehensive dataset for multiple autonomous driving tasks, and it draws massive attention from the community.

% \subsubsection{Waymo Dataset}
\smallskip
\noindent
\textbf{Waymo Dataset.}
The Waymo Open Dataset v1.3~\cite{sun2020scalability} contains $798$, $202$ and $80$ video sequences in the training, validation, and testing sets, respectively. Each sequence has 5 LiDARs and 5 views of side left, front left, front, front right and side right  image resolution is $1920 \times 1280$ pixels or $1920 \times 886$ pixels. 
% Due to limited computational resource, we sample $1$ frame out of every $5$ frames from the training set to form a \emph{mini-train} set to quickly verify 
% the effect of 
% different implementations~\cite{sun2020scalability}.
Waymo is large-scale and diverse. It is evolving as the dataset version keeps updating. Each year Waymo Open Challenge would define new tasks and encourage the community to work on the problems.

% \subsubsection{nuScenes Dataset}
\smallskip
\noindent
\textbf{nuScenes Dataset.}
The nuScenes Dataset~\cite{caesar2020nuscenes} is a large scale autonomous driving dataset which contains 1000 driving scenes in two cities. 850 scenes are for training/validation and 150 are for testing. Each scene is 20s long. It has 40k keyframes with entire sensor suite including 6 cameras, 1 LiDAR and 5 Radars. The camera image resolution is $1600 \times 900$. Meanwhile, corresponding HD-Map and CAN-bus data are released to explore assistance of multiple inputs. nuScenes becomes more and more popular in the academic literature as it provides a diverse multi-sensor setting; the data scale is not as large as Waymo's, making it efficient to fast verify idea on this benchmark.

\subsubsection{Evaluation Metrics}

% \subsubsection{Waymo Open Dataset}
\textbf{LET-3D-APL.} 
In the camera-only 3D detection, LET-3D-APL is used as the metric instead of 3D-AP. 
Compared with the 3D intersection over union (IoU), the LET-3D-APL allows longitudinal localization errors of the predicted bounding boxes up to a given tolerance. 
LET-3D-APL penalizes longitudinal localization errors by scaling the precision using the localization affinity.
The definition of LET-3D-APL is mathematically defined as:
\begin{equation}
    % \mathrm{LET-3D-APL} = \int_{0}^{1} p_{L}(r)dr
    \mathrm{LET\mbox{-}3D\mbox{-}APL} = \int_{0}^{1} p_{L}(r)dr = \int_{0}^{1}\overline{a}_{l}\cdot p(r)dr,
\label{eqn:let}
\end{equation}
where $p_{L}(r)$ indicates the longitudinal affinity weighted precision value, the $p(r)$ means the precision value at recall $r$, and the multiplier $\overline{a}_{l}$ is the average longitudinal affinity of all matched predictions treated as $TP$ (true positive).

\smallskip
\noindent \textbf{mAP.}
The mean Average Precision~(mAP) is similar to the well-known AP metric in the 2D object detection, but the matching strategy is replaced from IoU to the 2D center distance on the BEV plane. The AP is calculated under different distance thresholds: 0.5, 1, 2, and 4 meters. The mAP is computed by averaging the AP in the above thresholds.

\smallskip
\noindent \textbf{NDS.}
The nuScenes detection score~(NDS) is a combination of several metrics: mAP, mATE~(Average Translation Error), mASE~(Average Scale Error), mAOE~(Average Orientation Error), mAVE~(Average Velocity Error) and mAAE~(Average Attribute Error).
The NDS is computed by using the weight-sum of the above metrics. The weight of mAP is 5 and 1 for the rest. In the first step the $\mathrm{TP_{error}}$ is converted to $\mathrm{TP_{score}}$ as shown in Eqn.~\ref{eqn:tp_score}, then Eqn.~\ref{eqn:nds} defines the NDS:
\begin{align}
    \mathrm{TP_{score}} = max(1-\mathrm{TP_{error}}, 0.0), \label{eqn:tp_score} \\
    \mathrm{NDS} = \frac{5\cdot \mathrm{mAP}+\sum_{i=1}^{5} \mathrm{TP_{score}^{i}}
}{10}.
\label{eqn:nds}
\end{align}

\begin{table*}[tb]
   \centering
   \caption{
%   Important 
   BEV perception literature in recent years. Under \textbf{Input Modality}, ``L'' for LiDAR, ``SC'' for single camera, ``MC'' for multi camera, ``T'' for temporal information. Under \textbf{Task}, ``ODet'' for 3D object detection, ``LDet'' for 3D lane detection, ``MapSeg'' for map segmentation, ``Plan'' for motion planning, ``MOT'' for multi-object tracking. \textbf{Depth Supervision} means either camera-only model uses sparse/dense depth map to supervise the model, \cmark~for yes, \xmark~for no, ${\text -}$ for LiDAR-input model. Under \textbf{Dataset}, 
   ``nuS'' %means 
   nuScenes dataset~\cite{caesar2020nuscenes}, 
   ``WOD'' %means 
   Waymo Open Dataset~\cite{sun2020scalability}, 
   ``KITTI'' %means 
   KITTI dataset~\cite{geiger2012we}, ``Lyft'' 
%   means 
   Lyft Level 5 Dataset~\cite{houston2020one}, ``OpenLane'' 
%   means 
   OpenLane dataset~\cite{chen2022persformer}, ``AV'' 
%   means 
   Argoverse Dataset~\cite{Argoverse}, ``Carla'' %means 
   carla simulator~\cite{dosovitskiy2017carla}, ``SUN'' 
%   means 
   SUN RGB-D dataset~\cite{song2015sun}, ``ScanNet'' 
%   means 
   ScanNet indoor scenes dataset~\cite{dai2017scannet}. 
   }
   \scalebox{0.92}{
   \begin{tabular}{ l |c|c|c|c|c|c }
   \toprule 
   \textbf{Method   } & \textbf{Venue }   & \textbf{Input Modality  }& \textbf{Task }        & \textbf{Depth Supervision} & \textbf{Dataset}   
& \textbf{Contribution}            \\ 
\midrule
OFT~\cite{Roddick2019OrthographicFT}  & \textit{BMVC 2019} & SC & ODet & \xmark & KITTI & Feature Projection to BEV \\
% \midrule

VoxelNet~\cite{zhou2018voxelnet} & \textit{CVPR 2018} & L & ODet & - & KITTI & Implicit voxel grids transformed to BEV \\

PointPillars~\cite{lang2019pointpillars} & \textit{CVPR 2019} & L & ODet & - & KITTI & Voxelization with pillars as BEV \\

CaDDN~\cite{reading2021categorical} & \textit{CPVR 2021} & SC & ODet & \cmark  & KITTI/WOD & Depth Distribution with Supervision \\
% \midrule
DfM~\cite{wang2022dfm}  & \textit{ECCV 2022} & SC & ODet & \cmark & KITTI & Motion to Depth to Voxel to BEV \\
% \midrule
BEVDet~\cite{huang2021bevdet}  & \textit{arXiv 2022} & MC & ODet & \xmark & nuS & BEV space data augmentation \\
PETR~\cite{liu2022petr} & \textit{arXiv 2022} & MC & ODet & \xmark & nuS & Implicit BEV Pos Embed \\
% \midrule
BEVDepth~\cite{li2022bevdepth}  & \textit{arXiv 2022} & MC/T  & ODet & \cmark & nuS & Depth Correction \\
% \midrule
ImVoxelNet~\cite{rukhovich2022imvoxelnet}  & \textit{WACV 2022} & SC/MC & ODet & \xmark & \begin{tabular}[c]{@{}c@{}}nuS/KITTI\\ SUN/ScanNet\end{tabular} & Camera Pose with Grid Sampler to BEV \\
% \midrule

\midrule
M$^2$BEV~\cite{xie2022m2bev}  & \textit{arXiv 2022} & MC & ODet/MapSeg & \xmark & nuS & BEV Representation without Depth \\
% \midrule
PolarFormer~\cite{jiang2022polarformer}  & \textit{arXiv 2022} & MC & ODet/MapSeg & \xmark & nuS & Polar-ray Representation in BEV \\
% \midrule
BEVFormer~\cite{li2022bevformer}  & \textit{ECCV 2022} & MC/T & ODet/MapSeg & \xmark & nuS/WOD & Transformer for BEV feature \\
% \midrule
BEVFusion~\cite{liu2022bevfusion}  & \textit{arXiv 2022} & MC/L & ODet/MapSeg & - & nuS & Fusion on BEV from Camera and LiDAR \\
\midrule
% \midrule
Cam2BEV~\cite{reiher2020sim2real} & \textit{ITSC 2020} & MC & MapSeg & \xmark & Synthetic & Homo-graphic Projection to BEV \\
% \midrule
% \midrule
% PYVA~\cite{yang2021projecting}  & & & & & \\
% \midrule
FIERY~\cite{hu2021fiery}  & \textit{ICCV 2021} & MC & MapSeg & \xmark & nuS/Lyft & Future Prediction in BEV space \\
CVT~\cite{zhou2022cross}  & \textit{CVPR 2022} & MC & MapSeg & \xmark & nuS & Camera Intrinsics BEV Projection \\
% \midrule
HDMapNet~\cite{li2021hdmapnet}  & \textit{ICRA 2022} & MC/L/T & MapSeg & - & nuS & Feature Fusion under BEV \\
Image2Map~\cite{saha2021translating}  & \textit{ICRA 2022} & SC/T & MapSeg & \xmark & nuS/Lyft/AV & Polar-ray Transformer in BEV  \\
\midrule
LSS~\cite{philion2020lift}  & \textit{ECCV 2020} & MC & MapSeg/Plan & \xmark & nuS/Lfyt & First Depth Distribution \\
% \midrule
ST-P3~\cite{hu2022st}  & \textit{ECCV 2022} & MC/T & MapSeg/Plan & \cmark & nuS/Carla & End to End P3 with Temporal Info \\
\midrule
% BEVerse~\cite{}  & & & & & \\
% \midrule
3D LaneNet~\cite{garnett20193d}  & \textit{ICCV 2019} & SC & LDet & \xmark & OpenLane & IPM Projection to BEV space \\
% \midrule
STSU~\cite{can2021structured}  & \textit{ICCV 2021} & SC & LDet & \xmark & nuS & Query-based Centerline in BEV \\
% \midrule
PersFormer~\cite{chen2022persformer}  & \textit{ECCV 2022} & SC & LDet & \xmark & OpenLane & Perspective Transformer for BEV \\
% \midrule
% LaRa~\cite{}  & & & & & \\
% \midrule
% \midrule
% MonoDTR~\cite{}  & & & & & \\
% \midrule
% PETRv2~\cite{}  & & & & & \\
% \midrule
% BEVSegFormer~\cite{}  & & & & & \\
% \midrule
% Ego3RT~\cite{}  & & & & & \\
% \midrule
% UVTR~\cite{}  & & & & & \\
% \midrule
% FUTR3D~\cite{}  & & & & & \\
% \midrule
% TransFusion~\cite{}  & & & & & \\
% \midrule
% AutoAlignV2~\cite{}  & & & & & \\
% \midrule
% CenterFusion~\cite{}  & & & & & \\
% \midrule

% XXX & xx \\
% XXX & xx \\
% XXX & xx \\
    \bottomrule
   \end{tabular}
   }
   
   \label{tab:method_list}
   \vspace{-0.4cm}

\end{table*}

% \clearpage

\section{Methodology of BEV Perception}\label{bev_method}

In this section, we describe in great details various perspectives of BEV perception from both academia and industry.
We differentiate BEV pipeline in three settings based on the input modality, namely {BEV Camera} (camera-only 3D perception) in Sec.~\ref{bev-camera}, {BEV LiDAR} in Sec.~\ref{bev-lidar} and {BEV Fusion} in Sec.~\ref{sec: bev-fusion}, and summarize industrial design of BEV perception in Sec.~\ref{industrial_bev_perception}.

% \noindent 
Tab.~\ref{tab:method_list} summarizes the taxonomy of BEV perception literature based on input data and task type. We can see that there is trending research for BEV perception published in top-tiered venues. The task topics as well as the formulation pipelines (contribution) can be various, indicating a blossom of the 3D autonomous driving community. 
Tab.~\ref{tab:numerical_results} depicts the performance gain of 3D object detection and segmentation on popular leaderboards over the years.
We can observe that the performance gain improves significantly in spirit of BEV perception knowledge.

% \newpage
\begin{table*}[tb]
\renewcommand\arraystretch{1.2}
\centering
    \caption{Performance comparison of BEV perception algorithms on popular benchmarks. We classify different approaches following Tab.~\ref{tab:method_list}. Under \textbf{Modality}, ``SC'', ``MC'', ``L'' denote single camera, multi camera and LiDAR, respectively. Under \textbf{Task Head}, ``Det'' designates 3D object/lane detection task, and ``Seg'' indicates BEV map segmentation task. Under \textbf{KITTI ODet}, we report $AP_{40}$ of 3D object at Easy, Moderate and Hard level in KITTI dataset~\cite{geiger2012we}. Under \textbf{nuS ODet}, we report \textbf{NDS} and $mAP$ of 3D object in nuScenes dataset~\cite{caesar2020nuscenes}. Under \textbf{nuS MapSeg}, we report mIOU score of DRI(drivable area) and LAN(lane, a.k.a divider) categories in nuScenes Map Segmentation setting. Under \textbf{OL}, we report F1 score of 3D laneline in OpenLane dataset~\cite{chen2022persformer}. Under \textbf{WOD}, we report LET-APL~\cite{hung2022let3dap} for camera-only 3D object detection and APH/L2~\cite{sun2020scalability} for any modality 3D object detection in Waymo Open Dataset~\cite{sun2020scalability}. $^*$ denotes results reported by the original papers.
    % with specific settings.
    }
\begin{center}
    \begin{tabular}{c|c|c|c|c|c|c|c|c|c|c|c|c|c|c|c} % p{2.2cm}<{\centering}
    \toprule
    \multirow{2}{*}{\textbf{Methods}} &
    % \multirow{2}[]{c|}{\textbf{Methods}} &
    \multicolumn{3}{c|}{\textbf{Modality}} &
    \multicolumn{2}{c|}{\textbf{Task Head}} & \multicolumn{3}{c|}{\textbf{KITTI ODet \%}} & \multicolumn{2}{c|}{\textbf{nuS ODet}} & \multicolumn{2}{c|}{\textbf{nuS MapSeg}} &
    \multicolumn{1}{c|}{\textbf{OL}} &
    \multicolumn{2}{c}{\textbf{WOD}}  \\
    \cline{2-16}
    % \cline{2-16}
    ~ & \scriptsize{SC} & \scriptsize{MC} & \scriptsize{L} & \scriptsize{Det} & \scriptsize{Seg} & \scriptsize{Easy}  & \scriptsize{Mod.}  & \scriptsize{Hard}  & \scriptsize{NDS}  & \scriptsize{mAP}  & \scriptsize{DRI}  & \scriptsize{LAN}  & \scriptsize{F1}  & \scriptsize{LET}  & \scriptsize{APH/L2} \\
    \midrule
    OFT~\cite{Roddick2019OrthographicFT} & \cmark & - & - & \cmark & - & 2.50 & 3.28 & 2.27 & - & - & - & - & - & - & - \\ 
    % AM3D~\cite{ma2019accurate} & \cmark & - & - & \cmark & - & 16.50 & 10.74 & 9.52 & - & - & - & - & - & - & - \\
    CaDDN~\cite{reading2021categorical} & \cmark & - & - & \cmark & - & 19.17 & 13.41 & 11.46 & - & - & - & - & - & - & - \\
    ImVoxelNet~\cite{rukhovich2022imvoxelnet} & \cmark & \cmark & - & \cmark & - & 17.15 & 10.97 & 9.15 & - & 0.518$^{*}$ & - & - & - & - & - \\
    % SMOKE~\cite{liu2020smoke} & \cmark & - & - & \cmark & - & 14.03 & 9.76 & 7.84 & - & - & - & - & - & - & - \\
    DfM~\cite{wang2022dfm} & \cmark & - & - & \cmark & - & 22.94 & 16.82 & 14.65 & - & - & - & - & - & - & - \\
    PGD~\cite{wang2022probabilistic} & \cmark & - & - & \cmark & - & 19.05 & 11.76 & 9.39 & 0.448 & 0.386 & - & - & - & 0.5127 & - \\
         BEVerse~\cite{zhang2022beverse} & - & \cmark & - & \cmark & - & - & - & - & 0.531 & 0.393 & - & - & - & - & - \\
    PointPillars~\cite{lang2019pointpillars} & - & - & \cmark & \cmark & - & - & - & - &  0.550 & 0.401 &   - & - & - & - & - \\
         BEVDet~\cite{huang2021bevdet} & - & \cmark & - & \cmark & - & - & - & - & 0.552 & 0.422 & - & - & - & - & - \\
     BEVDet4D~\cite{huang2022bevdet4d} & - & \cmark & - & \cmark & - & - & - & - & 0.569 & 0.451 & - & - & - & - & - \\
     BEVDepth~\cite{li2022bevdepth} & - & \cmark & - & \cmark & - & - & - & - & 0.609 & 0.520 & - & - & - & - & - \\
     DSGN~\cite{chen2020dsgn} & - & \cmark& -  & \cmark & - & 73.50 & 52.18 & 45.14 & - & - &   - & - & - & - & - \\
     PV-RCNN~\cite{shi2019pvrcnn} & - & - & \cmark & \cmark & - & - & - & - & - & - &   - & - & - & - & 0.7152 \\
    CenterPoint~\cite{yin2021center} & - & - & \cmark & \cmark & - & - & - & - & 0.673 & 0.603 &   - & - & - & - & 0.7193 \\
    SST~\cite{fan2022sst} & - & - & \cmark & \cmark & - & - & - & - & - & - &   - & - & - & - & 0.7281 \\
    %CenterPoint++~\cite{yin2021center}& - & - & \cmark & \cmark & - & - & - & - & - & - &   - & - & - & - & 0.7282 \\   
    AFDetV2~\cite{hu2022afdetv2} & - & - & \cmark & \cmark & - & - & - & - & 0.685 &  0.624 &   - & - & - & - & 0.7312 \\     
     PV-RCNN++~\cite{shi2021pvrcnn++} & - & - & \cmark & \cmark & - & - & - & - & - & - &   - & - & - & - & 0.7352 \\
     Pyramid-PV~\cite{mao2021pyramid} & - & - & \cmark & \cmark & - & - & - & - & - & - &   - & - & - & - & 0.7443 \\
     MPPNet~\cite{chen2022mppnet} & - & - & \cmark & \cmark & - & - & - & - & - & - &   - & - & - & - & 0.7567 \\

    \midrule
    M$^2$BEV~\cite{xie2022m2bev} & - & \cmark & - & \cmark & \cmark & - & - & - & 0.474 & 0.429 & 77.3 & 40.5 & - & - & - \\
     BEVFormer~\cite{li2022bevformer} & - & \cmark & - & \cmark & \cmark & - & - & - & 0.569 & 0.481 & 80.1 & 25.7 & - & 0.5616 & - \\
     PolarFormer~\cite{jiang2022polarformer} & - & \cmark & - & \cmark & \cmark & - & - & - & 0.572 & 0.493 & 82.6 & 46.2 & - & - & - \\
     PointPainting~\cite{vora2020pointpainting}& - & \cmark & \cmark & \cmark & - & - & - & - & 0.581 &  0.464 &- & - & - & - & - \\
     MVP~\cite{yin2021mvp}& - & \cmark & \cmark & \cmark & - & - & - & - & 0.705 &  0.664 &- & - & - & - & - \\
     AutoAlign~\cite{chen2022autoalign}& - & \cmark & \cmark & \cmark & - & - & - & - & 0.709 &  0.658 &- & - & - & - & - \\
     %PointAugmenting~\cite{wang2021pointaugmenting} & - & \cmark & \cmark & \cmark & - & - & - & - & 0.711 & 0.668 &- & - & - & - & - \\
     TransFusion~\cite{bai2022transfusion}& - & \cmark & \cmark & \cmark & - & - & - & - & 0.717 & 0.689 & - &- & - & - & - \\
     DeepFusion~\cite{li2022deepfusion} & - & \cmark & \cmark & \cmark & - & - & - & - & - & - &   - & - & - & - & 0.7554 \\
     AutoAlignV2~\cite{chen2022autoalignv2} & - & \cmark & \cmark & \cmark & - & - & - & - & 0.724 & 0.684 &- & - & - & - & - \\
     BEVFusion~\cite{liu2022bevfusion} & - & \cmark & \cmark & \cmark & \cmark & - & - & - & 0.761 & 0.750 & 85.5 & 53.7 & - & - & 0.7633 \\

     \midrule
     3D-LaneNet~\cite{garnett20193d} & \cmark & - & - & \cmark & - & - & - & - & - & - & - & - & 0.441 & - & - \\
     PersFormer~\cite{chen2022persformer} & \cmark & - & - & \cmark & - & - & - & - & - & - & - & - & 0.505 & - & - \\
    \bottomrule
    \end{tabular}
\end{center}
\label{tab:numerical_results}
\vspace{-0.4cm}
\end{table*}

\subsection{BEV Camera}\label{bev-camera}
% written by smch
\subsubsection{General Pipeline}
Camera-only 3D perception attracts a large amount of attention from academia.
The core issue is that 2D imaging process inherently cannot preserve 3D information, hindering precise object localization without accurate depth extraction.
Camera-only 3D perception can be split into three domains: monocular setting, stereo setting and multi-camera setting.
%
% In this survey, we mainly focus on single-camera and multi-camera setting.
%
Since multi-camera methods often start with a monocular baseline, we start with monocular baseline setting as well.
We use ``2D space'' to refer to perspective view with pixel coordinates, ``3D space'' to refer to 3D real-world space with world coordinates, and ``BEV space'' to refer to bird's eye view in the following context.

\begin{figure*}
\centering
\includegraphics[width=0.9\textwidth]{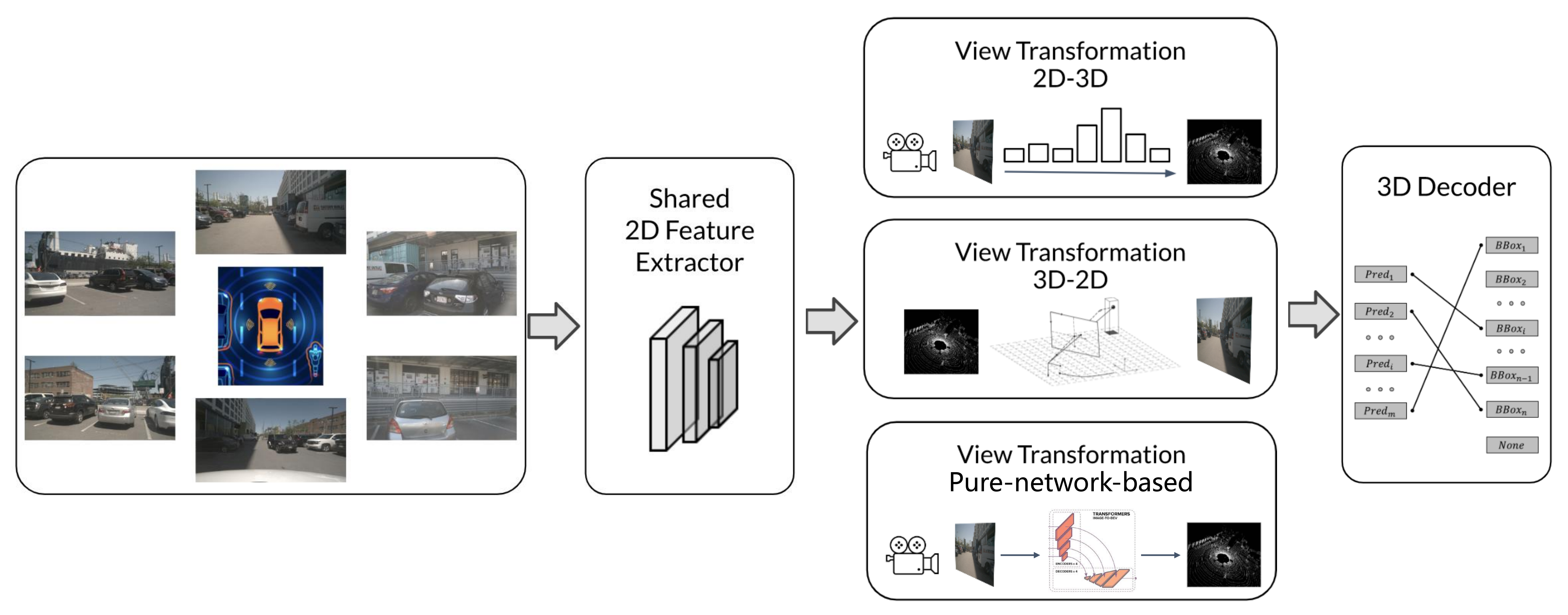}
\caption{The general pipeline of BEV Camera (camera-only perception). There are three parts, including 2D feature extractor, view transformation and 3D decoder. In view transformation, there are two ways to encode 3D information -  one is to predict depth information from 2D feature; the other is to sample 2D feature from 3D space.
% , namely {3D-2D}. 
%
% More details about view transformation are listed in Sec.~\ref{sec:view transform} and Fig.~\ref{fig:camera_project}
}
\label{fig:camera_pipeline}
\vspace{-0.4cm}

\end{figure*}

As depicted in Fig.~\ref{fig:camera_pipeline}, a general camera-only 3D perception system can be divided into three parts: {2D feature extractor}, {view transformation module} (optional), and {3D decoder}.
As camera-only 3D perception has the same input as 2D perception, the general feature extractor can be formulated as:
\begin{equation}\label{eqn:2d feat}
    \mathcal{F}_{2D}^{*}(u,v) = M_{feat}(\mathcal{I}^{*}(u,v)),
\end{equation}
where $\mathcal{F}_{2D}$ denotes 2D feature, $\mathcal{I}$ denotes image, $M_{feat}$ denotes 2D feature extractor, $u,v$ denote coordinates on 2D plane and $^{*}$ denotes one or more images and corresponding 2D features.
In 2D feature extractor, there exists a huge amount of experience in 2D perception that can be considered in 3D perception, in the form of backbone pretraining~\cite{he2019rethinking, park2021pseudo}.
The view transformation module largely differentiates from 2D perception system.
%
% First, it's worth noticing 
Note that not all 3D perception methods have a view transformation module, and some detect objects in 3D space directly from features in 2D space~\cite{wang2021fcos3d, park2021pseudo, liu2020smoke}.
As depicted in Fig.~\ref{fig:camera_pipeline}, there are generally three ways to perform view transformation.
Such a transformation can be formulated as:
\begin{equation}\label{eqn:view_trans}
    \mathcal{F}_{3D}(x,y,z) = M_{trans} \big(\mathcal{F}_{2D}^{*}(\hat{u}, \hat{v}), \begin{bmatrix}
\bm{R} & \bm{T}
\end{bmatrix}, \bm{K} \big),
\end{equation}
where $\mathcal{F}_{3D}$ denotes 3D (or voxel) feature, $x, y, z$ denote coordinates in 3D space, $M_{trans}$ denotes view transformation module, $\hat{u}, \hat{v}$ denote corresponding 2D coordinates in terms of $x, y, z$, $\begin{bmatrix}
\bm{R} & \bm{T}
\end{bmatrix}$ and $\bm{K}$ are camera extrinsics and intrinsics as described
% Eqn.~\ref{prel-3dv-E5}.
in Sec.~\ref{Preliminary_3dvision} of Appendix. Note that some methods do not rely on camera extrinsics and intrinsics.
The 3D decoders receive the feature in 2D/3D space and output 3D perception results such as 3D bounding boxes, BEV map segmentation, 3D lane keypoints and so on.
Most 3D decoders come from LiDAR-based methods~\cite{zhou2018voxelnet, yin2021center, tang2020spvcnn, yan2018second} which perform detection in voxel-space/BEV space, but there are still some camera-only 3D decoders that utilize the feature in 2D space~\cite{wang2021fcos3d, wang2022detr3d, liu2020smoke} and directly regress the localization of 3D objects.

%
% A general roadmap is depicted in Fig.~\ref{fig:camera_project}.
%
% As a side note, there are few attempt to solve 3D pretraining~\cite{park2021pseudo} and transfer recent success in 2D vision Transformer~\cite{li2022dn, meng2021conditional} to 3D space; this could be future investigation in 3D perception.

\subsubsection{View Transformation}\label{sec:view transform}
% \smch{formula and numbers}

\begin{figure}
\centering
\includegraphics[width=0.48\textwidth]{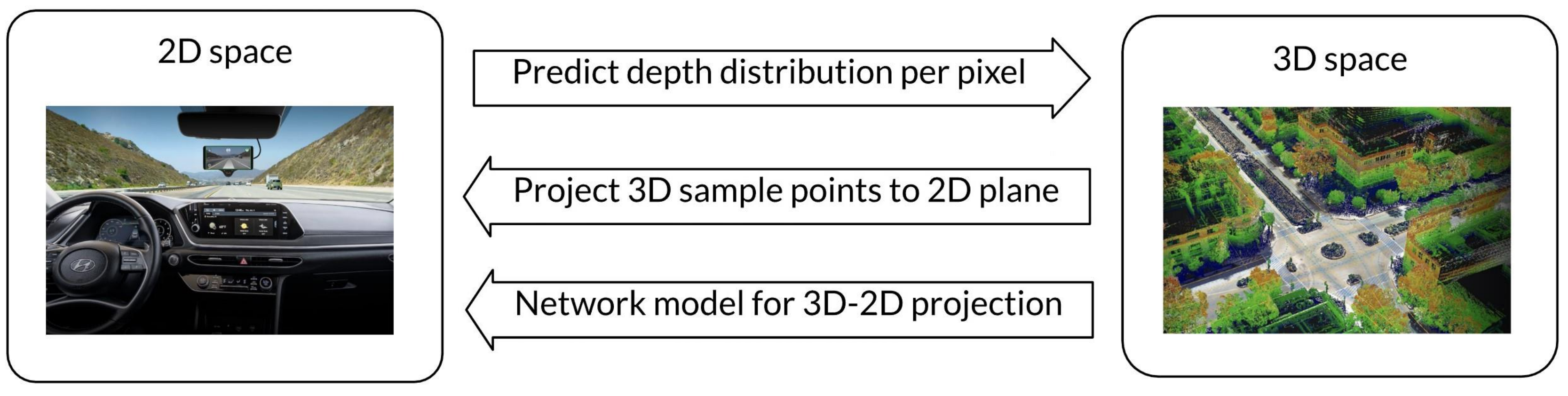}
\caption{
% Two major approaches to implement view transformation. 
Taxonomy of View Transformation.
From the \textbf{2D-3D} methods, LSS-based approaches~\cite{philion2020lift, reading2021categorical, huang2021bevdet, huang2022bevdet4d, li2022bevdepth, liu2022bevfusion, liang2022bevfusion} predict depth distribution per pixel from 2D feature. 
From the \textbf{3D-2D} methods, homographic matrix based methods~\cite{li2022bevformer, chen2022persformer, gong2022gitnet} presume sparse 3D sample points and project them to 2D plane via camera parameters. \textbf{Pure-network-based} methods
~\cite{pan2020cross, hendy2020fishing, chitta2021neat, yang2021projecting, gosala2022bird}
adopt MLP or transformer to implicitly model the projection from 3D space to 2D plane.
}
\label{fig:camera_project}
\vspace{-0.4cm}
\end{figure}

{View transformation module is pivotal in camera-only 3D perception, as it serves as the primary unit for constructing 3D data and encoding 3D prior assumptions. Recent studies~\cite{li2022bevformer, li2022bevdepth, huang2021bevdet, liu2022petr, jiang2022polarformer, wang2022dfm, xie2022m2bev, saha2021translating, chen2022persformer, garnett20193d} have centered on enhancing this module. }
{We divide the view transformation techniques into three principal streams. The first stream, designated as "2D-3D methods", commences with 2D image features and ``lifts'' 2D features to 3D space via depth estimation. The second stream, recognized as "3D-2D methods", originates in 3D space, and encode 2D feature to 3D space via 3D-to-2D projection mapping. The first two stream explicitly model geometric transformation relationships. In contrast, the third stream, denoted as "pure-network-based methods", utilizes neural networks to implicitly acquire the geometric transformation.}
% %
% Constructing 3D information from 2D feature is usually formulated as depth estimation or cost volume.
% %
% Constructing 3D information from 3D prior assumption is usually formulated as sampling 2D feature to construct 3D feature via 3D-2D projection mapping.
% %
% \textcolor{red}{ it can be divided into three aspects, the first one is to utilize 2D feature to construct depth information and ``lift'' 2D feature to 3D space, the second one is to encode 2D feature to 3D space via 3D-to-2D projection mapping, and the last one directly learns the 2D-to-3D space transformation  without relying on any priors.}
% %
% We name the first method as \textit{2D-3D} and the second one \textit{3D-2D}. 
%
Fig.~\ref{fig:camera_project} presents a summary roadmap of performing view transformation, and they are analyzed in details below.

\textbf{$\left(1\right)$ 2D-3D methods:} 
%The \textit{2D-3D} method is introduced by LSS~\cite{philion2020lift}, where it predicts depth distribution per grid on 2D feature, then ``lift'' the 2D feature via the corresponding depth to voxel space, and perform downstream tasks following LiDAR-based methods. %This process can be formulated as:
The \textit{2D-3D} method, firstly introduced by LSS~\cite{philion2020lift}, predicts grid-wise depth distribution on 2D features, then ``lifts'' the 2D features to voxel space based on depth, and conducts downstream tasks similar to LiDAR-based methods. This process can be formulated as:
\begin{equation}
    \mathcal{F}_{3D}(x,y,z) = \big[\mathcal{F}_{2D}^{*}(\hat{u}, \hat{v}) \otimes \mathcal{D}^{*}(\hat{u}, \hat{v}) \big ]_{xyz}, 
\end{equation}
where $\mathcal{F}_{3D}(x,y,z)$ and $\mathcal{F}_{2D}^{*}(\hat{u}, \hat{v})$ remain the same meaning as Eqn.~\ref{eqn:view_trans}, $\mathcal{D}^{*}(\hat{u}, \hat{v})$ represents predicted depth value or distribution at $(\hat{u}, \hat{v})$, and $\otimes$ denotes outer production or similar operations.
% \smch{introduce LSS or CaDDN formula}
%
Note that this is very different from pseudo-LiDAR methods~\cite{wang2019pseudo, you2019pseudo} whose depth information is extracted from a pretrained depth estimation model and the lifting process occurs before 2D feature extraction.
After LSS~\cite{philion2020lift}, there is another work following the same idea of formulating depth as bin-wise distribution, namely CaDDN~\cite{reading2021categorical}.
CaDDN employs a similar network to predict categorical depth distribution,
squeezes the voxel-space feature down to BEV space, 
and performs 3D detection at the end.
The main difference between LSS~\cite{philion2020lift} and CaDDN~\cite{reading2021categorical} is that CaDDN uses depth ground truth to supervise its categorical depth distribution prediction, thus owing a superior depth network to extract 3D information from 2D space.
%Note that when we claim ``a better depth network'', it is actually learning an implicit projection between the road surface and the perspective view at feature level.
%
This track comes subsequent work such as BEVDet~\cite{huang2021bevdet} and its temporal version BEVDet4D~\cite{huang2022bevdet4d}, BEVDepth~\cite{li2022bevdepth}, 
% the camera branch in 
BEVFusion~\cite{liu2022bevfusion, liang2022bevfusion}, and others~\cite{chen2020dsgn, guo2021liga, park2021pseudo}.
%
% It's worth noting 
Note that in stereo setting, depth value/distribution is much more easier to be obtained via the strong prior where the distance between the pair of cameras (namely \textit{baseline} of the system) are supposed to be constant.
This can be formulated as:
\begin{equation}
    \mathcal{D}(u, v) = f \times \frac{b}{d(u,v)},
\end{equation}
where $d(u,v)$ is the horizontal disparity on the pair of images at location $(u,v)$ (usually defined in the left image), $f$ is the focal length of cameras as
% in Eqn.~\ref{prel-3dv-E5},
in Sec.~\ref{Preliminary_3dvision} of Appendix,
$\mathcal{D}(u, v)$ is the depth value at $(u,v)$, and $b$ is the length of the baseline.
Stereo methods such as LIGA Stereo~\cite{guo2021liga} and DSGN~\cite{chen2020dsgn} utilize this strong prior and perform 
% at an approximate level 
on par with LiDAR-based alternatives on KITTI leaderboard~\cite{geiger2012we}. 
% compared to LiDAR-based methods.
% \smch{introduce stereo formula}
% SMCH: need MORE literature
% \smch{review to here}

%
%
\textbf{$\left(2\right)$ 3D-2D methods:} The second branch (3D to 2D) can be originated back to thirty years ago, when Inverse Perspective Mapping (IPM) \cite{mallot1991inverse} formulated the projection from 3D space to 2D space conditionally hypothesizing that the corresponding points in 3D space lie on a horizontal plane.
%
% \smch{introduce IPM formula}
%
Such a transformation matrix can be mathematically derived from camera intrinsic and extrinsic parameters~\cite{andrew2001multiple}, and the detail of this process is presented in Sec.~\ref{Preliminary_3dvision} of Appendix.
%\smch{review to here}
%
% Later, 
% A series of work~\cite{kim2019deep, palazzi2017learning, zhu2021monocular, loukkal2021driving, can2022understanding, sengupta2012automatic, song2021stacked}
A series of work 
apply IPM to transform elements from perspective view to bird's eye view in an either pre-processing or post-processing manner.
In context of view transformation, feature projection method from 3D to 2D is first introduced by OFT-Net~\cite{Roddick2019OrthographicFT}. 
OFT-Net forms a uniformly distributed 3D voxel feature grid, populating voxels by aggregating image features from corresponding projection regions. The orthographic BEV feature map is then acquired by summing voxel features vertically.
%where it projects 2D feature to voxel space (3D space).
%It is based on assumption that the depth distribution is uniform along the ray starting from camera origin to a specific point in 3D space.
%
%
Recently, inspired by Tesla's technical roadmap for the perception system~\cite{tesla_ai_day}, a combination of 3D-2D geometric projection and neural network becomes popular \cite{li2022bevformer, chen2022persformer, wang2022detr3d, gong2022gitnet}.
Note that the cross-attention mechanism in transformer architecture conceptually meets the need of such a geometric projection, as denoted by:
\begin{equation}
    \mathcal{F}_{3D}(x,y,z) = CrossAttn(q:P_{xyz}, k~v: \mathcal{F}_{2D}^{*}(\hat{u}, \hat{v})),
\end{equation}
where $q,k,v$ stand for query, key and value, $P_{xyz}$ is pre-defined anchor point in voxel space, and other notations follow Eqns.~\ref{eqn:2d feat} and ~\ref{eqn:view_trans}.
% \smch{introduce transformer formula, especially cross-attention}
%
Some approaches~\cite{li2022bevformer, wang2022detr3d} utilize camera parameters to project $P_{xyz}$ to image plane for fast convergence of the model.
To obtain robust detection results, BEVFormer~\cite{li2022bevformer} exploits the cross-attention mechanism in transformer to enhance the modeling of \textit{3D-2D} view transformation.
Others~\cite{rukhovich2022imvoxelnet, wang2022mvfcos3d++} alleviate grid sampler to accelerate this process efficiently for massive production.
Nonetheless, these approaches heavily rely on the precision of camera parameters, which are susceptible to fluctuations over extended durations of driving.
%
% \smch{XXXX}
% SMCH: need MORE literature

%

\textbf{$\left(3\right)$ Pure-network-based methods:}  Regardless of 2D-3D methods or 3D-2D methods, both techniques introduce inherit inductive bias contained in geometric projection. In contrast, some methods tend to ultilize neural networks for the implicit representation of camera projection relationships.
A lot of BEV map segmentation works~\cite{pan2020cross,saha2021translating,li2021hdmapnet}
utilize either multi-layer perceptron or transformer~\cite{vaswani2017attention} architecture to model the 3D-2D projection implicitly.
VPN~\cite{pan2020cross} introduces view relation module—a Multilayer Perceptron (MLP) for producing map-view features by processing inputs from all views, thus achieves the acquisition of a shared feature representation spanning various perspectives.
HDMapNet~\cite{li2021hdmapnet} employs MLP architecture to execute the view transformation of feature maps. BEVSegFormer builds dense BEV queries, and directly predicts their 2D projection points from query feature through MLP, then update the query embedding using deformable attention. CVT~\cite{zhou2022cross} incorporate image features with camera-aware position embedding derived from the camera intrinsic and extrinsic parameters, and introduces a cross-view attention module to produce map-view representation.
Some methods do not explicitly construct BEV features. PETR~\cite{liu2022petr} integrates 3D positional embedding, derived from camera parameters, into 2D multi-view features. This integration empowers sparse queries to directly interact with 3D position-aware image features through the vanilla cross attention.

%\textbf{$\left(4\right)$ Other methods:}

\subsubsection{Discussion on BEV and perspective methods}
At the very beginning of camera-only 3D perception, the main focus is how to predict 3D object localization from perspective view, \textit{a.k.a. 2D space}.
It is because 2D perception is well developed at that stage~\cite{tian2019fcos, ren2015faster, girshick2015fast, he2017mask}, and how to equip a 2D detector with the ability to perceive 3D scene becomes 
% mainstream methods~\cite{wang2021fcos3d, xu2018multi, chen20153d, mousavian20173d, brazil2019m3d, jrgensen2019monocular, simonelli2019disentangling, ma2021delving, wang2022probabilistic, zhang2021objects, shi2021geometry, li2019stereo, sun2020disp, xu2020zoomnet, peng2021side, liu2020smoke}.
mainstream methods~\cite{wang2021fcos3d, xu2018multi, wang2022probabilistic, liu2020smoke}.
Later, some research reached to BEV representation, since under this view, it is easy to tackle the problem where objects with the same size in 3D space have very different size on image plane due to the distance to camera.
This series of work~\cite{wang2019pseudo, Roddick2019OrthographicFT, reading2021categorical, chen2020dsgn, guo2021liga} either predict depth information or utilize 3D prior assumption to compensate the loss of 3D information in camera input.
While recent BEV-based methods~\cite{li2022bevformer, li2022bevdepth, huang2021bevdet, liu2022bevfusion, ng2020bev, liang2022bevfusion, xie2022m2bev} has taken the 3D perception world by storm, it is worth noticing that this success has been beneficial from mainly three parts.
The first reason is the trending nuScenes dataset~\cite{caesar2020nuscenes}, which has multi-camera setting and it is very suitable to apply multi-view feature aggregation under BEV.
The second reason is that most camera-only BEV perception methods have gained a lot of help from LiDAR-based methods~\cite{tang2020spvcnn, yin2021center, qi2017pointnet, qi2017pointnet++, lang2019pointpillars, yan2018second, zhou2018voxelnet} in form of detection head and corresponding loss design.
The third reason is that the long-term development of monocular methods~\cite{wang2021fcos3d, xu2018multi, liu2020smoke} has flourished BEV-based methods a good starting point in form of handling feature representation in the perspective view.
%
% Many papers claimed that performing 3D tasks under BEV can avoid directly estimating depth from perspective view as it is a difficult task.
% %
% This is a typical "Yes and No".
% %
% On one hand, 
% %
% On the other hand, \smch{XXX}
%
The core issue is how to reconstruct the lost 3D information from 2D images.
To this end, BEV-based methods and perspective methods are two different ways to resolve the same problem, and they are not excluding each other.
% \smch{review to here, section 4}

\subsection{BEV LiDAR}\label{bev-lidar}
% written by huijie

\subsubsection{General Pipeline}

%Nevertheless, point-level features lead to high computation overheads due to unordered storage. In contrast, the voxel-based structure is better suited for feature extraction but often yields lower accuracy because the input data are divided into grids.

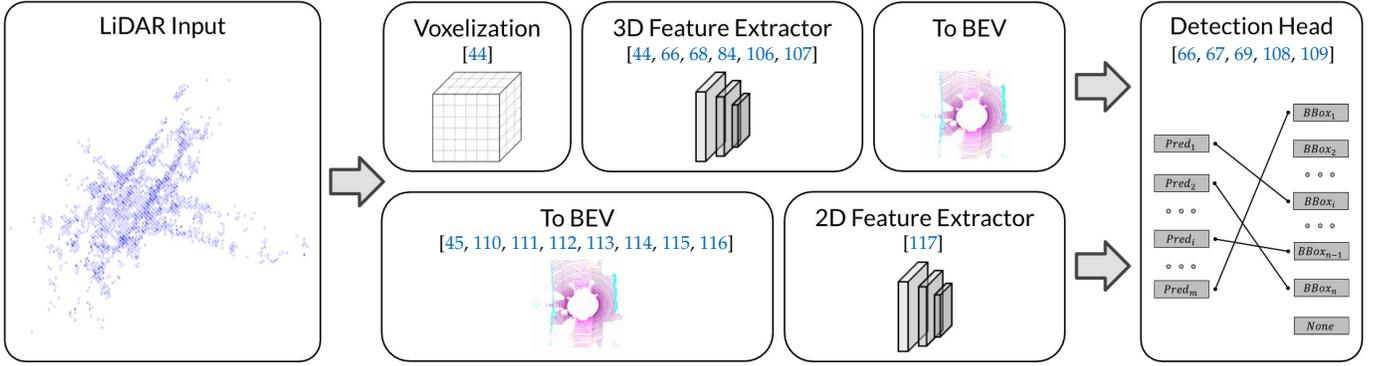
\begin{figure*}
\centering
\begin{overpic}[width=1.0\linewidth]{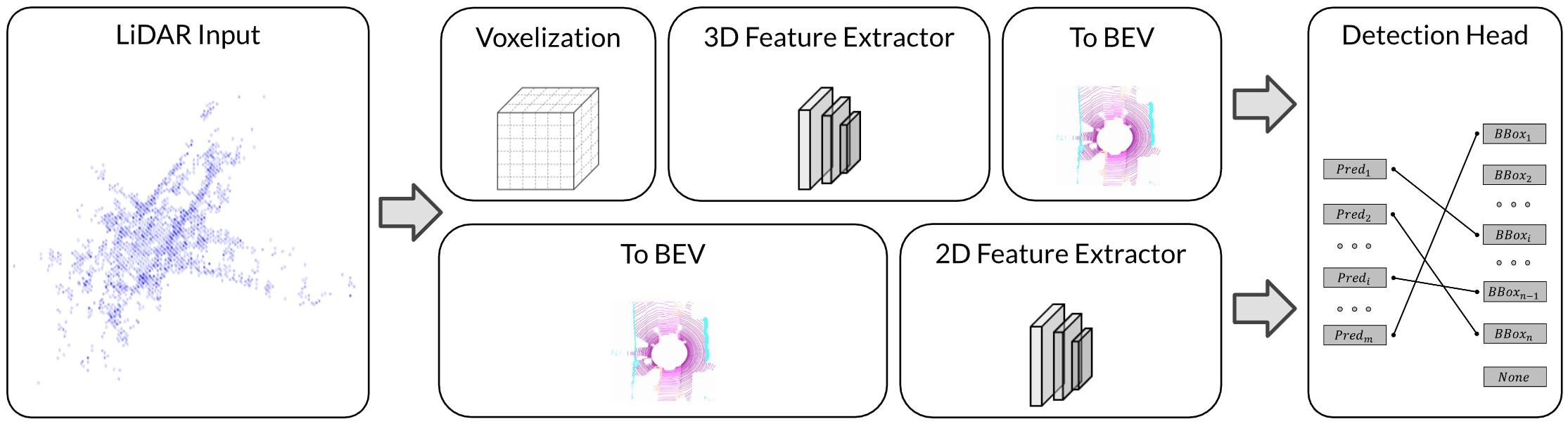}
\put(33.5,22.4){\scriptsize~\cite{zhou2018voxelnet}}
 \put(45.5,22.4){\scriptsize~\cite{yan2018second, zhou2018voxelnet, fan2022sst, he2020sassd, shi2019pvrcnn, mao2021votr}}
 \put(85.3,22.4){\scriptsize~\cite{yin2021center, shi2019pvrcnn, deng2020voxelrcnn, wang2021objectdgcnn, hu2022afdetv2}}
 \put(31.8,8.6){\scriptsize~\cite{lang2019pointpillars,chen2017multi,yang2018pixor, billard2018hd, beltran2018birdnet, zeng2018rt3d, ali2018yolo3d, simony2018complex}}
 \put(65.7,8.6){\scriptsize~\cite{he2016deep}}
\end{overpic}
\caption{The general pipeline of BEV LiDAR perception. There are mainly two branches to convert point cloud data into BEV representation. The upper branch extracts point cloud features in 3D space, providing more accurate detection results. The lower branch extracts BEV features in 2D space, providing more efficient networks.}
\label{fig:roadmap_lidar}
\vspace{-0.4cm}

\end{figure*}

Fig.~\ref{fig:roadmap_lidar} depicts the general pipeline for BEV LiDAR perception.
The extracted point cloud features are transformed into BEV feature maps.
The common detection heads generate 3D prediction results. 
In terms of the feature extraction part, 
there are mainly two branches to convert point cloud data into BEV representation.
According to the pipeline order, we term these two options as pre-BEV and post-BEV respectively, indicating whether
the input of 
backbone network is from 3D representation or BEV representation.

\subsubsection{Pre-BEV Feature Extraction}

%Intuitively, convolution can be described as the amount of overlap of one function as it is shifted over another function. When talking about convolution in \ac{CNN}s, it usually means cross-correlation. For an input image $x$, which is 2-dimensional, convolution at position $(i, j)$ can be written as:
%\begin{equation}
%    \begin{aligned}
%        conv(i, j) 
%        &= (x * K)(i, j)
%        \\ 
%        &= \sum_{h}^{H} \sum_{w}^{W} x(h, w) \cdot K(i - h, j - w),
%    \end{aligned}
%\end{equation}
%where $K$ is the kernel, and $H$ and $W$ are the height and width of the kernel, respectively. Similar to \ac{MLP}s, it is followed by an element-wise activation function.

%Typically, a frame of point cloud data contains about a hundred thousand points, pint-based methods usually waste time on processing irregular data. 
% relied on the volumetric representation and vanilla 3D convolution to process the 3D data
% the dense volumetric representation is inherently inefficient, and it also inevitably introduces information loss
% the resolution of dense voxels  is strictly constrained by the memory;
% Specifically, spatially sparse convolutional networks are introduced for LiDAR-based detection and are used to extract information from the z-axis before the 3D data are downsampled to something akin to 2D image data.

Besides point-based methods processing on the raw point cloud, voxel-based methods voxelize points into discrete grids, which provides a more efficient representation by discretizing the continuous 3D coordinate. 
Based on the discrete voxel representation, 3D convolution or 
% 3D sparse convolution~\cite{graham2014sparse, graham2017submanifold, tang2022torchsparse, choy2019minkowski, graham2018submanifold}
3D sparse convolution~\cite{graham2014sparse, choy2019minkowski}
can be used to extract point cloud features. 
%at the location with index $j$ at the location with index $i$
We use $Y_{j, c'}$ to represent the $j$-th voxel output $Y$ at output channel $c'$, and $X_{i, c}$ to represent the $i$-th voxel input $X$ at input channel $c$.
% \smch{what is index j? in the point cloud? voxel? - DONE} 
%
A normal 3D convolutional operation can be described as:
\begin{equation}
    Y_{j, c'} = \sum_{i \in P(j)} \sum_{c} W_{k,c,c'} X_{i, c},
\label{eqn:3dconv}
\end{equation}
where $P(j)$ denotes a function for obtaining the input index $i$ and the filter offset, and $W_{k,c,c'}$ denotes filter weight with kernel offset $k$. For sparse input $\tilde{X}$ and output $\tilde{Y}$, we can rewrite Eqn.~\ref{eqn:3dconv} into 3D sparse convolution:
\begin{equation}
    \tilde{Y}_{j, c'} = \sum_{k} \sum_{c} W_{k,c,c'} \tilde{X}_{R_{k, j}, k, c},
\label{eqn:3dsparseconv}
\end{equation}
where $R_{k, j}$ denotes a matrix that specifies the input index $i$ given the kernel offset $k$ and the output index $j$. Most state-of-the-art methods normally utilize 3D sparse convolution to conduct feature extraction. The 3D voxel features can then be formatted as a 2D tensor in BEV by densifying and compressing the height axis.

VoxelNet~\cite{zhou2018voxelnet} stacks multiple voxel feature encoding (VFE) layers to encode point cloud distribution in a voxel as a voxel feature. Given $\mathbf{V} = \{\mathbf{p_i} = [x_i, y_i, z_i, r_i]^T\}_{i=1...n}$ as $n \leq N$ points inside a non-empty voxel, where $x_i, y_i, z_i$ are coordinates in 3D space, $r_i$ is reflectance,  $N$ is the maximal number of points, and the centroid of $\mathbf{V}$ $(v_x, v_y, v_z)$ is the local mean of all points, the feature of each point is calculated by:
\begin{equation}
    f_i = FCN([x_i, y_i, z_i, r_i, x_i-v_x, y_i-v_y, z_i-v_z]^T).
\end{equation}
FCN is the composition of a linear layer, a batch normalization, and an activation function. Feature of the voxel is the element-wise max-pooling of all $f_i$ of $\mathbf{V}$. 
%Then 
A 3D convolution is applied to further aggregate local voxel features. After merging the dimension of channel and height, the feature maps, which are transformed implicitly into BEV, are processed by a region proposal network (RPN) to generate object proposals. SECOND~\cite{yan2018second} introduces sparse convolution in processing voxel representation to reduce training and inference speed by a large margin. CenterPoint~\cite{yin2021center}, which is a powerful center-based anchor-free 3D detector, also follows this detection pattern and becomes a baseline method for 3D object detection.

PV-RCNN~\cite{shi2019pvrcnn} combines point and voxel branches to learn more discriminative point cloud features. Specifically, high-quality 3D proposals are generated by the voxel branch, and the point branch provides extra information for proposal refinement. SA-SSD~\cite{he2020sassd} designs an auxiliary network, which converts the voxel features in the backbone network back to point-level representations, to explicitly leverage the structure information of the 3D point cloud and ease the loss in downsampling. Voxel R-CNN~\cite{deng2020voxelrcnn} adopts 3D convolution backbone to extract point cloud feature. A 2D network is then applied on the BEV to provide object proposals, which are refined via extracted features. It achieves comparable performance with point-based methods. Object DGCNN~\cite{wang2021objectdgcnn} models the task of 3D object detection as message passing on a dynamic graph in BEV. After transforming point cloud into BEV feature maps, predicted query points collect BEV features from key points iteratively. VoTr~\cite{mao2021votr} introduces Local Attention, Dilated Attention, and Fast Voxel Query to enable attention mechanism on numerous voxels for large context information. SST~\cite{fan2022sst} treats extracted voxel features as tokens and then applies Sparse Regional Attention and Region Shif in the non-overlapping region to avoid downsampling for voxel-based networks. AFDetV2~\cite{hu2022afdetv2} formulates a single-stage anchor-free network by introducing a keypoint auxiliary supervision and multi-task head.

% Besides using pure BEV representation, 

\subsubsection{Post-BEV Feature Extraction}

% PIXOR~\cite{yang2018pixor} keep the height information as channels assuming objects of interest are on the same ground.

As voxels in 3D space are sparse and irregular, applying 3D convolution is inefficient. For industrial applications, operators such as 3D convolution may not be supported; suitable and efficient 3D detection networks are desirable. 

MV3D~\cite{chen2017multi} is the first method to convert point cloud data into a BEV representation. After discretizing points into the BEV grid, the features of height, intensity, and density are obtained according to points in the grid to represent grid features. As there are many points in a BEV grid, in this processing, the loss of information is considerable. Other works~\cite{yang2018pixor, billard2018hd, beltran2018birdnet, zeng2018rt3d, ali2018yolo3d, simony2018complex} follow the similar pattern to represent point cloud using the statistic in a BEV grid, such as the maximum height and mean of intensity.

PointPillars~\cite{lang2019pointpillars} first introduces the concept of pillar, which is a special type of voxel with unlimited height. It utilizes a simplified version of PointNet~\cite{qi2017pointnet} to learn a representation of points in pillars. The encoded features can then be processed by standard 2D convolutional networks and detection heads. Though the performance of PointPillars is not as satisfactory as other 3D backbones, it and its variants enjoy high efficiency and thus are suitable for industrial applications. 

%\begin{figure*}[t]
%\centering
%\includegraphics[width=0.4\textwidth]{placeholder.png}
%\caption{The general pipeline in BEV Fusion track.}
%\label{fig:roadmap_fusion}
%\end{figure*}

\subsubsection{Discussion}

%As a dense representation, BEV feature map can be seen as normal 2D feature map, in which the center pixel is the location of ego car. Besides using pillar and voxel to transform point cloud data into BEV, previous work also use statistic to describe point cloud in BEV representation. 

% Originally, 
The point cloud data are directly processed by the neural network, as does in~\cite{qi2019votenet,  pan2020pointformer}. The neighborhood relationship among points is calculated in the continuous 3D space. 
This brings extra time consumption and limits the receptive field of the neural network.
Recent works~\cite{yan2018second, zhou2018voxelnet} utilize discrete grids to represent point cloud data; convolutional operations are adopted to extract features. However, converting point cloud data into any form of representation inevitably causes the loss of information. State-of-the-art methods in pre-BEV feature extraction utilize voxels with fine-grained size, preserving most of the 3D information from point cloud data and thus beneficial for 3D detection. As a trade-off, it requires high memory consumption and computational cost. Transforming point cloud data directly into BEV representation avoids complex operation in the 3D space. As the height dimension is compressed, a great loss of information becomes inevitable. The most efficient method is to represent the BEV feature map using statistics, and yet it provides inferior results.
% in low accuracy. 
Pillar-based methods~\cite{lang2019pointpillars} balance performance and cost, and become a popular choice for industrial applications. How to deal with the trade-off between performance and efficiency becomes a vital challenge for LiDAR-based applications.

\subsection{BEV Fusion}\label{sec: bev-fusion}
% written by huijie / yulu

%\subsubsection{General Pipeline}
%The general pipeline of BEV fusion is shown in Fig.~\ref{fig:roadmap_lidar}. 
%It mainly includes camera-to-BEV, lidar-to-BEV, fusion module and task head. Among them, camera-to-BEV and lidar-to-BEV are described in Sec.~\ref{bev-camera} and Sec.~\ref{bev-lidar}. The fusion module generally uses the convolution or attention model. Task head can directly apply various BEV-based heads.

\subsubsection{General Pipeline}

\begin{figure}
\centering
\includegraphics[width=0.48\textwidth]{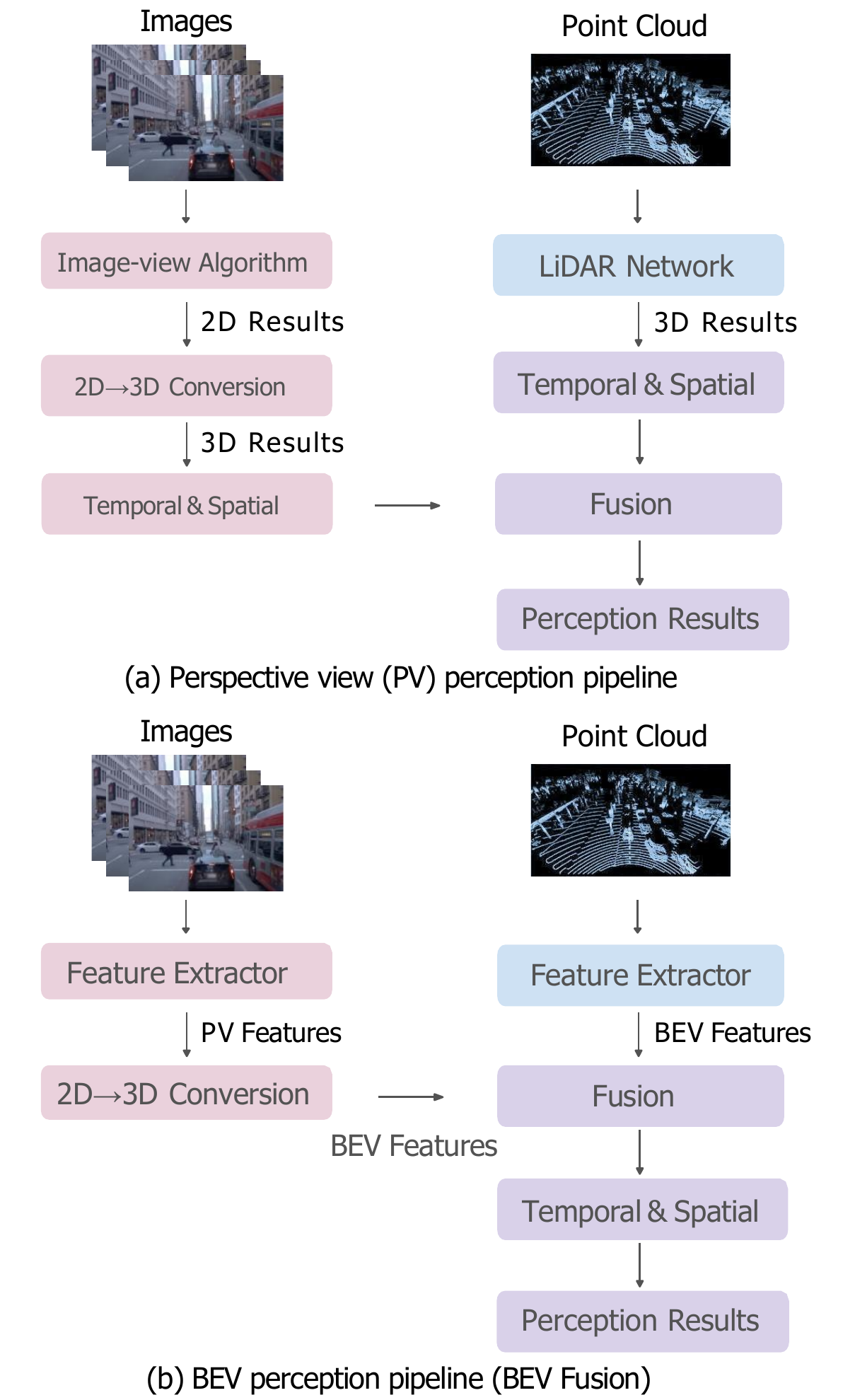}
\caption{
Two typical pipeline designs for BEV fusion algorithms, applicable to both academia and industry.
The main difference lies in 2D to 3D conversion and fusion modules. In the PV perception pipeline (a), results of different algorithm are first transformed into 3D space, then fused using prior or hand-craft rules. The BEV perception pipeline (b) first transforms PV features to BEV, then fuses features to obtain the ultimate predictions, thereby maintaining most original information and avoiding hand-crafted design.
}
\label{fig:industry_pipelines}
\vspace{-0.4cm}
\end{figure}

Inverse perspective mapping (IPM)~\cite{mallot1991ipm} is proposed to map pixels onto the BEV plane using the geometric constraint of the intrinsic and extrinsic matrix of cameras. Despite its inaccuracy due to the flat-ground assumption, it provides the possibility that images and point clouds can be unified in BEV. Lift-splat-shoot (LSS)~\cite{philion2020lift} is the first method to predict depth distribution of image features, introducing neural networks to learn the ill-posed camera-to-lidar transformation problem. Other works~\cite{li2022bevformer, li2022uvtr} develop different method to conduct view transformation. Given the view transformation methods from perspective view to BEV, Fig.~\ref{fig:industry_pipelines}b shows the general pipeline for fusing image and point cloud data. Modal-specific feature extractors are used to extract features in perspective view and BEV separately. After transforming to the representations in BEV, feature maps from different sensors are fused. The temporal and ego-motion information can be introduced in the BEV representation as well.

% \begin{figure}[t]
% \centering
%      \centering
%      \begin{subfigure}[b]{0.4\textwidth}
%          \includegraphics[width=\textwidth]{figures/Industry/industry_PV_pipeline.pdf}
%          \caption{Perspective view (PV) perception pipeline}
%          \label{fig:industry_pv_pipeline}
%      \end{subfigure}
%      \hfill
%      \begin{subfigure}[b]{0.4\textwidth}
%          \includegraphics[width=\textwidth]{figures/Industry/industry_BEV_pipeline.pdf}
%          \caption{BEV perception pipeline (BEV Fusion)}
%          \label{fig:industry_bev_pipeline}
%      \end{subfigure}
% \caption{
% Two typical pipeline designs for BEV fusion algorithms, applicable to both academia and industry.
% % Perception pipeline comparison (PV vs BEV) for industrial applications. 
% The main difference 
% % between these two paradigms 
% lies in 2D to 3D conversion and fusion modules. In the PV perception pipeline (a), results of different algorithm are first transformed into 3D space, then fused using prior or hand-craft rules. The BEV perception pipeline (b) first transforms PV features to BEV, then fuses features to obtain the ultimate predictions, thereby maintaining most original information and avoiding hand-crafted design.
% }
% \label{fig:industry_pipelines}
% \vspace{-0.4cm}
% \end{figure}

\subsubsection{LiDAR-camera Fusion}

%With the development of camera to voxel/bev technology, camera and lidar can be directly fusion in 3D space.

Concurrently, two works with the same name BEVFusion~\cite{liu2022bevfusion, liang2022bevfusion} explore fusion in BEV from different directions. As camera-to-lidar projection~\cite{vora2020pointpainting, wang2021pointaugmenting} throws away the semantic density of camera features, BEVFusion~\cite{liu2022bevfusion} designs an efficient camera-to-BEV transformation method, which efficiently projects camera features into BEV, and then fuses it with lidar BEV features using convolutional layers. BEVFusion~\cite{liang2022bevfusion} regards the BEV fusion as a robustness topic to maintain the stability of the perception system. It encodes camera and lidar features into the same BEV to ensure the independence of camera and lidar streams. This design enables the perception system to maintain stability against sensor failures.

% UTVR different branches are utilized to respectively generate and enhance voxel space for each modality
Apart from BEVFusion~\cite{liu2022bevfusion, liang2022bevfusion}, UVTR~\cite{li2022uvtr} represents different input modalities in modal-specific voxel spaces without height compression to avoid the semantic ambiguity and enable further interactions. Image voxel space is constructed by transforming the image feature of each view to the predefined space with depth distribution generated for each image. Point voxel space is constructed using common 3D convolutional networks. Cross-modality interaction is then conducted between two voxel spaces to enhance modal-specific information.

\subsubsection{Temporal Fusion}
% STINet~\cite{zhang2020stinet}
% HDNET~\cite{billard2018hd}
% MVFuseNet~\cite{daddha2021mvfusenet}

Temporal information plays an important role in inferring the motion state of objects and recognizing occlusions. BEV provides a desirable bridge to connect scene representations in different timestamps, as the central location of the BEV feature map is persistent to ego car. MVFuseNet~\cite{daddha2021mvfusenet} utilizes both BEV and range view for temporal feature extraction. Other works~\cite{hu2021fiery,zhang2022beverse,huang2022bevdet4d} use ego-motion to align the previous BEV features to the current coordinates, and then fuse the current BEV features to obtain the temporal features.
%As the temporal version of BEVDet~\cite{huang2021bevdet}, 
BEVDet4D~\cite{huang2022bevdet4d} fuses the previous feature maps with the current frame using a spatial alignment operation followed by a concatenation of multiple feature maps.  BEVFormer~\cite{li2022bevformer} and UniFormer~\cite{qin2022uniformer} adopt a soft way to fusion temporal information. The attention module is utilized to fuse temporal information from previous BEV feature maps and previous frames, respectively. Concerning the motion of ego car, locations for the attention module to attend among representations of different timestamps are also corrected by the ego-motion information.

% \subsubsection{Multiview}
% VISTA~\cite{deng2022vista}
% MVF~\cite{zhou2020mvf}
% MV3D~\cite{chen2017multi}
% Pillar-DO~\cite{wang2020pillarod}
% CVCNet~\cite{chen2020cvcnet}
% MVFuseNet~\cite{daddha2021mvfusenet}
% Objects do not overlap, and the size of each object is independent of the distance from ego-vehicle in Bev. Compared with Bev, RV could produce compact and dense features by using native representation of LiDAR point clouds. Different views have their own advantages and disadvantages.

\subsubsection{Discussion}

As images are in perspective coordinate and point clouds are in 3D coordinate, spacial alignment between two modalities becomes a vital problem. Though it is easy to project point cloud data onto image coordinates using geometric projection relationships, the sparse nature of point cloud data makes extracting informative features difficult. Inversely, transforming images in perspective view into 3D space would be an ill-posed problem, 
% because of the nature of 
due to the lack of depth information in perspective view. Based on prior knowledge, previous work such as IPM~\cite{mallot1991ipm} and LSS~\cite{philion2020lift} make it possible to transform information in the perspective view into BEV, providing a unified representation for multi-sensor and temporal fusion.

% We believe that with the development of camera to BEV/voxel technology,BEV fusion for multiple sensors beyond lidar and camera remains explorable.
Fusion in the BEV space for lidar and camera data provides satisfactory performance for the 3D detection task. Such a method also maintains the independence of different modalities, which provides the opportunity to build a more robust perception system. For temporal fusion, representations in different timestamps can be directly fused in the BEV space by concerning ego-motion information. Compensation for ego-motion is easy to obtain by monitoring control and motion information, as the BEV coordinate is consistent with the 3D coordinate. With the concern of robustness and consistency, BEV is an ideal representation for multi-sensor and temporal fusion.

\subsection{Industrial Design of BEV Perception}\label{industrial_bev_perception}

Recent years have witnessed trending popularity for BEV perception in the industry. In this section, we describe the architecture design for BEV perception on a system level.

% \gxw{Information is collected from pubic resources}

\begin{figure}[t]
\centering
     \centering
     \begin{subfigure}[b]{0.22\textwidth}
     \centering
         \includegraphics[width=0.95\textwidth]{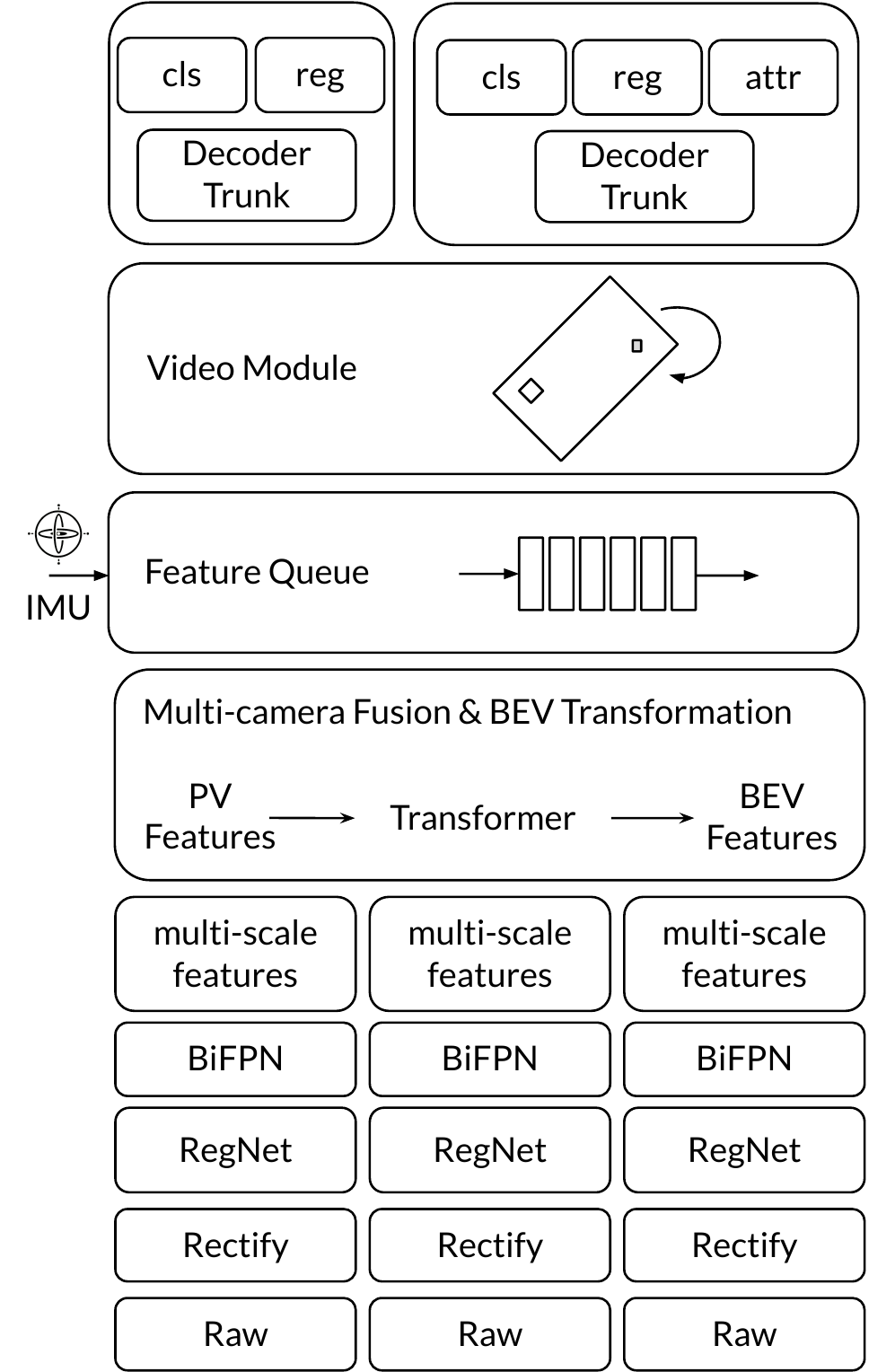}
         \caption{Video predictor with vision input only~\cite{tesla_ai_day}}
         \label{fig:industry_tesla_arch}
     \end{subfigure}
     \hfill
     \begin{subfigure}[b]{0.25\textwidth}
     \centering
         \includegraphics[width=0.95\textwidth]{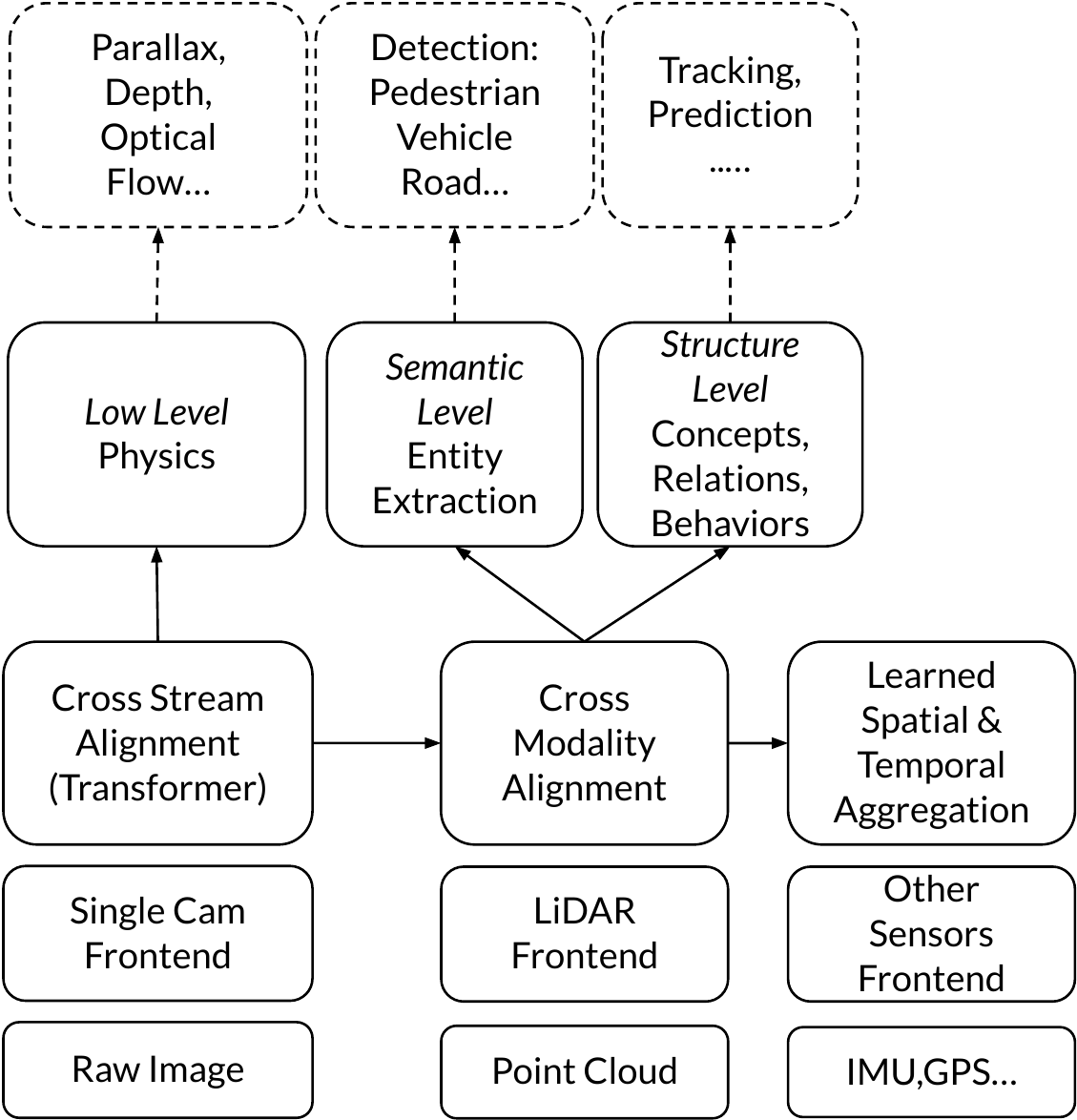}
         \caption{Multi-modality input with various perception tasks~\cite{horizon2022bev}}
         \label{fig:industry_horizon_arch}
     \end{subfigure}
    % \hfill
    %  \begin{subfigure}[b]{0.325\textwidth}
    %  \centering
    %      \includegraphics[width=0.95\textwidth]{figures/Industry/industry_HAOMO_AI.pdf}
    %      \caption{HAOMO~\cite{haomo2022aiday}}
    %      \label{fig:industry_haomo_arch}
    %  \end{subfigure}
    %  \hfill
    %  \begin{subfigure}[b]{0.24\textwidth}
    %      \includegraphics[width=\textwidth]{figures/Industry/industry_phigent.pdf}
    %      \caption{PhiGent Robotics's network architecture}
    %      \label{fig:industry_phigent_arch}
    %  \end{subfigure}
\caption{BEV architecture comparison from industry solutions.
% corporations. 
%
These paradigms follow similar workflow as illustrated in Fig.~\ref{fig:industry_pipelines}b.
% unanimously (input sensors might vary). 
% HY: Does the main context mention this details below?
% First they encode input data with backbone and perform BEV projection via transformer. Then,  BEV features are fused temporally and spatially. At last, they decode BEV feature with different heads. 
There are slight differences in each design. Fig.~\ref{fig:industry_tesla_arch} from Tesla \cite{tesla_ai_day} takes vision as main input and incorporates video module,
% and IMU as input 
while 
Fig.~\ref{fig:industry_horizon_arch}
from Horizon \cite{horizon2022bev} embraces multi-modality to tackle multiple perception tasks.
% to the inputs additionally. 
% Backbone varies in different architectures.
}
\label{fig:industry_arch}
\vspace{-0.4cm}
\end{figure}

Fig.~\ref{fig:industry_pipelines} depicts two typical paradigms for sensor fusion in the industrial applications. 
Prior to BEV perception research, most autonomous driving companies construct the perception system based on perspective view inputs. 
As shown in Fig.~\ref{fig:industry_pipelines}a, 
in the perspective view (PV) pipeline, LiDAR tracks provide 3D results directly, while image-based 3D results are converted from 2D results via geometry prior. Then the predictions from images and LiDAR are fused through hand-crafted approaches.
On the contrary, BEV based methods, as illustrated in Fig.~\ref{fig:industry_pipelines}b, perform feature-level 2D to 3D transformation and integrate features instead of the direct detection outputs from different modalities, leading to less hand-crafted design and more robustness.

Fig.~\ref{fig:industry_arch} summarizes various BEV perception architecture proposed by corporations around the globe.
The detailed model/input options are described in Sec.~\textcolor{red}{D} of Appendix. 
Note that all the information presented in this survey are collected from public resource; comparison and analysis among different plans are based on facts. 
%\textbf{We do NOT advocate, oppose or advertise some plan towards other alternatives.}
%
The BEV fusion architectures in Fig.~\ref{fig:industry_arch}
follow the pipeline as depicted in Fig.~\ref{fig:industry_pipelines}b, consisting of input data, feature extractor, PV to BEV transformation, feature fusion module, temporal \& spatial module and prediction head. We elaborate on each module in details below.

\vspace{-1pt}
\subsubsection{Input Data}
% As mentioned before, 
BEV based perception algorithms support different data modalities, including camera, LiDAR, Radar, IMU and GPS. Camera and LiDAR are the main perception sensors for autonomous driving. Some products use camera only as the input sensor, e.g., Tesla~\cite{tesla_ai_day}, PhiGent~\cite{phigent2022bev}, Mobileye~\cite{mobile2020ces}. 
The others adopt a suite of camera and LiDAR combination, e.g., Horizon~\cite{horizon2022bev}, HAOMO~\cite{haomo2022aiday}. 
Note that IMU and GPS signals are often adopted for sensor fusion plans~\cite{tesla_ai_day, horizon2022bev, haomo2022aiday}, as do in Tesla and Horizon, etc.

\subsubsection{Feature Extractor}

The feature extractor serves as transforming raw data to appropriate feature representations, and this module often consists of backbone and neck. There are different selections for backbone and neck. For instance, ResNet~\cite{he2016deep} in HAOMO~\cite{haomo2022aiday} and  RegNet~\cite{radosavovic2020designing} in Tesla~\cite{tesla_ai_day} can be employed as the image backbone. The neck could be FPN~\cite{Lin2017FeaturePN} from HAOMO~\cite{haomo2022aiday}, BiFPN~\cite{tan2020efficientdet} from Tesla~\cite{tesla_ai_day}, etc. As for the point cloud input, pillar based option from HAOMO~\cite{haomo2022aiday} or voxel based choice from Mobileye~\cite{mobile2020ces} are ideal candidates for the backbone.

\subsubsection{PV to BEV Transformation}

There are mainly four approaches to perform view transformation in industry:
% , namely IPM projection, transformer based transformation and ViDAR~(pseudo LiDAR's name in industry). By far, transformer and ViDAR are most used in industry.
%
(a) \textbf{Fixed IPM}. Based on the flat ground assumption, a fixed transformation can project PV feature to BEV space. Fixed IPM projection handles ground plane well. However, it is sensitive to vehicle jolting and road flatness.
(b) \textbf{Adaptive IPM} utilizes the extrinsic parameters of SDV, which are obtained by some pose estimation approaches, and projects features to BEV accordingly. Although adaptive IPM is robust to vehicle pose, it still hypothesizes on the flat ground assumption.
(c) \textbf{Transformer} based BEV transformation employs dense transformer to project PV feature into BEV space. Such data driven transformation-based methods are widely adopted by Tesla, Horizon, HAOMO.
(d) \textbf{ViDAR}
is first proposed in early 2018 by Waymo and Mobileye in parallel at different venues~\cite{mobile2020ces,waymo2020cvprworkshop}, to indicate the practice of using pixel-level depth to
% ViDAR used pixel level depth to 
project PV feature to BEV space based on camera or vision inputs, resembling the representation form as does in LiDAR. The term ViDAR is equivalent to the concept of pseudo-LiDAR presented in most academic literature. 
Equipped With ViDAR, one can transform images and subsequently features into point cloud directly. Then point cloud based methods can be applied to get BEV features. 
We have seen many ViDAR applications~\cite{tesla_ai_day, mobile2020ces, waymo2020cvprworkshop, zhou2017unsupervised, gordon2019depth} recently, e.g., Tesla, Mobileye, Waymo, Toyota,  etc.
Overall, the options of Transformer and ViDAR are most prevailing in industry.

\subsubsection{Fusion Module}
The alignment among different camera sources has been accomplished in the previous BEV transformation module. In the fusion unit, they step further to aggregate BEV features from camera and LiDAR. By doing so, features from different modalities are ultimately integrated into one unified form.

\subsubsection{Temporal \& Spatial Module}
By stacking BEV features temporally and spatially, a feature queue can be constructed. The temporal stack pushes and pops a feature blob every fixed time, while the spatial stack does it every fixed distance. 
After fusing feature in these stacks into one form, they can obtain a spatial-temporal BEV feature, which is robust to occlusion~\cite{tesla_ai_day, haomo2022aiday}. The aggregation module can be in form of 3D convolution, RNN or Transformer. 
Based on the temporal module and vehicle kinematics, one can maintain a large BEV feature map surrounding ego vehicle and update the feature map locally, as does in the spatial RNN module from Tesla~\cite{tesla_ai_day}.

\subsubsection{Prediction Head}

In BEV perception, the multi-head design is widely adopted. Since BEV feature aggregates information from all sensors, all 3D detection results are decoded from BEV feature space. 
In the meanwhile, PV results are also decoded from the corresponding PV features in some design. The prediction results can be classified into three categories~\cite{horizon2022bev}:
(a) \textbf{Low level results} are related to physics constrains, such as optical flow, depth, etc.
(b) \textbf{Entity level results} include concepts of objects, i.e., vehicle detection, laneline detection, etc.
(c) \textbf{Structure level results} represent relationship between objects, including object tracking, motion prediction, etc.

\section{Empirical Evaluation and Recipe}\label{recipe}
% lidar-only relative written by huijie / tianhao

In this section, we summarize the bag of tricks and the most useful practices to achieve top results on various benchmarks.
This is based on our contest entries to Waymo Open Challenge 2022, namely BEVFormer++ built on top of BEVFormer \cite{li2022bevformer} for camera-only detection and Voxel-SPVCNN derived from SPVCNN \cite{tang2020spvcnn} for LiDAR segmentation.
These practical experiences can serve as a reference to seamlessly integrate into other BEV perception models and evaluate the efficacy. We present practical experiences related to data augmentation, BEV encoder, loss selection in the following content, and place additional practical experiences regarding  detection head design, model ensemble, post-processing in Sec.~\ref{additional_recipes_apdx} of Appendix.
% ~\cite{sun2020scalability} 2022 perception challenge.
%\smch{should be based on some conclusion in section \ref{bev_method}, this is from our waymo challenge XXX

%\lzq{\iffalse}
% This is based on our contest entries
% to 
% % ~\cite{li2022bevformer, tang2020spvcnn} on 
% Waymo Open Challenge 2022, namely BEVFormer++ built on top of BEVFormer \cite{li2022bevformer} for camera-only detection and Voxel-SPVCNN derived from SPVCNN \cite{tang2020spvcnn} for LiDAR segmentation.
% ~\cite{sun2020scalability} 2022 perception challenge.
% \smch{should be based on some conclusion in section \ref{bev_method}, this is from our waymo challenge XXX}
%\lzq{\fi}

 \begin{table*}
   \centering
    \renewcommand\arraystretch{0.8}
   \setlength\tabcolsep{0.1cm}
   \caption{\textbf{BEV camera detection track.} Ablation studies on \texttt{val} set with improvements over BEVFormer~\cite{li2022bevformer}, i.e., BEVFormer++. Some results are only reported as \textit{L1/mAPH} on \textit{car} category with \textit{iou$\geq$0.5}.
      DeD (Deformable DETR head). FrA (FreeAnchor head). CeP (Centerpoint head).
      ConvO (Conv offsets in Temporal Self Attention (TSA)). DE (Deformable view Encoder). CoP (Corner Pooling). DA (2D Auxiliary loss). GR (Global location regression). MS (Multi-Scale), FL (filp), SL (Smoooth L1 Loss), EMA (Exponential Moving Average), SB (Sync BN), 2BS (2x BEV Scale), LLW (Leanable Loss Weight), LS (Label Smoothing), LE (LET-IoU based Assignment) and TTA(Test Time Augmentation). DS (Dataset). The \texttt{mini} dataset contains \nicefrac{1}{5} training data. *denotes the model is trained with 24 epochs, otherwise with 12 epochs.}
      
   \begin{tabular}{c |c c c |cc c c c c c c c c c c c c c |c|c | c c c }
   \toprule 
      ID &DeD & FrA & CeP& ConvO & DE & CoP &  DA & GR & MS & FL & SL & EMA & SB & 2BS & LLW & LS &LE & TTA & Backbone&DS & \makecell{LET- \\ mAPL} & \makecell{LET- \\ mAPH}  & \makecell{L1/ \\ mAPH}  \\
      \midrule
    0&\checkmark & && & & &&&&&&&& &&& &&R101&mini& 34.6 & 46.1 & 25.5   \\
     1&\checkmark & && \checkmark& &&&&&&&&&&&& &&R101&mini& 35.9 & 48.1 & 25.6   \\
     2&\checkmark & && & \checkmark& &&&&&&&&&&& &&R101&mini& 36.1 & 48.1 & 25.9   \\
     3&\checkmark & && & &\checkmark& &&&&&&&&&& &&R101&mini& 35.6 & 46.9 & 26.0   \\
     4&\checkmark & && & &&\checkmark && &&& & & &\checkmark && &&R101&mini& 36.2 & 48.1 & 25.4   \\
     5&\checkmark & && & &&&\checkmark &&&&&&&&& &&R101&mini& 35.4 & 47.2 & 27.2   \\
     6&\checkmark & && & &&&&\checkmark & \checkmark &&&&&&& &&R101&mini&  - &  - & 26.8  \\
     7&\checkmark & && & &&&& &&\checkmark&&&&&& &&R101&mini&  - & - & 27.3 \\
     8&\checkmark & && & &&&& &&&\checkmark &&&&& &&R101&mini&  - & - & 26.2 \\
     9&\checkmark & && & &&&& &&& &\checkmark &&&& &&R101&mini&  - & - & 25.6 \\
     9&\checkmark & && & &&&& &&& & &\checkmark &&& &&R101&mini&  - & - & 25.5 \\
     10&\checkmark & && & && && &&& & & &\checkmark && &&R101&mini&  - & - & 26.5 \\
     11&\checkmark & && & && && &&& & & & &\checkmark & &&R101&mini&  36.0 & 46.7 & - \\
     12&\checkmark & && & && && &&& & & & & &\checkmark &&R101&mini& 34.7 & 44.2 & - \\
     13&   \checkmark &  & & \checkmark& \checkmark&\checkmark&\checkmark&\checkmark& \checkmark &\checkmark &\checkmark &\checkmark &\checkmark &\checkmark & &&& \checkmark &R101&mini& - & - & 37.5 \\

     14*&\checkmark & && & & &&&&&&&&&&& &&SwinL&mini& 40.0&	55.6&	51.9  \\
    %\checkmark &  & & \checkmark& \checkmark&\checkmark&&&\checkmark & \checkmark& & SwinL &mini &  42.4	&57.6	&50.6\\
      15*&   \checkmark &  & & \checkmark& \checkmark&\checkmark&\checkmark&\checkmark& \checkmark &\checkmark &\checkmark &\checkmark &\checkmark &\checkmark & &&& &SwinL &mini &  44.7&60.8	&55.5	\\
    \midrule
     16&& \checkmark && & & &&&&&&&&&&& &&R101& mini& 35.9 & 49.9 & 45.9  \\
     17&& \checkmark && & & &&&&& &&&&&&\checkmark &&R101&mini& 36.3  & 51.1 & 46.6  \\
    \midrule
     18&&  & \checkmark && & &&&& &&&&&&&&&R101 &mini&34.0 & 47.9 & 43.5   \\
    \midrule
       19&   \checkmark &  & & \checkmark& \checkmark&\checkmark&\checkmark&\checkmark& \checkmark &\checkmark &\checkmark &\checkmark &\checkmark &\checkmark & &&&& SwinL &full & 48.4&64.8 	&60.4	\\
      20&&\checkmark   & & \checkmark& \checkmark&\checkmark&\checkmark&\checkmark& \checkmark &\checkmark &\checkmark &\checkmark &\checkmark &\checkmark & &&&&SwinL &full & 47.2&61.2&56.8\\
       21&&\checkmark   & & \checkmark& \checkmark&\checkmark&\checkmark&\checkmark& \checkmark &\checkmark &\checkmark &\checkmark &\checkmark &\checkmark & &&\checkmark & &SwinL &full & 47.6 & 61.4 & 57.0   \\
     22&&&\checkmark & \checkmark& \checkmark&\checkmark&\checkmark&\checkmark& \checkmark &\checkmark &\checkmark &\checkmark &\checkmark &\checkmark & &&&&SwinL&full & 41.9 &54.6  &48.2    \\
        \bottomrule
   \end{tabular}
%   \smallskip
   \label{tab:camera_tricks}
\end{table*}

% 根据不同输入模态，讨论不同的augmentation setting
\subsection{Data Augmentation}

Data augmentation plays a crucial role in enhancing the robustness and generalization capabilities of perception models. By synthetically expanding the diversity of the training dataset, we can improve the model's ability to handle variations in real-world scenarios. 
Certain non-affine transformations are challenging to apply for data augmentation for images or BEV space. For instance, methodologies like copy-paste and Mosaic are problematic due to their potential to cause misalignment between image semantics and 3D spatial information. 
%Common augmentations adopted in recent camera-only methods are color jitter, flip, resize, rotation, crop, and Grid Mask. Meanwhile, it is feasible to perform rotation, flipping, and resizing on the BEV space for any given modality.

\subsubsection{BEV Camera (Camera-only) Detection}
% \noindent\textbf{Camera Input for 3D Object Detection}
% written by hanming
% 常见augmentation， lidar和camera的，整个表？包括augmetation名字和常用参数
Common data augmentations on images for 2D recognition tasks are applicable for the tasks of camera based BEV perception.
In general, we can divide augmentations into static augmentation involving color variation alone and spatial transformation moving pixels around. 
Augmentations based on color variation are directly applicable. For augmentations involving spatial transformation, apart from ground truth transformed accordingly, calibration in camera parameter is also necessary. 
Common augmentations adopted in recent camera-only methods are color jitter, flip, resize, rotation, crop, and Grid Mask. 
%
%Common augmentations adopted in recent work are color jitter, flip, multi-scale resize,  rotation, crop and grid mask.

% Taking the first place solution BEVFormer~\cite{li2022bevformer} in Waymo Open Challenge 2022 as an example,
In BEVFormer++,
color jitter, flip, multi-scale resize and grid mask are employed.
The input image is scaled by a factor between 0.5 and 1.2, flipped by a ratio of 0.5; the maximum $30\%$ of total area is randomly masked with square masks.
Notably, there are two ways for flipping images in BEV perception, which we refer to as image-level flipping and BEV-level flipping.
During image-level flipping, we flip the image and adjust camera intrinsic parameters, while camera extrinsic parameters and GT boxes remain unchanged. This preserves the 3D-to-2D projection relationship. Note that image-level flipping only enhances 2D feature extraction without exerting any influence on the subsequent modules associated with BEV.
During BEV-level flipping, we flip the image and rearrange multi-view images symmetrically, such as switching the front-left camera to the front-right. Overlapped area coherence is preserved. Camera intrinsic parameters stay constant, while extrinsic parameters are adjusted, and GT boxes on the BEV plane are flipped. BEV-level flipping boosts the entire BEV perception model.
BEV-level flipping is adopted in BEVFormer++.
Related ablation study is described in Tab.~\ref{tab:camera_tricks} ID 6 experiment, indicating that data augmentation plays a vital role in improving 3D model's performance.
Since BEVFormer \cite{li2022bevformer} employs sequence input, it ensures that the transformation is consistent for each frame of the sequence after input augmentation.

\subsubsection{LiDAR Segmentaion}
Different from the task of detection, heavy data augmentation can be applied in the task of segmentation, including random rotation, scaling, flipping, and point translation. For random rotation, an angle is picked from the range of $[0, 2\pi)$, rotation is applied to every point on the $x$-$y$ plane. A scale factor is chosen from the range of $[0.9, 1.1]$, and then multiplied on the point cloud coordinate. Random flipping is conducted along $X$ axis, $Y$ axis, or both $X$ and $Y$ axes. For random translation, offsets for each axis are sampled separately from a normal distribution with a mean of $0$ and a standard deviation of $0.1$.

Besides coordinates and reflectance, extra information can be utilized to boost model performance. Painting~\cite{vora2020pointpainting, wang2021pointaugmenting} is a common technique to enhance point cloud data with image information. For unlabeled image data, by projecting point cloud labels onto corresponding images and densifying the sparse annotations, semantic labels on images are obtained from annotated point cloud data. An image model is trained to provide 2D semantic segmentation results. Then, predicted semantic labels are painted as one-hot vectors to point cloud data as additional channels to represent semantic information from images. Besides, temporal information can also be used, as datasets in autonomous driving are usually collected sequentially. Past consecutive frames are concatenated with the current frame. An additional channel is appended to represent the relative time information of different frames. To reduce the number of points, a small voxelization network is applied. Then, voxels treated as points are served as input to our models.

As shown in Tab.~\ref{tab:lidarseg_tricks} ID 1, heavy data augmentation brings an improvement of mIoU of $0.4$. By introducing information from images and previous frames, gains in model performance are $0.5$ and $0.8$ mIoU, respectively (Tab.~\ref{tab:lidarseg_tricks} ID 5 \& 6). It indicates that extra information, especially temporal information, is beneficial for the per-point classification task.

\begin{table*}[t]
   \centering
   \caption{\textbf{BEV LiDAR segmentation track.} Ablation studies on \texttt{val} set with improvements over SPVCNN~\cite{tang2020spvcnn}, i.e., Voxel-SPVCNN. Aug (heavy data augmentation). Arch (adjustments on model architecture). TTA (test-time augmentation). Painting (one-hot painting from image semantic segmentation). Temporal (multi-frame input). V-SPV (Voxel-SPVCNN). Expert (ensemble with more expert models). Post (post-processing techniques including object-level refinement and segmentation with tracking).}
   \label{tab:ablation_tricks}
   
   \begin{tabular}{c|p{1.1cm}<{\centering}p{1.1cm}<{\centering}p{1.1cm}<{\centering}p{1.1cm}<{\centering}p{1.1cm}<{\centering}p{1.1cm}<{\centering}p{1.1cm}<{\centering}p{1.1cm}<{\centering}p{1.1cm}<{\centering}p{1.1cm}<{\centering} | p{1.5cm}<{\centering} }
   \toprule 
   
    ID & Aug & Loss & Arch & TTA & Painting & Temporal & V-SPV & Ensemble & Expert & Post & mIoU  \\
    \midrule
    
    0 &&&&&&&&&&& 67.4 \\
    1 &\checkmark &&&&&&&&&& 67.8 \\
    2 &\checkmark & \checkmark &&&&&&&&& 68.4 \\
    3 &\checkmark & \checkmark & \checkmark &&&&&&&& 69.6 \\
    4 &\checkmark & \checkmark & \checkmark & \checkmark & &&&&&& 71.1 \\
    5 &\checkmark & \checkmark & \checkmark & \checkmark & \checkmark &&&&&& 71.6 \\
    6 &\checkmark & \checkmark & \checkmark & \checkmark & \checkmark & \checkmark &&&&& 72.4 \\
    7 &\checkmark & \checkmark & \checkmark & \checkmark & \checkmark & \checkmark & \checkmark &&&& 73.5 \\
    8 &\checkmark & \checkmark & \checkmark & \checkmark & \checkmark & \checkmark & \checkmark & \checkmark &&& 74.2 \\
    9 &\checkmark & \checkmark & \checkmark & \checkmark & \checkmark & \checkmark & \checkmark & \checkmark & \checkmark && 74.5 \\
    10 &\checkmark & \checkmark & \checkmark & \checkmark & \checkmark & \checkmark & \checkmark & \checkmark & \checkmark &\checkmark & 75.4 \\
    
    \bottomrule
    \end{tabular}
    \label{tab:lidarseg_tricks}
    \vspace{-0.4cm}
\end{table*}

% 选择代表性的bev encoder进行recipe讨论
\subsection{BEV Encoder}
\subsubsection{BEV Camera: BEVFormer++}
% wirtten by hanming
% TBD: 1. will bevformer be introduced in previous chapters or introduced detaily here? 2. a uniform symbol?
% \smch{maybe talk a little bit about backbone?}
BEVFormer++ has multiple encoder layers, each of which follows the conventional structure of transformers~\cite{vaswani2017attention}, except for three tailored designs, namely BEV queries, spatial cross-attention, and temporal self-attention. Specifically, BEV queries are grid-shaped learnable parameters, which is designed to query features in BEV space from multi-camera views via attention mechanisms.
Spatial cross-attention and temporal self-attention are attention layers working with BEV queries, which are used to lookup and aggregate spatial features from multi-camera images as well as temporal features from history BEV feature.

During inference, at timestamp $t$, we feed multi-camera images to the backbone network (e.g., ResNet-101~\cite{he2016deep}), and obtain the features $F_{t}=\{F_{t}^{i}\}_{i=1}^{N_{view}}$ of different camera views, where $F_{t}^{i}$ is the feature of the $i$-th view, $N_{view}$ is the total number of camera views. At the same time, we preserve BEV features $B_{t-1}$ at the prior timestamp $t-1$. 
In each encoder layer, we first use BEV queries $Q$ to query the temporal information from the prior BEV features $B_{t-1}$ via the temporal self-attention.
We then employ BEV queries $Q$ to inquire about the spatial information from  multi-camera features $F_{t}$ via the spatial cross-attention. After the feed-forward network~\cite{vaswani2017attention}, the encoder layer generates the refined BEV features, which is consequently the input of the next encoder layer. 
After six stacking encoder layers, unified BEV features $B_{t}$ at current timestamp $t$ are generated. Taking the BEV features $B_{t}$ as input, the 3D detection head and map segmentation head predict the perception results such as 3D bounding boxes and semantic map.

To improve the feature quality contributing from BEV encoder, three main aspects are to be discussed as follows.
% , including image backbone, view transform and temporal BEV fusion. 

 \textbf{(a) 2D Feature Extractor.} 
Techniques for improving backbone representation quality in 2D perception tasks are most likely to improve presentation quality for BEV tasks as well. For convenience, in the image backbone we adopt feature pyramid that is widely used in most 2D perception tasks. As depicted in Tab.~\ref{tab:camera_tricks}, the structural design of 2D feature extractor, e.g. state-of-the-art image feature extractor~\cite{liu2021swin}, global information interaction~\cite{law2018cornernet}, multi-level feature fusion~\cite{Lin2017FeaturePN,zhu2020deformable} etc. all contribute to better feature representation for BEV perception. Apart from the structural design, auxiliary tasks supervising backbones are also important for the performance of BEV perception, which will be discussed in Sec.~\ref{sec:camera-loss}.

 \textbf{(b) View transformation.} 
The transformation takes in image features and reorganizes them into BEV space. Hyper-parameters, including the image feature's sampling range and frequency, as well as BEV resolution are of vital importance for the performance of BEV perception. The sampling range decides how much of the viewing frustum behind an image will be sampled into BEV space. By default this range is equal to the effective range of LiDAR annotation. When efficiency is of higher priority, the upper z-axis part of the viewing frustum can be compromised since it only contains unimportant information such as sky in most cases. The sampling frequency decides the utility of image features. Higher frequency ensures the model to accurately sample corresponding image features for each BEV location with the cost of higher computation. BEV resolution decides the representation granularity of the BEV feature, where each feature can be accurately traced back to a grid in the world coordinates. High resolution is required for better representation of small scale objects such as traffic lights and pedestrians. 
Related experiments are depicted in Tab.~\ref{tab:camera_tricks} ID 2\&3.
In view transformation, feature extraction operations, e.g. convolution block or Transformer block are also present in many BEV perception networks. Adding better feature extraction sub-networks in the BEV space can also improve the BEV perception performance.

 \textbf{(c) Temporal BEV fusion.} 
Given the structure of BEV feature, temporal fusion in BEV space often leverages ego-car pose information to align temporal BEV features.
% and use cross-attention for feature aggregation. 
However, other agents' movements are not modeled explicitly in this alignment process and
it requires additional learning by the model. As a result, to enhance fusion on features of other moving agents, it is reasonable to increase the perception range of cross-attention as we perform temporal fusion. For instance, we might enlarge the kernel size of attention offset in the deformable attention module or use global attention. Related improvements can be observed in Tab.~\ref{tab:camera_tricks} ID 1.

\subsubsection{BEV LiDAR: Voxel-SPVCNN}
Existing 3D perception models are not ideal to recognize small instances due to the coarse voxelization and aggressive downsampling. SPVCNN~\cite{tang2020spvcnn} utilizes Minkowski U-Net \cite{choy2019minkowski} in the voxel-based branch. To preserve point cloud resolution, an extra point-based branch without downsampling is used. Features of the point-based and voxel-based branches would be propagated to each other at different stages of the network.

We propose Voxel-SPVCNN by making two effective modifications to the original SPVCNN~\cite{tang2020spvcnn}. Compared to simply performing voxelization on the raw input feature, a lightweight 3-layer MLP is applied to extract point features and then the voxelization process is applied. Besides, the input of the point-based branch is replaced by the voxel-as-point branch. The network structure of this branch is still a MLP; but the input is replaced as voxels. 
Voxel-SPVCNN is more efficient, as computation on the point-based branch is greatly reduced, especially in the case where the input is multi-scan point cloud. The change of model architecture brings an improvement of $1.1$ mIoU (see Tab.~\ref{tab:lidarseg_tricks} ID 7).

% \textbf{Different 3D Decoders}
% \smch{}hanming

%Inspired by  Cyl@inder3D~\cite{zhu2020cylinder3d} that a cylindrical partition voxelization eases the problem of imbalance point distribution, an additional cylinder feature extraction branch is fused to the point branch by simple addition. By fusing multiple point cloud representations, a gain of $1.1$ mIoU is achieved as shown in Tab.~\ref{tab:lidarseg_tricks}.

% \subsubsection{Camera-LiDAR: BEVFusion}
% % 这部分是谁负责？
% merged by 4.3 BEV Fusion ?

% XXX

% XXX

% XXX

% \lzq{\iffalse}
% % 讨论3D检测与地图分割，或许还有其它任务。每种任务讨论有代表性的head

% \lzq{\fi}

\subsection{Loss}

\subsubsection{BEV Camera-only Detection}
\label{sec:camera-loss}
%written by hanming
One of the  versatile benefits from BEV feature representation 
% in BEV perception 
is to enable models to be trained with losses proposed in both 2D and 3D object detection. As reported in Tab.~\ref{tab:camera_tricks} ID 16-20, when leveraging different head design, we conclude that the corresponding losses could be transferred with minimum modification, e.g. a tuning in the loss weight.

Apart from the training loss for 3D target, auxiliary loss plays an important role in Camera-only BEV detection. One type of auxiliary loss is to add 2D detection loss on top of 2D feature extractors. Such supervision enhances the localization on 2D image feature, which in turn contributes to 3D representations provided by view transformation in BEV perception. One example of utilizing such auxiliary loss can be observed in Tab.~\ref{tab:camera_tricks} ID 4. Another type of auxiliary loss is depth supervision~\cite{li2022bevdepth}. When utilizing ground truth depth generated from LiDAR systems, the implicit depth estimation capability of BEV perception can be improved to obtain accurate 3D object localization. Both of these auxiliary tasks can be applied during 
% 3D object detection 
training to improve 
% the 3D object detection 
performance. 
As a side note,
% Meanwhile, as a common practice, 
2D detection or depth pretrained backbone are commonly adopted as initialization weights~\cite{wang2022detr3d,li2022bevformer}.

\subsubsection{LiDAR segmentation}
Instead of a conventional cross-entropy loss, Geo loss~\cite{liu2020geo} and Lov\'{a}sz loss~\cite{berman2018lovasz} are utilized to train all models. To have a better boundary of different classes, Geo loss has a strong response to the voxels with rich details. Lov\'{a}sz loss serves as a differentiable intersection-over-union (IoU) loss to mitigate the class imbalance problem. It improves the model performance by $0.6$ mIoU as shown in Tab.~\ref{tab:lidarseg_tricks} ID 2.

\section{Conlusion}
In this survey, we conduct a thorough review on BEV perception in recent years and provide a practical recipe according to our analysis in BEV design pipeline.
Grand challenges and future endeavors could be: (a) how to devise a more accurate depth estimator; (b) how to better align feature representations from multiple sensors in a novel fusion mechanism; (c) how to design a parameter-free network such that the algorithm performance is free to pose variation or sensor location, achieving better generalization ablity across various scenarios; and (d) how to incorporate the successful knowledge from foundation models to facilitate BEV perception.
More detailed discussion could be found in the Appendix.
We hope this survey can benefit the community and be an insightful guidebook for later research in 3D perception.

\section*{Author Contributions}

H. Li, J. Dai and L. Lu lead the project, provide mentorship and allocate resources across task tracks.
H. Li, W. Wang, L. Lu drafted the main outline of the project, worked on a preliminary version of the manuscript, supervised key milestones of the project.
In Waymo Open Challenge 2022, H. Deng, H. Tian, J. Yang and X. Jia served as team lead.
C. Sima, H. Wang, J. Zeng, Z. Li, L. Chen and T. Li are core members during the Challenge, implementing ideas, running experiments, responsible for key outputs of each track.
Y. Li is in charge of the whole data processing team.
Y. Gao implemented the model ensemble part.
For survey writing, J. Zeng wrote the 3D preliminary section. E. Xie wrote and revised the dataset and evaluation part. J. Xie completed the toolbox for bag of tricks.
%
%
%
% Y. Huang, J. Li, T. Li, W. Qin, Y. Bai are team members during Waymo challenge and contribute to experiment implementations.
%
S. Liu and J. Shi oversaw the project from their inception.
D. Lin advised on general picture of the research and gave additional comment on the manuscript.
Y. Qiao was the primary consultant, eliciting key goals and tracking progress. 

\small{
\bibliographystyle{IEEEtran}  
\bibliography{egbib_new}

\begin{thebibliography}{100}
\providecommand{\url}[1]{#1}
\csname url@samestyle\endcsname
\providecommand{\newblock}{\relax}
\providecommand{\bibinfo}[2]{#2}
\providecommand{\BIBentrySTDinterwordspacing}{\spaceskip=0pt\relax}
\providecommand{\BIBentryALTinterwordstretchfactor}{4}
\providecommand{\BIBentryALTinterwordspacing}{\spaceskip=\fontdimen2\font plus
\BIBentryALTinterwordstretchfactor\fontdimen3\font minus \fontdimen4\font\relax}
\providecommand{\BIBforeignlanguage}[2]{{%
\expandafter\ifx\csname l@#1\endcsname\relax
\typeout{** WARNING: IEEEtran.bst: No hyphenation pattern has been}%
\typeout{** loaded for the language `#1'. Using the pattern for}%
\typeout{** the default language instead.}%
\else
\language=\csname l@#1\endcsname
\fi
#2}}
\providecommand{\BIBdecl}{\relax}
\BIBdecl

\bibitem{ren2015faster}
S.~Ren, K.~He, R.~Girshick, and J.~Sun, ``{Faster R-CNN}: Towards real-time object detection with region proposal networks,'' \emph{IEEE Transactions on Pattern Analysis and Machine Intelligence}, vol.~39, no.~6, pp. 1137--1149, 2017.

\bibitem{he2017mask}
K.~He, G.~Gkioxari, P.~Doll{\'a}r, and R.~Girshick, ``{Mask R-CNN},'' in \emph{{CVPR}}, 2017.

\bibitem{xie2022m2bev}
E.~Xie, Z.~Yu, D.~Zhou, J.~Philion, A.~Anandkumar, S.~Fidler, P.~Luo, and J.~M. Alvarez, ``{M$^{2}$BEV}: Multi-camera joint 3d detection and segmentation with unified birds-eye view representation,'' \emph{arXiv preprint arXiv:2204.05088}, 2022.

\bibitem{li2022bevformer}
Z.~Li, W.~Wang, H.~Li, E.~Xie, C.~Sima, T.~Lu, Q.~Yu, and J.~Dai, ``{BEVFormer}: Learning bird's-eye-view representation from multi-camera images via spatiotemporal transformers,'' \emph{arXiv preprint arXiv:2203.17270}, 2022.

\bibitem{liu2022bevfusion}
Z.~Liu, H.~Tang, A.~Amini, X.~Yang, H.~Mao, D.~Rus, and S.~Han, ``{BEVFusion}: Multi-task multi-sensor fusion with unified bird's-eye view representation,'' \emph{arXiv preprint arXiv:2205.13542}, 2022.

\bibitem{tesla_ai_day}
\BIBentryALTinterwordspacing
(2021) {Tesla AI Day}. [Online]. Available: \url{{https://www.youtube.com/watch?v=j0z4FweCy4M}}
\BIBentrySTDinterwordspacing

\bibitem{caesar2020nuscenes}
H.~Caesar, V.~Bankiti, A.~H. Lang, S.~Vora, V.~E. Liong, Q.~Xu, A.~Krishnan, Y.~Pan, G.~Baldan, and O.~Beijbom, ``nuscenes: A multimodal dataset for autonomous driving,'' in \emph{{CVPR}}, 2020.

\bibitem{sun2020scalability}
P.~Sun, H.~Kretzschmar, X.~Dotiwalla, A.~Chouard, V.~Patnaik, P.~Tsui, J.~Guo, Y.~Zhou, Y.~Chai, B.~Caine, V.~Vasudevan, W.~Han, J.~Ngiam, H.~Zhao, A.~Timofeev, S.~Ettinger, M.~Krivokon, A.~Gao, A.~Joshi, Y.~Zhang, J.~Shlens, C.~Zhifeng, and D.~Anguelov, ``Scalability in perception for autonomous driving: Waymo open dataset,'' in \emph{{CVPR}}, 2020.

\bibitem{chen2022_epro-pnp}
H.~Chen, P.~Wang, F.~Wang, W.~Tian, L.~Xiong, and H.~Li, ``{EPro-PnP}: Generalized end-to-end probabilistic perspective-n-points for monocular object pose estimation,'' in \emph{{CVPR}}, 2022.

\bibitem{wang2022dfm}
T.~Wang, J.~Pang, and D.~Lin, ``Monocular 3d object detection with depth from motion,'' \emph{arXiv preprint arXiv:2207.12988}, 2022.

\bibitem{geiger2012we}
A.~Geiger, P.~Lenz, and R.~Urtasun, ``Are we ready for autonomous driving? the kitti vision benchmark suite,'' in \emph{{CVPR}}, 2012.

\bibitem{Argoverse2}
B.~Wilson, W.~Qi, T.~Agarwal, J.~Lambert, J.~Singh, S.~Khandelwal, B.~Pan, R.~Kumar, A.~Hartnett, J.~K. Pontes, D.~Ramanan, P.~Carr, and J.~Hays, ``Argoverse 2: Next generation datasets for self-driving perception and forecasting,'' in \emph{Neural Information Processing Systems Track on Datasets and Benchmarks}, 2021.

\bibitem{waymo2020cvprworkshop}
\BIBentryALTinterwordspacing
(2020) {Drago Anguelov -- Machine Learning for Autonomous Driving at Scale }. [Online]. Available: \url{https://youtu.be/BV4EXwlb3yo}
\BIBentrySTDinterwordspacing

\bibitem{vaswani2017_transformer}
A.~Vaswani, N.~Shazeer, N.~Parmar, J.~Uszkoreit, L.~Jones, A.~N. Gomez, {\L}.~Kaiser, and I.~Polosukhin, ``Attention is all you need,'' in \emph{{NeurIPS}}, 2017.

\bibitem{dosovitskiy2020_vit}
A.~Dosovitskiy, L.~Beyer, A.~Kolesnikov, D.~Weissenborn, X.~Zhai, T.~Unterthiner, M.~Dehghani, M.~Minderer, G.~Heigold, S.~Gelly, J.~Uszkoreit, and N.~Houlsby, ``An image is worth 16x16 words: Transformers for image recognition at scale,'' \emph{arXiv preprint arXiv:2010.11929}, 2020.

\bibitem{han2022survey}
K.~Han, Y.~Wang, H.~Chen, X.~Chen, J.~Guo, Z.~Liu, Y.~Tang, A.~Xiao, C.~Xu, Y.~Xu \emph{et~al.}, ``A survey on vision transformer,'' \emph{IEEE transactions on pattern analysis and machine intelligence}, 2022.

\bibitem{he2022_mae}
K.~He, X.~Chen, S.~Xie, Y.~Li, P.~Doll{\'a}r, and R.~Girshick, ``Masked autoencoders are scalable vision learners,'' in \emph{{CVPR}}, 2022.

\bibitem{radford2021_clip}
A.~Radford, J.~W. Kim, C.~Hallacy, A.~Ramesh, G.~Goh, S.~Agarwal, G.~Sastry, A.~Askell, P.~Mishkin, J.~Clark, G.~Krueger, and I.~Sutskever, ``Learning transferable visual models from natural language supervision,'' in \emph{{ICML}}, 2021.

\bibitem{arnold2019survey}
E.~Arnold, O.~Y. Al-Jarrah, M.~Dianati, S.~Fallah, D.~Oxtoby, and A.~Mouzakitis, ``A survey on 3d object detection methods for autonomous driving applications,'' \emph{IEEE Transactions on Intelligent Transportation Systems}, vol.~20, no.~10, pp. 3782--3795, 2019.

\bibitem{liang2021survey}
W.~Liang, P.~Xu, L.~Guo, H.~Bai, Y.~Zhou, and F.~Chen, ``A survey of 3d object detection,'' \emph{Multimedia Tools and Applications}, vol.~80, no.~19, pp. 29\,617--29\,641, 2021.

\bibitem{qian20223d}
R.~Qian, X.~Lai, and X.~Li, ``3d object detection for autonomous driving: a survey,'' \emph{Pattern Recognition}, p. 108796, 2022.

\bibitem{ma2022vision}
Y.~Ma, T.~Wang, X.~Bai, H.~Yang, Y.~Hou, Y.~Wang, Y.~Qiao, R.~Yang, D.~Manocha, and X.~Zhu, ``Vision-centric bev perception: A survey,'' \emph{arXiv preprint arXiv:2208.02797}, 2022.

\bibitem{mao20223d}
J.~Mao, S.~Shi, X.~Wang, and H.~Li, ``3d object detection for autonomous driving: A review and new outlooks,'' \emph{arXiv preprint arXiv:2206.09474}, 2022.

\bibitem{Argoverse}
M.-F. Chang, J.~W. Lambert, P.~Sangkloy, J.~Singh, S.~Bak, A.~Hartnett, D.~Wang, P.~Carr, S.~Lucey, D.~Ramanan, and J.~Hays, ``Argoverse: 3d tracking and forecasting with rich maps,'' in \emph{{CVPR}}, 2019.

\bibitem{wang2019apolloscape}
X.~Huang, P.~Wang, X.~Cheng, D.~Zhou, Q.~Geng, and R.~Yang, ``The apolloscape open dataset for autonomous driving and its application,'' \emph{IEEE transactions on pattern analysis and machine intelligence}, vol.~42, no.~10, pp. 2702--2719, 2019.

\bibitem{chen2022persformer}
L.~Chen, C.~Sima, Y.~Li, Z.~Zheng, J.~Xu, X.~Geng, H.~Li, C.~He, J.~Shi, Y.~Qiao, and J.~Yan, ``{PersFormer}: 3d lane detection via perspective transformer and the openlane benchmark,'' \emph{arXiv preprint arXiv:2203.11089}, 2022.

\bibitem{yan2022once}
F.~Yan, M.~Nie, X.~Cai, J.~Han, H.~Xu, Z.~Yang, C.~Ye, Y.~Fu, M.~B. Mi, and L.~Zhang, ``Once-3dlanes: Building monocular 3d lane detection,'' in \emph{Proceedings of the IEEE/CVF Conference on Computer Vision and Pattern Recognition}, 2022, pp. 17\,143--17\,152.

\bibitem{houston2020one}
J.~Houston, G.~Zuidhof, L.~Bergamini, Y.~Ye, L.~Chen, A.~Jain, S.~Omari, V.~Iglovikov, and P.~Ondruska, ``One thousand and one hours: Self-driving motion prediction dataset,'' in \emph{Conference on Robot Learning}, 2020.

\bibitem{pham20203d}
Q.-H. Pham, P.~Sevestre, R.~S. Pahwa, H.~Zhan, C.~H. Pang, Y.~Chen, A.~Mustafa, V.~Chandrasekhar, and J.~Lin, ``A* 3d dataset: Towards autonomous driving in challenging environments,'' in \emph{{ICRA}}, 2020.

\bibitem{Patil2019TheHD}
A.~Patil, S.~Malla, H.~Gang, and Y.-T. Chen, ``The h3d dataset for full-surround 3d multi-object detection and tracking in crowded urban scenes,'' in \emph{{ICRA}}, 2019.

\bibitem{behley2019iccv}
J.~Behley, M.~Garbade, A.~Milioto, J.~Quenzel, S.~Behnke, C.~Stachniss, and J.~Gall, ``{SemanticKITTI: A Dataset for Semantic Scene Understanding of LiDAR Sequences},'' in \emph{{ICCV}}, 2019.

\bibitem{geyer2020a2d2}
J.~Geyer, Y.~Kassahun, M.~Mahmudi, X.~Ricou, R.~Durgesh, A.~S. Chung, L.~Hauswald, V.~H. Pham, M.~M{\"u}hlegg, S.~Dorn \emph{et~al.}, ``A2d2: Audi autonomous driving dataset,'' \emph{arXiv preprint arXiv:2004.06320}, 2020.

\bibitem{gahlert2020cityscapes}
N.~G{\"a}hlert, N.~Jourdan, M.~Cordts, U.~Franke, and J.~Denzler, ``Cityscapes 3d: Dataset and benchmark for 9 dof vehicle detection,'' \emph{arXiv preprint arXiv:2006.07864}, 2020.

\bibitem{xiao2021pandaset}
P.~Xiao, Z.~Shao, S.~Hao, Z.~Zhang, X.~Chai, J.~Jiao, Z.~Li, J.~Wu, K.~Sun, K.~Jiang \emph{et~al.}, ``Pandaset: Advanced sensor suite dataset for autonomous driving,'' in \emph{IEEE International Intelligent Transportation Systems Conference}, 2021.

\bibitem{Liao2021ARXIV}
Y.~Liao, J.~Xie, and A.~Geiger, ``{KITTI}-360: A novel dataset and benchmarks for urban scene understanding in 2d and 3d,'' \emph{arXiv preprint arXiv:2109.13410}, 2021.

\bibitem{wang2021cirrus}
Z.~Wang, S.~Ding, Y.~Li, J.~Fenn, S.~Roychowdhury, A.~Wallin, L.~Martin, S.~Ryvola, G.~Sapiro, and Q.~Qiu, ``Cirrus: A long-range bi-pattern lidar dataset,'' in \emph{{ICRA}}, 2021.

\bibitem{mao2021one}
J.~Mao, M.~Niu, C.~Jiang, X.~Liang, Y.~Li, C.~Ye, W.~Zhang, Z.~Li, J.~Yu, C.~Xu \emph{et~al.}, ``One million scenes for autonomous driving: Once dataset,'' \emph{arXiv preprint arXiv:2106.11037}, 2021.

\bibitem{Weng2020_AIODrive}
X.~Weng, Y.~Man, J.~Park, Y.~Yuan, D.~Cheng, M.~O'Toole, and K.~Kitani, ``{All-In-One Drive: A Large-Scale Comprehensive Perception Dataset with High-Density Long-Range Point Clouds},'' \emph{arXiv}, 2021.

\bibitem{wangdeepaccident}
T.~Wang, W.~Ji, S.~Chen, G.~Chongjian, E.~Xie, and P.~Luo, ``{DeepAccident}: A large-scale accident dataset for multi-vehicle autonomous driving,'' 2022.

\bibitem{dosovitskiy2017carla}
A.~Dosovitskiy, G.~Ros, F.~Codevilla, A.~Lopez, and V.~Koltun, ``Carla: An open urban driving simulator,'' in \emph{{CoRL}}, 2017.

\bibitem{song2015sun}
S.~Song, S.~P. Lichtenberg, and J.~Xiao, ``{SUN RGB-D}: A rgb-d scene understanding benchmark suite,'' in \emph{{CVPR}}, 2015.

\bibitem{dai2017scannet}
A.~Dai, A.~X. Chang, M.~Savva, M.~Halber, T.~Funkhouser, and M.~Nie{\ss}ner, ``{ScanNet}: Richly-annotated 3d reconstructions of indoor scenes,'' in \emph{{CVPR}}, 2017.

\bibitem{Roddick2019OrthographicFT}
T.~Roddick, A.~Kendall, and R.~Cipolla, ``Orthographic feature transform for monocular 3d object detection,'' in \emph{British Machine Vision Conference}, 2019.

\bibitem{zhou2018voxelnet}
Y.~Zhou and O.~Tuzel, ``Voxelnet: End-to-end learning for point cloud based 3d object detection,'' in \emph{{CVPR}}, 2018.

\bibitem{lang2019pointpillars}
A.~H. Lang, S.~Vora, H.~Caesar, L.~Zhou, J.~Yang, and O.~Beijbom, ``Pointpillars: Fast encoders for object detection from point clouds,'' in \emph{{CVPR}}, 2019.

\bibitem{reading2021categorical}
C.~Reading, A.~Harakeh, J.~Chae, and S.~L. Waslander, ``Categorical depth distribution network for monocular 3d object detection,'' in \emph{{CVPR}}, 2021.

\bibitem{huang2021bevdet}
J.~Huang, G.~Huang, Z.~Zhu, and D.~Du, ``{BEVDet}: High-performance multi-camera 3d object detection in bird-eye-view,'' \emph{arXiv preprint arXiv:2112.11790}, 2021.

\bibitem{liu2022petr}
Y.~Liu, T.~Wang, X.~Zhang, and J.~Sun, ``Petr: Position embedding transformation for multi-view 3d object detection,'' \emph{arXiv preprint arXiv:2203.05625}, 2022.

\bibitem{li2022bevdepth}
Y.~Li, Z.~Ge, G.~Yu, J.~Yang, Z.~Wang, Y.~Shi, J.~Sun, and Z.~Li, ``{BEVDepth}: Acquisition of reliable depth for multi-view 3d object detection,'' \emph{arXiv preprint arXiv:2206.10092}, 2022.

\bibitem{rukhovich2022imvoxelnet}
D.~Rukhovich, A.~Vorontsova, and A.~Konushin, ``Imvoxelnet: Image to voxels projection for monocular and multi-view general-purpose 3d object detection,'' in \emph{IEEE/CVF Winter Conference on Applications of Computer Vision}, 2022.

\bibitem{jiang2022polarformer}
Y.~Jiang, L.~Zhang, Z.~Miao, X.~Zhu, J.~Gao, W.~Hu, and Y.-G. Jiang, ``Polarformer: Multi-camera 3d object detection with polar transformers,'' \emph{arXiv preprint arXiv:2206.15398}, 2022.

\bibitem{reiher2020sim2real}
L.~Reiher, B.~Lampe, and L.~Eckstein, ``A sim2real deep learning approach for the transformation of images from multiple vehicle-mounted cameras to a semantically segmented image in bird’s eye view,'' in \emph{IEEE 23rd International Conference on Intelligent Transportation Systems}, 2020.

\bibitem{hu2021fiery}
A.~Hu, Z.~Murez, N.~Mohan, S.~Dudas, J.~Hawke, V.~Badrinarayanan, R.~Cipolla, and A.~Kendall, ``{FIERY}: Future instance prediction in bird's-eye view from surround monocular cameras,'' in \emph{{ICCV}}, 2021.

\bibitem{zhou2022cross}
B.~Zhou and P.~Kr{\"a}henb{\"u}hl, ``Cross-view transformers for real-time map-view semantic segmentation,'' in \emph{{CVPR}}, 2022.

\bibitem{li2021hdmapnet}
Q.~Li, Y.~Wang, Y.~Wang, and H.~Zhao, ``Hdmapnet: An online hd map construction and evaluation framework,'' in \emph{{ICRA}}, 2022.

\bibitem{saha2021translating}
A.~Saha, O.~Mendez, C.~Russell, and R.~Bowden, ``Translating images into maps,'' in \emph{{ICRA}}, 2022.

\bibitem{philion2020lift}
J.~Philion and S.~Fidler, ``Lift, splat, shoot: Encoding images from arbitrary camera rigs by implicitly unprojecting to 3d,'' in \emph{{ECCV}}, 2020.

\bibitem{hu2022st}
S.~Hu, L.~Chen, P.~Wu, H.~Li, J.~Yan, and D.~Tao, ``{ST-P3}: End-to-end vision-based autonomous driving via spatial-temporal feature learning,'' \emph{arXiv preprint arXiv:2207.07601}, 2022.

\bibitem{garnett20193d}
N.~Garnett, R.~Cohen, T.~Pe'er, R.~Lahav, and D.~Levi, ``3d-lanenet: end-to-end 3d multiple lane detection,'' in \emph{{ICCV}}, 2019.

\bibitem{can2021structured}
Y.~B. Can, A.~Liniger, D.~P. Paudel, and L.~Van~Gool, ``Structured bird's-eye-view traffic scene understanding from onboard images,'' in \emph{{ICCV}}, 2021.

\bibitem{hung2022let3dap}
W.-C. Hung, H.~Kretzschmar, V.~Casser, J.-J. Hwang, and D.~Anguelov, ``Let-3d-ap: Longitudinal error tolerant 3d average precision for camera-only 3d detection,'' \emph{arXiv preprint arXiv:2206.07705}, 2022.

\bibitem{wang2022probabilistic}
T.~Wang, Z.~Xinge, J.~Pang, and D.~Lin, ``Probabilistic and geometric depth: Detecting objects in perspective,'' in \emph{{CoRL}}, 2022.

\bibitem{zhang2022beverse}
Y.~Zhang, Z.~Zhu, W.~Zheng, J.~Huang, G.~Huang, J.~Zhou, and J.~Lu, ``{BEVerse}: Unified perception and prediction in birds-eye-view for vision-centric autonomous driving,'' \emph{arXiv preprint arXiv:2205.09743}, 2022.

\bibitem{huang2022bevdet4d}
J.~Huang and G.~Huang, ``{BEVDet4D}: Exploit temporal cues in multi-camera 3d object detection,'' \emph{arXiv preprint arXiv:2203.17054}, 2022.

\bibitem{chen2020dsgn}
C.~Yilun, S.~Liu, X.~Shen, and J.~Jia, ``Dsgn: Deep stereo geometry network for 3d object detection,'' \emph{{CVPR}}, 2020.

\bibitem{shi2019pvrcnn}
S.~Shi, C.~Guo, L.~Jiang, Z.~Wang, J.~Shi, X.~Wang, and H.~Li, ``{PV-RCNN}: Point-voxel feature set abstraction for 3d object detection,'' in \emph{{CVPR}}, 2020, pp. 10\,529--10\,538.

\bibitem{yin2021center}
T.~Yin, X.~Zhou, and P.~Krahenbuhl, ``Center-based 3d object detection and tracking,'' in \emph{{CVPR}}, 2021.

\bibitem{fan2022sst}
L.~Fan, Z.~Pang, T.~Zhang, Y.-X. Wang, H.~Zhao, F.~Wang, N.~Wang, and Z.~Zhang, ``Embracing single stride 3d object detector with sparse transformer,'' in \emph{{CVPR}}, 2022.

\bibitem{hu2022afdetv2}
Y.~Hu, Z.~Ding, R.~Ge, W.~Shao, L.~Huang, K.~Li, and Q.~Liu, ``{AFDetV2}: Rethinking the necessity of the second stage for object detection from point clouds,'' \emph{{AAAI}}, 2022.

\bibitem{shi2021pvrcnn++}
S.~Shi, L.~Jiang, J.~Deng, Z.~Wang, C.~Guo, J.~Shi, X.~Wang, and H.~Li, ``{PV-RCNN++}: Point-voxel feature set abstraction with local vector representation for 3d object detection,'' \emph{arXiv preprint arXiv:2102.00463}, 2021.

\bibitem{mao2021pyramid}
J.~Mao, M.~Niu, H.~Bai, X.~Liang, H.~Xu, and C.~Xu, ``{Pyramid R-CNN}: Towards better performance and adaptability for 3d object detection,'' in \emph{{ICCV}}, 2021.

\bibitem{chen2022mppnet}
X.~Chen, S.~Shi, B.~Zhu, K.~C. Cheung, H.~Xu, and H.~Li, ``{MPPNet}: Multi-frame feature intertwining with proxy points for 3d temporal object detection,'' \emph{arXiv preprint arXiv:2205.05979}, 2022.

\bibitem{vora2020pointpainting}
S.~Vora, A.~H. Lang, B.~Helou, and O.~Beijbom, ``{PointPainting}: Sequential fusion for 3d object detection,'' in \emph{{CVPR}}, 2020.

\bibitem{yin2021mvp}
T.~Yin, X.~Zhou, and P.~Kr{\"a}henb{\"u}hl, ``Multimodal virtual point 3d detection,'' \emph{{NeurIPS}}, 2021.

\bibitem{chen2022autoalign}
Z.~Chen, Z.~Li, S.~Zhang, L.~Fang, Q.~Jiang, F.~Zhao, B.~Zhou, and H.~Zhao, ``{AutoAlign}: Pixel-instance feature aggregation for multi-modal 3d object detection,'' \emph{arXiv preprint arXiv:2201.06493}, 2022.

\bibitem{bai2022transfusion}
X.~Bai, Z.~Hu, X.~Zhu, Q.~Huang, Y.~Chen, H.~Fu, and C.-L. Tai, ``{TransFusion}: Robust lidar-camera fusion for 3d object detection with transformers,'' in \emph{{CVPR}}, 2022.

\bibitem{li2022deepfusion}
Y.~Li, A.~W. Yu, T.~Meng, B.~Caine, J.~Ngiam, D.~Peng, J.~Shen, B.~Wu, Y.~Lu, D.~Zhou, Q.~V. Le, A.~Yuille, and M.~Tan, ``Deepfusion: Lidar-camera deep fusion for multi-modal 3d object detection,'' in \emph{{CVPR}}, 2022.

\bibitem{chen2022autoalignv2}
Z.~Chen, Z.~Li, S.~Zhang, L.~Fang, Q.~Jiang, and F.~Zhao, ``{AutoAlignV2}: Deformable feature aggregation for dynamic multi-modal 3d object detection,'' \emph{arXiv preprint arXiv:2207.10316}, 2022.

\bibitem{he2019rethinking}
K.~He, R.~Girshick, and P.~Doll{\'a}r, ``Rethinking imagenet pre-training,'' in \emph{{ICCV}}, 2019.

\bibitem{park2021pseudo}
D.~Park, R.~Ambrus, V.~Guizilini, J.~Li, and A.~Gaidon, ``Is pseudo-lidar needed for monocular 3d object detection?'' in \emph{{ICCV}}, 2021.

\bibitem{wang2021fcos3d}
T.~Wang, X.~Zhu, J.~Pang, and D.~Lin, ``{FCOS3D}: Fully convolutional one-stage monocular 3d object detection,'' in \emph{{ICCV}}, 2021, pp. 913--922.

\bibitem{liu2020smoke}
Z.~Liu, Z.~Wu, and R.~T\'oth, ``{SMOKE}: Single-stage monocular 3d object detection via keypoint estimation,'' \emph{arXiv preprint arXiv:2002.10111}, 2020.

\bibitem{tang2020spvcnn}
H.~Tang, Z.~Liu, S.~Zhao, Y.~Lin, J.~Lin, H.~Wang, and S.~Han, ``Searching efficient 3d architectures with sparse point-voxel convolution,'' in \emph{{ECCV}}, 2020.

\bibitem{yan2018second}
Y.~Yan, Y.~Mao, and B.~Li, ``Second: Sparsely embedded convolutional detection,'' \emph{Sensors}, vol.~18, no.~10, p. 3337, 2018.

\bibitem{wang2022detr3d}
Y.~Wang, V.~C. Guizilini, T.~Zhang, Y.~Wang, H.~Zhao, and J.~Solomon, ``Detr3d: 3d object detection from multi-view images via 3d-to-2d queries,'' in \emph{{CoRL}}, 2022.

\bibitem{wang2019pseudo}
Y.~Wang, W.-L. Chao, D.~Garg, B.~Hariharan, M.~Campbell, and K.~Q. Weinberger, ``Pseudo-lidar from visual depth estimation: Bridging the gap in 3d object detection for autonomous driving,'' in \emph{{CVPR}}, 2019.

\bibitem{you2019pseudo}
Y.~You, Y.~Wang, W.-L. Chao, D.~Garg, G.~Pleiss, B.~Hariharan, M.~Campbell, and K.~Q. Weinberger, ``Pseudo-lidar++: Accurate depth for 3d object detection in autonomous driving,'' \emph{arXiv preprint arXiv:1906.06310}, 2019.

\bibitem{liang2022bevfusion}
T.~Liang, H.~Xie, K.~Yu, Z.~Xia, Z.~Lin, Y.~Wang, T.~Tang, B.~Wang, and Z.~Tang, ``{BEVFusion}: A simple and robust lidar-camera fusion framework,'' \emph{arXiv preprint arXiv:2205.13790}, 2022.

\bibitem{guo2021liga}
X.~Guo, S.~Shi, X.~Wang, and H.~Li, ``{LIGA-Stereo}: Learning lidar geometry aware representations for stereo-based 3d detector,'' \emph{{ICCV}}, 2021.

\bibitem{mallot1991inverse}
H.~A. Mallot, H.~H. B{\"u}lthoff, J.~Little, and S.~Bohrer, ``Inverse perspective mapping simplifies optical flow computation and obstacle detection,'' \emph{Biological cybernetics}, vol.~64, no.~3, pp. 177--185, 1991.

\bibitem{andrew2001multiple}
A.~M. Andrew, ``Multiple view geometry in computer vision,'' \emph{Kybernetes}, 2001.

\bibitem{gong2022gitnet}
S.~Gong, X.~Ye, X.~Tan, J.~Wang, E.~Ding, Y.~Zhou, and X.~Bai, ``{GitNet}: Geometric prior-based transformation for birds-eye-view segmentation,'' \emph{arXiv preprint arXiv:2204.07733}, 2022.

\bibitem{wang2022mvfcos3d++}
T.~Wang, Q.~Lian, C.~Zhu, X.~Zhu, and W.~Zhang, ``{MV-FCOS3D++: Multi-View} camera-only 4d object detection with pretrained monocular backbones,'' \emph{arXiv preprint}, 2022.

\bibitem{pan2020cross}
B.~Pan, J.~Sun, H.~Y.~T. Leung, A.~Andonian, and B.~Zhou, ``Cross-view semantic segmentation for sensing surroundings,'' \emph{IEEE Robotics and Automation Letters}, vol.~5, no.~3, pp. 4867--4873, 2020.

\bibitem{hendy2020fishing}
N.~Hendy, C.~Sloan, F.~Tian, P.~Duan, N.~Charchut, Y.~Xie, C.~Wang, and J.~Philbin, ``Fishing net: Future inference of semantic heatmaps in grids,'' \emph{arXiv preprint arXiv:2006.09917}, 2020.

\bibitem{chitta2021neat}
K.~Chitta, A.~Prakash, and A.~Geiger, ``{NEAT}: Neural attention fields for end-to-end autonomous driving,'' in \emph{{ICCV}}, 2021.

\bibitem{yang2021projecting}
W.~Yang, Q.~Li, W.~Liu, Y.~Yu, Y.~Ma, S.~He, and J.~Pan, ``Projecting your view attentively: Monocular road scene layout estimation via cross-view transformation,'' in \emph{{CVPR}}, 2021.

\bibitem{gosala2022bird}
N.~Gosala and A.~Valada, ``Bird’s-eye-view panoptic segmentation using monocular frontal view images,'' \emph{IEEE Robotics and Automation Letters}, vol.~7, no.~2, pp. 1968--1975, 2022.

\bibitem{vaswani2017attention}
A.~Vaswani, N.~Shazeer, N.~Parmar, J.~Uszkoreit, L.~Jones, A.~N. Gomez, {\L}.~Kaiser, and I.~Polosukhin, ``Attention is all you need,'' in \emph{{NeurIPS}}, 2017.

\bibitem{tian2019fcos}
Z.~Tian, C.~Shen, H.~Chen, and T.~He, ``{FCOS}: Fully convolutional one-stage object detection,'' in \emph{{ICCV}}, 2019.

\bibitem{girshick2015fast}
R.~Girshick, ``{Fast R-CNN},'' in \emph{{ICCV}}, 2015.

\bibitem{xu2018multi}
B.~Xu and Z.~Chen, ``Multi-level fusion based 3d object detection from monocular images,'' in \emph{{CVPR}}, 2018.

\bibitem{ng2020bev}
M.~H. Ng, K.~Radia, J.~Chen, D.~Wang, I.~Gog, and J.~E. Gonzalez, ``{BEV-Seg}: Bird's eye view semantic segmentation using geometry and semantic point cloud,'' \emph{arXiv preprint arXiv:2006.11436}, 2020.

\bibitem{qi2017pointnet}
C.~R. Qi, H.~Su, K.~Mo, and L.~J. Guibas, ``{PointNet}: Deep learning on point sets for 3d classification and segmentation,'' in \emph{{CVPR}}, 2017.

\bibitem{qi2017pointnet++}
C.~R. Qi, L.~Yi, H.~Su, and L.~J. Guibas, ``{PointNet++}: Deep hierarchical feature learning on point sets in a metric space,'' in \emph{{NeurIPS}}, 2017.

\bibitem{he2020sassd}
C.~He, H.~Zeng, J.~Huang, X.-S. Hua, and L.~Zhang, ``Structure aware single-stage 3d object detection from point cloud,'' in \emph{{CVPR}}, 2020.

\bibitem{mao2021votr}
J.~Mao, Y.~Xue, M.~Niu, H.~Bai, J.~Feng, X.~Liang, H.~Xu, and C.~Xu, ``Voxel transformer for 3d object detection,'' in \emph{{ICCV}}, 2021.

\bibitem{deng2020voxelrcnn}
J.~Deng, S.~Shi, P.~Li, W.~Zhou, Y.~Zhang, and H.~Li, ``{Voxel R-CNN}: Towards high performance voxel-based 3d object detection,'' in \emph{{AAAI}}, 2021.

\bibitem{wang2021objectdgcnn}
Y.~Wang and J.~M. Solomon, ``Object dgcnn: 3d object detection using dynamic graphs,'' in \emph{{NeurIPS}}, M.~Ranzato, A.~Beygelzimer, Y.~Dauphin, P.~Liang, and J.~W. Vaughan, Eds., 2021.

\bibitem{chen2017multi}
X.~Chen, H.~Ma, J.~Wan, B.~Li, and T.~Xia, ``Multi-view 3d object detection network for autonomous driving,'' in \emph{{CVPR}}, 2017.

\bibitem{yang2018pixor}
B.~Yang, W.~Luo, and R.~Urtasun, ``{PIXOR}: Real-time 3d object detection from point clouds,'' in \emph{{CVPR}}, 2018.

\bibitem{billard2018hd}
B.~Yang, M.~Liang, and R.~Urtasun, ``Hdnet: Exploiting hd maps for 3d object detection,'' in \emph{{CoRL}}, 2018.

\bibitem{beltran2018birdnet}
J.~Beltrán, C.~Guindel, F.~M. Moreno, D.~Cruzado, F.~García, and A.~De~La~Escalera, ``{BirdNet}: A 3d object detection framework from lidar information,'' in \emph{International Conference on Intelligent Transportation Systems}, 2018.

\bibitem{zeng2018rt3d}
Y.~Zeng, Y.~Hu, S.~Liu, J.~Ye, Y.~Han, X.~Li, and N.~Sun, ``Rt3d: Real-time 3-d vehicle detection in lidar point cloud for autonomous driving,'' \emph{IEEE Robotics and Automation Letters}, vol.~3, no.~4, pp. 3434--3440, 2018.

\bibitem{ali2018yolo3d}
W.~Ali, S.~Abdelkarim, M.~Zidan, M.~Zahran, and A.~El~Sallab, ``Yolo3d: End-to-end real-time 3d oriented object bounding box detection from lidar point cloud,'' in \emph{European Conference on Computer Vision Workshops}, 2018.

\bibitem{simony2018complex}
M.~Simony, S.~Milzy, K.~Amendey, and H.-M. Gross, ``{Complex-YOLO}: An euler-region-proposal for real-time 3d object detection on point clouds,'' in \emph{European Conference on Computer Vision Workshops}, 2018.

\bibitem{he2016deep}
K.~He, X.~Zhang, S.~Ren, and J.~Sun, ``Deep residual learning for image recognition,'' in \emph{{CVPR}}, 2016.

\bibitem{graham2014sparse}
B.~Graham, ``Spatially-sparse convolutional neural networks,'' \emph{arXiv preprint arXiv:1409.6070}, 2014.

\bibitem{choy2019minkowski}
C.~Choy, J.~Gwak, and S.~Savarese, ``4d spatio-temporal convnets: Minkowski convolutional neural networks,'' in \emph{{CVPR}}, 2019.

\bibitem{qi2019votenet}
C.~R. Qi, O.~Litany, K.~He, and L.~J. Guibas, ``Deep hough voting for 3d object detection in point clouds,'' in \emph{{ICCV}}, 2019.

\bibitem{pan2020pointformer}
X.~Pan, Z.~Xia, S.~Song, L.~E. Li, and G.~Huang, ``3d object detection with pointformer,'' in \emph{{CVPR}}, 2021.

\bibitem{mallot1991ipm}
H.~Mallot, H.~B{\"u}lthoff, J.~Little, and S.~Bohrer, ``Inverse perspective mapping simplifies optical flow computation and obstacle detection,'' \emph{Biological Cybernetics}, vol.~64, no.~3, pp. 177--185, 1991.

\bibitem{li2022uvtr}
Y.~Li, Y.~Chen, X.~Qi, Z.~Li, J.~Sun, and J.~Jia, ``Unifying voxel-based representation with transformer for 3d object detection,'' \emph{arXiv preprint arXiv:2206.00630}, 2022.

\bibitem{wang2021pointaugmenting}
C.~Wang, C.~Ma, M.~Zhu, and X.~Yang, ``{PointAugmenting}: Cross-modal augmentation for 3d object detection,'' in \emph{{CVPR}}, 2021.

\bibitem{daddha2021mvfusenet}
A.~Laddha, S.~Gautam, S.~Palombo, S.~Pandey, and C.~Vallespi-Gonzalez, ``{MVFuseNet}: Improving end-to-end object detection and motion forecasting through multi-view fusion of lidar data,'' in \emph{CVPR Workshops}, 2021.

\bibitem{qin2022uniformer}
Z.~Qin, J.~Chen, C.~Chen, X.~Chen, and X.~Li, ``{UniFormer}: Unified multi-view fusion transformer for spatial-temporal representation in bird's-eye-view,'' \emph{arXiv preprint arXiv:2207.08536}, 2022.

\bibitem{horizon2022bev}
\BIBentryALTinterwordspacing
(2022) {Horizon Algorithms }. [Online]. Available: \url{https://en.horizon.ai/products/applications-algorithms/}
\BIBentrySTDinterwordspacing

\bibitem{phigent2022bev}
\BIBentryALTinterwordspacing
(2022) {PhiGent: Technical Roadmap }. [Online]. Available: \url{https://43.132.128.84/coreTechnology}
\BIBentrySTDinterwordspacing

\bibitem{mobile2020ces}
\BIBentryALTinterwordspacing
(2020) {CES 2020 by Mobileye }. [Online]. Available: \url{https://youtu.be/HPWGFzqd7pI}
\BIBentrySTDinterwordspacing

\bibitem{haomo2022aiday}
\BIBentryALTinterwordspacing
(2022) {HAOMO AI DAY }. [Online]. Available: \url{https://www.bilibili.com/video/BV1Wr4y1H7W7}
\BIBentrySTDinterwordspacing

\bibitem{radosavovic2020designing}
I.~Radosavovic, R.~P. Kosaraju, R.~Girshick, K.~He, and P.~Doll{\'a}r, ``Designing network design spaces,'' in \emph{{CVPR}}, 2020.

\bibitem{Lin2017FeaturePN}
T.-Y. Lin, P.~Doll{\'a}r, R.~B. Girshick, K.~He, B.~Hariharan, and S.~J. Belongie, ``Feature pyramid networks for object detection,'' \emph{{CVPR}}, 2017.

\bibitem{tan2020efficientdet}
M.~Tan, R.~Pang, and Q.~V. Le, ``{EfficientDet}: Scalable and efficient object detection,'' in \emph{{CVPR}}, 2020.

\bibitem{zhou2017unsupervised}
T.~Zhou, M.~Brown, N.~Snavely, and D.~G. Lowe, ``Unsupervised learning of depth and ego-motion from video,'' in \emph{{CVPR}}, 2017.

\bibitem{gordon2019depth}
A.~Gordon, H.~Li, R.~Jonschkowski, and A.~Angelova, ``Depth from videos in the wild: Unsupervised monocular depth learning from unknown cameras,'' in \emph{{ICCV}}, 2019.

\bibitem{liu2021swin}
Z.~Liu, Y.~Lin, Y.~Cao, H.~Hu, Y.~Wei, Z.~Zhang, S.~Lin, and B.~Guo, ``Swin transformer: Hierarchical vision transformer using shifted windows,'' in \emph{{ICCV}}, 2021.

\bibitem{law2018cornernet}
H.~Law and J.~Deng, ``Cornernet: Detecting objects as paired keypoints,'' in \emph{{ECCV}}, 2018.

\bibitem{zhu2020deformable}
X.~Zhu, W.~Su, L.~Lu, B.~Li, X.~Wang, and J.~Dai, ``Deformable detr: Deformable transformers for end-to-end object detection,'' in \emph{{ICLR}}, 2020.

\bibitem{liu2020geo}
J.~Li, Y.~Liu, X.~Yuan, C.~Zhao, R.~Siegwart, I.~Reid, and C.~Cadena, ``Depth based semantic scene completion with position importance aware loss,'' \emph{IEEE Robotics and Automation Letters}, vol.~5, no.~1, pp. 219--226, 2019.

\bibitem{berman2018lovasz}
M.~Berman, A.~R. Triki, and M.~B. Blaschko, ``The lovász-softmax loss: A tractable surrogate for the optimization of the intersection-over-union measure in neural networks,'' in \emph{{CVPR}}, 2018.

\bibitem{Chen_2016_CVPR}
X.~Chen, K.~Kundu, Z.~Zhang, H.~Ma, S.~Fidler, and R.~Urtasun, ``Monocular 3d object detection for autonomous driving,'' in \emph{{CVPR}}, 2016.

\bibitem{mousavian20173d}
A.~Mousavian, D.~Anguelov, J.~Flynn, and J.~Kosecka, ``3d bounding box estimation using deep learning and geometry,'' in \emph{{CVPR}}, 2017.

\bibitem{kundu20183d}
A.~Kundu, Y.~Li, and J.~M. Rehg, ``{3D-RCNN}: Instance-level 3d object reconstruction via render-and-compare,'' in \emph{{CVPR}}, 2018.

\bibitem{zhou2019objects}
X.~Zhou, D.~Wang, and P.~Kr{\"a}henb{\"u}hl, ``Objects as points,'' \emph{arXiv preprint arXiv:1904.07850}, 2019.

\bibitem{brazil2019m3d}
G.~Brazil and X.~Liu, ``{M3D-RPN}: Monocular 3d region proposal network for object detection,'' in \emph{{ICCV}}, 2019.

\bibitem{ku2019monocular}
J.~Ku, A.~D. Pon, and S.~L. Waslander, ``Monocular 3d object detection leveraging accurate proposals and shape reconstruction,'' in \emph{{CVPR}}, 2019.

\bibitem{manhardt2019roi}
F.~Manhardt, W.~Kehl, and A.~Gaidon, ``{ROI-10D}: Monocular lifting of 2d detection to 6d pose and metric shape,'' in \emph{{CVPR}}, 2019.

\bibitem{shi2019pointrcnn}
S.~Shi, X.~Wang, and H.~Li, ``{PointRCNN}: 3d object proposal generation and detection from point cloud,'' in \emph{{CVPR}}, 2019.

\bibitem{shi2019parta2}
S.~Shi, Z.~Wang, J.~Shi, X.~Wang, and H.~Li, ``From points to parts: 3d object detection from point cloud with part-aware and part-aggregation network,'' \emph{IEEE transactions on pattern analysis and machine intelligence}, vol.~43, no.~8, pp. 2647--2664, 2020.

\bibitem{zhang2020h3dnet}
Z.~Zhang, B.~Sun, H.~Yang, and Q.~Huang, ``{H3DNet}: 3d object detection using hybrid geometric primitives,'' in \emph{{ECCV}}, 2020.

\bibitem{cheng2021back}
B.~Cheng, L.~Sheng, S.~Shi, M.~Yang, and D.~Xu, ``Back-tracing representative points for voting-based 3d object detection in point clouds,'' in \emph{{CVPR}}, 2021.

\bibitem{liu2021groupfree}
Z.~Liu, Z.~Zhang, Y.~Cao, H.~Hu, and X.~Tong, ``Group-free 3d object detection via transformers,'' in \emph{{ICCV}}, 2021.

\bibitem{wang2022rbgnet}
H.~Wang, S.~Shi, Z.~Yang, R.~Fang, Q.~Qian, H.~Li, B.~Schiele, and L.~Wang, ``{RBGNet}: Ray-based grouping for 3d object detection,'' in \emph{{CVPR}}, 2022.

\bibitem{yang20203dssd}
Z.~Yang, Y.~Sun, S.~Liu, and J.~Jia, ``{3DSSD}: Point-based 3d single stage object detector,'' in \emph{{CVPR}}, 2020.

\bibitem{xu2018spidercnn}
Y.~Xu, T.~Fan, M.~Xu, L.~Zeng, and Y.~Qiao, ``{SpiderCNN}: Deep learning on point sets with parameterized convolutional filters,'' in \emph{{ECCV}}, 2018.

\bibitem{wang2018dgcnn}
Y.~Wang, Y.~Sun, Z.~Liu, S.~E. Sarma, M.~M. Bronstein, and J.~M. Solomon, ``Dynamic graph cnn for learning on point clouds,'' \emph{ACM Transactions On Graphics}, vol.~38, no.~5, pp. 1--12, 2019.

\bibitem{li2018pointcnn}
Y.~Li, R.~Bu, M.~Sun, W.~Wu, X.~Di, and B.~Chen, ``{PointCNN}: Convolution on x-transformed points,'' in \emph{{NeurIPS}}, 2018.

\bibitem{thomas2019kpconv}
H.~Thomas, C.~R. Qi, J.-E. Deschaud, B.~Marcotegui, F.~Goulette, and L.~J. Guibas, ``{KPConv}: Flexible and deformable convolution for point clouds,'' in \emph{{ICCV}}, 2019.

\bibitem{zhao2021pointtransformer}
H.~Zhao, L.~Jiang, J.~Jia, P.~H. Torr, and V.~Koltun, ``Point transformer,'' in \emph{{ICCV}}, 2021.

\bibitem{hu2020RandLA}
Q.~Hu, B.~Yang, L.~Xie, S.~Rosa, Y.~Guo, Z.~Wang, A.~Trigoni, and A.~Markham, ``{RandLA-Net}: Efficient semantic segmentation of large-scale point clouds,'' \emph{{CVPR}}, 2020.

\bibitem{zhang2020polarnet}
Y.~Zhang, Z.~Zhou, P.~David, X.~Yue, Z.~Xi, B.~Gong, and H.~Foroosh, ``{PolarNet}: An improved grid representation for online lidar point clouds semantic segmentation,'' in \emph{{CVPR}}, 2020.

\bibitem{zhu2020cylinder3d}
X.~Zhu, H.~Zhou, T.~Wang, F.~Hong, Y.~Ma, W.~Li, H.~Li, and D.~Lin, ``Cylindrical and asymmetrical 3d convolution networks for lidar segmentation,'' in \emph{{CVPR}}, 2021.

\bibitem{cheng2021af2}
R.~Cheng, R.~Razani, E.~Taghavi, E.~Li, and B.~Liu, ``{(AF)2-S3Net}: Attentive feature fusion with adaptive feature selection for sparse semantic segmentation network,'' in \emph{{CVPR}}, 2021.

\bibitem{gerdzhev2021tornado}
M.~Gerdzhev, R.~Razani, E.~Taghavi, and L.~Bingbing, ``{TORNADO-Net}: multiview total variation semantic segmentation with diamond inception module,'' in \emph{{ICRA}}, 2021.

\bibitem{liong2020amvnet}
V.~E. Liong, T.~N.~T. Nguyen, S.~Widjaja, D.~Sharma, and Z.~J. Chong, ``{AMVNet}: Assertion-based multi-view fusion network for lidar semantic segmentation,'' \emph{arXiv preprint arXiv:2012.04934}, 2020.

\bibitem{ye2021drinet}
M.~Ye, S.~Xu, T.~Cao, and Q.~Chen, ``{DRINet}: A dual-representation iterative learning network for point cloud segmentation,'' in \emph{{ICCV}}, 2021.

\bibitem{ye2021drinet++}
M.~Ye, R.~Wan, S.~Xu, T.~Cao, and Q.~Chen, ``Drinet++: Efficient voxel-as-point point cloud segmentation,'' \emph{arXiv preprint arXiv:2111.08318}, 2021.

\bibitem{xu2021rpv}
J.~Xu, R.~Zhang, J.~Dou, Y.~Zhu, J.~Sun, and S.~Pu, ``{RPVNet}: A deep and efficient range-point-voxel fusion network for lidar point cloud segmentation,'' in \emph{{ICCV}}, 2021.

\bibitem{genova20212d3dnet}
K.~Genova, X.~Yin, A.~Kundu, C.~Pantofaru, F.~Cole, A.~Sud, B.~Brewington, B.~Shucker, and T.~Funkhouser, ``Learning 3d semantic segmentation with only 2d image supervision,'' in \emph{International Conference on 3D Vision}, 2021.

\bibitem{yan20222dpass}
X.~Yan, J.~Gao, C.~Zheng, C.~Zheng, R.~Zhang, S.~Cui, and Z.~Li, ``{2DPASS}: 2d priors assisted semantic segmentation on lidar point clouds,'' \emph{arXiv preprint arXiv:2207.04397}, 2022.

\bibitem{mvxnet2019}
V.~A. Sindagi, Y.~Zhou, and O.~Tuzel, ``{MVX-Net}: Multimodal voxelnet for 3d object detection,'' in \emph{{ICRA}}, 2019.

\bibitem{liang2019mmf}
M.~Liang, B.~Yang, Y.~Chen, R.~Hu, and R.~Urtasun, ``Multi-task multi-sensor fusion for 3d object detection,'' in \emph{{CVPR}}, 2019.

\bibitem{liang2019contfuse}
M.~Liang, B.~Yang, S.~Wang, and R.~Urtasun, ``Deep continuous fusion for multi-sensor 3d object detection,'' in \emph{{ECCV}}, 2018.

\bibitem{nabati2021centerfusion}
R.~Nabati and H.~Qi, ``{CenterFusion}: Center-based radar and camera fusion for 3d object detection,'' in \emph{Proceedings of the IEEE/CVF Winter Conference on Applications of Computer Vision}, 2021.

\bibitem{chen2022futr3d}
X.~Chen, T.~Zhang, Y.~Wang, Y.~Wang, and H.~Zhao, ``Futr3d: A unified sensor fusion framework for 3d detection,'' \emph{arXiv preprint arXiv:2203.10642}, 2022.

\bibitem{yang2022deepinteraction}
Z.~Yang, J.~Chen, Z.~Miao, W.~Li, X.~Zhu, and L.~Zhang, ``{DeepInteraction}: 3d object detection via modality interaction,'' \emph{arXiv preprint arXiv:2208.11112}, 2020.

\bibitem{qi2018frustumpointnets}
C.~R. Qi, W.~Liu, C.~Wu, H.~Su, and L.~J. Guibas, ``Frustum pointnets for 3d object detection from rgb-d data,'' in \emph{{CVPR}}, 2018.

\bibitem{ku2018avod}
J.~Ku, M.~Mozifian, J.~Lee, A.~Harakeh, and S.~L. Waslander, ``Joint 3d proposal generation and object detection from view aggregation,'' in \emph{{IROS}}, 2018.

\bibitem{pang2020clocs}
S.~Pang, D.~Morris, and H.~Radha, ``{CLOCs}: Camera-lidar object candidates fusion for 3d object detection,'' in \emph{{IROS}}, 2020.

\bibitem{philion2020learning}
J.~Philion, A.~Kar, and S.~Fidler, ``Learning to evaluate perception models using planner-centric metrics,'' in \emph{{CVPR}}, 2020.

\bibitem{phigent2022vidar}
\BIBentryALTinterwordspacing
(2022) {ViDAR -- Forward Visual Perception for Advanced Self-Driving }. [Online]. Available: \url{https://43.132.128.84/productions/visualRadar}
\BIBentrySTDinterwordspacing

\bibitem{carion2020end}
N.~Carion, F.~Massa, G.~Synnaeve, N.~Usunier, A.~Kirillov, and S.~Zagoruyko, ``End-to-end object detection with transformers,'' in \emph{{ECCV}}, 2020.

\bibitem{zhang2019freeanchor}
X.~Zhang, F.~Wan, C.~Liu, R.~Ji, and Q.~Ye, ``Freeanchor: Learning to match anchors for visual object detection,'' in \emph{{NeurIPS}}, 2019.

\bibitem{li2022dn}
F.~Li, H.~Zhang, S.~Liu, J.~Guo, L.~M. Ni, and L.~Zhang, ``Dn-detr: Accelerate detr training by introducing query denoising,'' in \emph{{CVPR}}, 2022.

\bibitem{meng2021conditional}
D.~Meng, X.~Chen, Z.~Fan, G.~Zeng, H.~Li, Y.~Yuan, L.~Sun, and J.~Wang, ``Conditional detr for fast training convergence,'' in \emph{{ICCV}}, 2021.

\bibitem{solovyev2019weighted}
R.~Solovyev, W.~Wang, and T.~Gabruseva, ``Weighted boxes fusion: Ensembling boxes from different object detection models,'' \emph{Image and Vision Computing}, vol. 107, pp. 104--117, 2021.

\bibitem{liu20201st}
Y.~Liu, G.~Song, Y.~Zang, Y.~Gao, E.~Xie, J.~Yan, C.~C. Loy, and X.~Wang, ``1st place solutions for openimage2019--object detection and instance segmentation,'' \emph{arXiv preprint arXiv:2003.07557}, 2020.

\bibitem{wang2020solov2}
X.~Wang, R.~Zhang, T.~Kong, L.~Li, and C.~Shen, ``{SOLOv2}: Dynamic and fast instance segmentation,'' \emph{{NeurIPS}}, 2020.

\bibitem{NNI}
\BIBentryALTinterwordspacing
``Neural network intelligence.'' [Online]. Available: \url{https://github.com/microsoft/nni}
\BIBentrySTDinterwordspacing

\bibitem{Zhou2021_parameterfree}
Y.~Zhou, Y.~He, H.~Zhu, C.~Wang, H.~Li, and Q.~Jiang, ``Monocular 3d object detection: An extrinsic parameter free approach,'' in \emph{{CVPR}}, 2021.

\bibitem{bommasani2021_foundationmodel}
R.~Bommasani, D.~A. Hudson, E.~Adeli, R.~Altman, S.~Arora, S.~von Arx, M.~S. Bernstein, J.~Bohg, A.~Bosselut, E.~Brunskill, E.~Brynjolfsson, S.~Buch, D.~Card, R.~Castellon, N.~Chatterji, A.~Chen, and et~al., ``On the opportunities and risks of foundation models,'' \emph{arXiv preprint arXiv:2108.07258}, 2021.

\bibitem{brown2020_gpt}
T.~Brown, B.~Mann, N.~Ryder, M.~Subbiah, J.~D. Kaplan, P.~Dhariwal, A.~Neelakantan, P.~Shyam, G.~Sastry, A.~Askell, S.~Agarwal, A.~Herbert-Voss, G.~Krueger, T.~Henighan, R.~Child, A.~Ramesh, D.~Ziegler, J.~Wu, C.~Winter, C.~Hesse, M.~Chen, E.~Sigler, M.~Litwin, S.~Gray, B.~Chess, J.~Clark, C.~Berner, S.~McCandlish, A.~Radford, I.~Sutskever, and D.~Amodei, ``Language models are few-shot learners,'' in \emph{{NeurIPS}}, 2020.

\bibitem{wang2022ofa}
P.~Wang, A.~Yang, R.~Men, J.~Lin, S.~Bai, Z.~Li, J.~Ma, C.~Zhou, J.~Zhou, and H.~Yang, ``{OFA}: Unifying architectures, tasks, and modalities through a simple sequence-to-sequence learning framework,'' in \emph{{ICML}}, 2022.

\bibitem{zhu2022_uniperceiver-moe}
J.~Zhu, X.~Zhu, W.~Wang, X.~Wang, H.~Li, X.~Wang, and J.~Dai, ``{Uni-Perceiver-MoE}: Learning sparse generalist models with conditional moes,'' \emph{arXiv preprint arXiv:2206.04674}, 2022.

\bibitem{scott2022_gato}
S.~Reed, K.~Zolna, E.~Parisotto, S.~G. Colmenarejo, A.~Novikov, G.~Barth-Maron, M.~Gimenez, Y.~Sulsky, J.~Kay, J.~T. Springenberg, T.~Eccles, J.~Bruce, A.~Razavi, A.~Edwards, N.~Heess, Y.~Chen, R.~Hadsell, O.~Vinyals, M.~Bordbar, and N.~de~Freitas, ``A generalist agent,'' \emph{arXiv preprint arXiv:2205.06175}, 2022.

\end{thebibliography}
}
\vskip -2\baselineskip plus -1fil
\begin{IEEEbiographynophoto}
% [{\includegraphics[width=1in,height=1.25in,clip,keepaspectratio]{figures/bio/Hongyang.jpg}}]
{Hongyang Li} (S'13-M'20-SM'23) received PhD  from The Chinese University of Hong Kong in 2019.
He is currently a Research Scientist at OpenDriveLab, Shanghai AI Lab. His expertise focuses on perception and cognition, end-to-end autonomous driving and foundation model. He is also affiliated with 
Shanghai Jiao Tong University. He serves as Area Chair for top-tiered conferences multiple times, including CVPR, NeurIPS. He won as PI the CVPR 2023 Best Paper Award, and proposed BEVFormer.
% as a
% prominent 3D detection baseline.
\end{IEEEbiographynophoto}
\vskip -2\baselineskip plus -1fil
\begin{IEEEbiographynophoto}
% [{\includegraphics[width=1in,height=1.25in,clip,keepaspectratio]{figures/bio/chonghao.jpg}}]
{Chonghao Sima} received the B.E. in Computer Science from Huazhong University of Science and Technology in 2019. 
He is currently a Researcher at OpenDriveLab, Shanghai AI Lab and a Ph.D. candidate at The University of Hong Kong.
 % , Shanghai, China. 
 His research interests include autonomous driving, computer vision and foundation model.
 He was an Outstanding Reviewer at CVPR 2023. He is the core author of a few popular
 % autonomous driving 
 work, including BEVFormer, OccNet, etc. He won the First Place at Waymo Challenge 2022. 
 % He served as the organization team on a few international workshops.
\end{IEEEbiographynophoto}
\vskip -2\baselineskip plus -1fil
\begin{IEEEbiographynophoto}
%[{\includegraphics[width=1in,height=1.25in,clip,keepaspectratio]{figures/bio/jifeng.jpeg}}]
{Jifeng Dai} is an Associate Professor at Department of Electronic Engineering of Tsinghua University. His current research focus is on deep learning for high-level vision.
Prior to that, he was an Executive Research Director at SenseTime Research, headed by Professor Xiaogang Wang, between 2019 and 2022. He was a Principle Research Manager in Visual Computing Group at Microsoft Research Asia (MSRA) between 2014 and 2019, headed by Dr. Jian Sun.
He got Ph.D. degree from the Department of Automation, Tsinghua University in 2014, under the supervison of Professor Jie Zhou. 
%He served as Area Chair for CVPR, ICCV, NeurIPS, AAAI, etc. multiple times. %He is on the editorial board of Internation Journal of Computer Vision (IJCV).
\end{IEEEbiographynophoto}
\vskip -2\baselineskip plus -1fil
\begin{IEEEbiographynophoto}
%[{\includegraphics[width=1in,height=1.25in,clip,keepaspectratio]{figures/bio/wangwenhai.png}}]
{Wenhai Wang} is a post-doctoral fellow at the Chinese University of Hong Kong. He obtained the Ph.D. degree from Nanjing University. He was a research scientist at the Shanghai AI Laboratory. His main research interests include foundation models, object detection/segmentation, autonomous driving perception, and optical character recognition.
\end{IEEEbiographynophoto}

\vskip -2\baselineskip plus -1fil

\begin{IEEEbiographynophoto}
{Lewei Lu} received the M.S in Computer Science from SCUT-MSRA joint-training program in 2017. He is currently a Researcher at SenseTime. He was a Researcher at Microsoft Research Asia between 2017 and 2019. His research interests focus on high-level vision and multi-modal learning, foundation model.
\end{IEEEbiographynophoto}

\vskip -2\baselineskip plus -1fil

\begin{IEEEbiographynophoto}
{Huijie Wang} received the B.Sc. in Computer Science from Karlsruhe Institute of Technology in 2019, and the M.Sc. in Computer Science from Technical University of Munich in 2022. He is currently a Researcher at OpenDriveLab, Shanghai AI Lab. His research interests include autonomous driving and computer vision.
\end{IEEEbiographynophoto}

\vskip -2\baselineskip plus -1fil

\begin{IEEEbiographynophoto}
{Jia Zeng} received received the B.E. from Central South University in 2017 and the Ph.D. degree from Shanghai Jiao Tong University in 2023. He is currently a researcher at OpenDriveLab, Shanghai AI Laboratory. His reseach interests lie in computer vision and autonomous driving.  
\end{IEEEbiographynophoto}

\vskip -2\baselineskip plus -1fil

\begin{IEEEbiographynophoto}%[{\includegraphics[width=1in,height=1.25in,clip,keepaspectratio]{figures/bio/lizhiqi.jpeg}}]
{Zhiqi Li} received the B.E. in Computer Science from Nanjing University in 2020. He is currently a Ph.D. candidate at Nanjing University. He is also a research intern at Shanghai AI lab.
His research interests include autonomous driving perception and computer version. He won Frist Place at Waymo Open Dataset Challenge 2022 and the Best Champion at the Occupancy Predcition Task of Autonomous Driving Challenge at CVPR 2023.
\end{IEEEbiographynophoto}

\vskip -2\baselineskip plus -1fil

\begin{IEEEbiographynophoto}
%[{\includegraphics[width=1in,height=1.25in,clip,keepaspectratio]{figures/bio/jiazhi.jpg}}]
{Jiazhi Yang} received the B.E. in Computer Science from Sichuan University in 2022. He is currently a Researcher at OpenDriveLab, Shanghai AI Lab. His research interests include autonomous driving and visual intelligence. He is the first author of UniAD that won Best Paper at CVPR 2023. He won the First Place at Waymo Challenge 2022. 
\end{IEEEbiographynophoto}

\vskip -2\baselineskip plus -1fil

\begin{IEEEbiographynophoto}
{Hanming Deng} received the M.S in Computer Science from Shanghai Jiaotong Univercity in 2020. He is currently a Researcher at Sensetime. His research interests focus on perception and planning in autonomous driving. He won the First Place at Waymo Challenge 2022.
\end{IEEEbiographynophoto}

\vskip -2\baselineskip plus -1fil

\begin{IEEEbiographynophoto}
{Hao Tian} received the B.E. in Electronic Information from Huazhong University of Science and Technology in 2018 and M.S in Mathmatic from Uni-Heidelberg in 2022. He is currently a Researcher at Sensetime. His research interests include autonomous driving, multi modal foundation model. 
\end{IEEEbiographynophoto}

\vskip -2\baselineskip plus -1fil

\begin{IEEEbiographynophoto}
{Enze Xie} received the B.S. degree from the Nanjing University of Aeronautics and Astronautics, Nanjing, China, in 2016, and the M.S. degree from Tongji University, Shanghai, China, in 2019. He is currently working toward the Ph.D. degree with the Department of Computer Science, The University of Hong Kong, Hong Kong.,His main research interests include object detection in 2-D and 3-D and transformers.
\end{IEEEbiographynophoto}

\vskip -2\baselineskip plus -1fil

\begin{IEEEbiographynophoto}
{Jiangwei Xie} is a researcher of the Fundamental Vision group in Sensetime. He received his M.S. degree in computer science from the ShanghaiTech University in 2022. His current research interest lies in autonomous driving.
\end{IEEEbiographynophoto}

\vskip -2\baselineskip plus -1fil

\begin{IEEEbiographynophoto}
%[{\includegraphics[width=1in,height=1.25in,clip,keepaspectratio]{figures/bio/chenli.jpeg}}]
{Li Chen} received the B.E. in Mechanical Engineering from Shanghai Jiao Tong University in 2019, and the M.S. in Robotics from University of Michigan, Ann Arbor, USA in 2020. 
He is currently a Researcher at OpenDriveLab, Shanghai AI Lab.
 % , Shanghai, China. 
 His research interests include autonomous driving and computer vision.
 % and machine learning. 
 He is a Recipient of WAIC Yunfan Award, Outstanding Reviewer at CVPR 2023 and the first author of UniAD that won Best Paper at CVPR 2023.
 % has published first/co-authored paper in ECCV, CoRL, NeurIPS on autonomous driving.
\end{IEEEbiographynophoto}

\vskip -2\baselineskip plus -1fil

\begin{IEEEbiographynophoto}
{Tianyu Li} received a Master's degree from Beihang University and is currently pursuing a Ph.D. at Fudan University. Additionally, he serves as an intern at OpenDriveLab, Shanghai AI Laboratory. His primary research interests revolve around autonomous driving and object detection. 
\end{IEEEbiographynophoto}

\vskip -2\baselineskip plus -1fil

\begin{IEEEbiographynophoto}
{Yang Li} received a Master's degree from Donghua University and is currently a researcher at OpenDriveLab, Shanghai AI Lab. His research interests include data engine and computer vision.
\end{IEEEbiographynophoto}

\vskip -2\baselineskip plus -1fil

\begin{IEEEbiographynophoto}
{Yulu Gao} is a PhD student from Computer Science and Engineering, Beihang University. He received the Bachelor degree from Beihang University. His research interests include object detection and visual tracking.
\end{IEEEbiographynophoto}

\vskip -2\baselineskip plus -1fil

\begin{IEEEbiographynophoto}
{Xiaosong Jia} is currently a PhD student at Department of Computer Science and Engineering, Shanghai Jiao Tong University (SJTU), Shanghai. Before that, he earned B.E. in IEEE Honor class at SJTU. His research interests include autonomous driving and machine learning, with (co-) first-authored papers published in TPAMI, CVPR, ICCV, NeurIPS, RAL, etc.
\end{IEEEbiographynophoto}

\vskip -2\baselineskip plus -1fil

\begin{IEEEbiographynophoto}
{Si Liu} is a Professor in Beihang University. She received Ph.D. degree from Institute of Automation, Chinese Academy of Sciences. She has been Research Assistant and Postdoc in National University of Singapore. She was a visiting scholar of Microsoft Research Asia. She has published over 40 cuttingedge papers on image editing and segmentation on TPAMI, IJCV, CVPR, ECCV and ICCV. She was the recipient of Best Paper of ACM MM 2013 and 2021, Best demo award of ACM MM 2012. 
%She was the Champion of CVPR 2017 Look Into Person Challenge. She is the organizer of ECCV 2018 and ICCV 2019 Person in Context Workshop/challenge. 
She servers as an area chair of ICCV 2019, CVPR 2020 and ECCV 2020, SPC of IJCAI 2019 and AAAI 2019. 
\end{IEEEbiographynophoto}

\vskip -2\baselineskip plus -1fil

\begin{IEEEbiographynophoto}
{Jianping Shi} (Member, IEEE) received the B.Eng. degree from Zhejiang University in 2011 and the Ph.D. degree in Computer Science and Engineering Department, Chinese University of Hong Kong, in 2015, under the supervision of Prof. Jiaya Jia. She is currently an Executive Research Director with SenseTime. Her team works on developing algorithms for autonomous driving, scene understanding, and remote sensing. She has served regularly on the organization committees for numerous conferences, such as the Area Chair of CVPR and ICCV.
\end{IEEEbiographynophoto}
\vskip -2\baselineskip plus -1fil

% \textbf{Ping Luo} received the Ph.D. degree in information engineering from The Chinese University of Hong Kong (CUHK) in 2014 supervised by Prof. Xiaoou Tang and Prof. Xiaogang Wang. He is currently an Assistant Professor with the Department of Computer Science, HKU. He was a Postdoctoral Fellow with CUHK from 2014 to 2016. He joined SenseTime Research as a Principal Research Scientist from 2017 to 2018. His research interests are machine learning and computer vision. He has published more than 100 peer-reviewed papers in top-tier conferences and journals, such as IEEE Transactions on Pattern Analysis and Machine Intelligence, IJCV, ICML, ICLR, CVPR, and NIPS. His work has high impact with 18000 citations according to Google Scholar and his current research interests include autonomous driving, deep learning, and computer version.

% \textbf{Junchi Yan} (S'10-M'11-SM'21) is currently an full Professor with Department of Computer Science and Engineering, Shanghai Jiao Tong University, Shanghai, China. Before that, he was a Senior Research Staff Member with IBM Research where he started his career since April 2011. His research interests include machine learning and computer vision. He regularly serves as Senior PC/Area Chair for NeurIPS, ICLR, ICML, CVPR, AAAI, IJCAI, ACM-MM and Associate Editor for the Pattern Recognition Journal.

\begin{IEEEbiographynophoto}
%[{\includegraphics[width=1in,height=1.25in,clip,keepaspectratio]{figures/bio/lindahua.png}}]
{Dahua Lin} received the B.Eng. degree from the University of Science and Technology of China, Hefei, China, in 2004, the M. Phil. degree from The Chinese University of Hong Kong, Hong Kong, in 2006, and the Ph.D. degree from the Massachusetts Institute of Technology, Cambridge, MA, USA, in 2012. From 2012 to 2014, he was a Research Assistant Professor with Toyota Technological Institute at Chicago, Chicago, IL, USA. He is currently an Associate Professor with the Department of Information Engineering, The Chinese University of Hong Kong (CUHK), and the Director of CUHK-SenseTime Joint Laboratory. His research interests include computer vision and machine learning.
\end{IEEEbiographynophoto}
\vskip -2\baselineskip plus -1fil
\begin{IEEEbiographynophoto}
%[{\includegraphics[width=1in,height=1.25in,clip,keepaspectratio]{figures/bio/qiaoyu.jpg}}]
{Yu Qiao}(Senior Member, IEEE) is a professor with Shanghai AI Laboratory and the Shenzhen Institutes of Advanced Technology (SIAT), Chinese Academy of Science. His research interests include computer vision, deep learning, and bioinformation. He has published more than 300 papers in IEEE TPAMI, IJCV, IEEE TIP, CVPR, ICCV etc. His H-index is 72, with 42,700 citations in Google scholar. He is a recipient of the distinguished paper award in AAAI 2021. His group achieved the first runner-up at the ImageNet Large Scale Visual Recognition Challenge 2015 in scene recognition, and the winner at the ActivityNet Large Scale Activity Recognition Challenge 2016 in video classification.
\end{IEEEbiographynophoto}

\clearpage
\normalsize	
\appendices
\section{More Related Work}\label{sec: task_def_sol_apdx}
In this section, we describe 3D perception tasks and their conventional solutions, including monocular camera-based 3D object detection, LiDAR-based 3D object detection and segmentation and sensor-fusion-based 3D object detection.

\subsection{Monocular Camera-based Object Detection}
% \smch{should be re-written in an uniform way, say what's done in perspective view, consider structure in \cite{ma2022vision}}
% written by Enze
Monocular camera-based methods take an RGB image as input and attempt to predict the 3D location and category of each object. The main challenge of monocular 3D detection is that the RGB image lacks depth information, so these kinds of methods need to predict the depth. Due to estimating depth from a single image being an ill-posed problem, typically monocular camear-based methods have worse performance than LiDAR-based methods.

Mono3D~\cite{Chen_2016_CVPR} proposes a 3D proposal generation method with contextual and semantic information and uses an R-CNN-like architecture to predict the 3D boxes.
Deep3DBox~\cite{mousavian20173d} predicts the 2D bounding box first, then estimate the 3D information from the 2D box, extending an advanced 2D detector.
3D-RCNN~\cite{kundu20183d} also extends R-CNN in 2D detection, not only predicting the 3D bounding box but also rendering the shape of each object.
CenterNet~\cite{zhou2019objects} is an anchor-free 2D detector and predicts the object center and the distance to the box. It can also easily extend to 3D detection by predicting the 3D size, depth, and orientation.
Pseudo-Lidar~\cite{wang2019pseudo} first predicts the depth map, then uses camera parameters to back-project the image with depth to the 3D point cloud in the LiDAR coordinate. Then any LiDAR-based detectors can be applied to the predicted ``pseudo LiDAR''. 
The M3D-RPN~\cite{brazil2019m3d} uses a single-stage 3D region proposal network, with two parallel paths for feature extraction: one global path with regular convolution, and one local depth-aware path using the non-shared Conv kernels in the \textit{H} space.
MonoPSR~\cite{ku2019monocular} first detects 2D boxes and inputs the whole image and the cropped objects, then two parallel branches are performed: a proposal generation module to predict the coarse boxes, and an instance reconstruction module to predict the mesh of the cropped object. In the second proposal refinement module, the reconstructed instance is used to get the refined 3D box.
% OFTNet~\cite{Roddick2019OrthographicFT} transforms the image feature map to the bird's eye view feature map with an orthographic feature transform, which does not rely on any explicit depth but only uses camera parameters. 
ROI-10D~\cite{manhardt2019roi} predicts the 2D box and depth first, then lift them to 3D RoIs and uses RoIAlign to crop the 3D feature for regressing the 3D bounding box. It can also easily extend to 3D mesh prediction.
SMOKE~\cite{liu2020smoke} is a one-stage anchor-free method that directly predicts the 3D center of the objects and the 3D box size and orientation.  
% CaDDN~\cite{reading2021categorical} predicts the implicit depth estimation and projects the image feature to the BEV feature using the predicted depth.
% %
% The feature transformation is similar to OFTNet, but CaDDN uses the depth label to supervise the depth estimation, which obtains sharper and more accurate depth.
FCOS3D~\cite{wang2021fcos3d} is a recent representative monocular 3D detection method which extends the state-of-the-art 2D detector, FCOS~\cite{tian2019fcos}. The regression branch is similar to CenterNet; they all add the depth and the box size prediction in the regression branch. FCOS3D also introduces some tricks \textit{e.g.} flip augmentation and test-time augmentation.
PGD~\cite{wang2022probabilistic} analyzes the importance of depth estimation in the monocular 3D detection and then proposes a geometric relation graph to capture the relation of each object, resulting in better depth estimation.

\subsection{LiDAR Detection and Segmentation}
% \smch{summerizes as point base, voxel base, range view base, consider the structure from \cite{mao20223d}}
% written by huijie
LiDAR describes surrounding environments with a set of points in the 3D space, which capture geometry information of objects. Despite the lack of color and texture information and the limited perception range, LiDAR-based methods outperform camera-based methods by a large margin benefit from depth prior.

\subsubsection{Detection}
% written by huijie
As data collected by LiDAR is formatted as point clouds, it is natural to construct neural networks directly on points. Point-based methods process on raw point cloud data to conduct feature extraction. VoteNet~\cite{qi2019votenet} directly detects objects in point clouds based on Hough voting. 
PointRCNN~\cite{shi2019pointrcnn} is a two-stage method for more accurate 3D detection. In the first stage, 3D proposals are generated via segmenting the point cloud into foreground and background points, and then proposals are refined to obtain the final bounding boxes in the second stage. 
Part-$A^2$~\cite{shi2019parta2} extends PointRCNN with the part-aware and neural aggregation network. 
H3DNet~\cite{zhang2020h3dnet} predicts a hybrid set of geometric primitives, which are then converted to object proposals by defining a distance function between an object and the geometric primitives. A drawback of voting methods is that outlier votes from backgrounds hurt the performance.
Pointformer~\cite{pan2020pointformer} designs a Transformer backbone, which consists of Local Transformer and Global Transformer, to learn features effectively. 
BRNet~\cite{cheng2021back} back-traces the representative points from the vote centers and revisits complementary seed points around these generated points.
Instead of grouping local features, Group-Free~\cite{liu2021groupfree} obtain the feature of an object from all the points using an attention mechanism. 
RBGNet~\cite{wang2022rbgnet} proposes a ray-based feature grouping module to learn better representations of object shapes to enhance cluster features. 
3DSSD~\cite{yang20203dssd} first presents a lightweight and effective point-based single-stage 3D detector. It utilizes 3D Euclidean distance and Feature-FPS as sampling strategy.

%SegVoxelNet~\cite{yi2020segvoxelnet}
%SE-SSD~\cite{zheng2021sessd}
% SSN~\cite{zhu2020ssn}

\subsubsection{Segmentation}
% written by huijie
Besides 3D object detection, the task of point cloud segmentation provides the whole scene understanding from point cloud data. 
Some works focus on indoor point cloud segmentation.
PointNet~\cite{qi2017pointnet} provides a unified operator by combining MLP and max pooling to learn point-wise features directly from point clouds. 
PointNet++~\cite{qi2017pointnet++} further introduces set abstraction to form a local operation for more representative feature extraction. 
SpiderCNN~\cite{xu2018spidercnn} proposes a novel convolutional architecture for efficient geometric feature extraction. 
DGCNN~\cite{wang2018dgcnn} proposes EdgeConv to learn incorporates local neighborhood information. PointCNN~\cite{li2018pointcnn} uses $X$-transformation to simultaneously weight and permute the input features for subsequent convolution on the transformed features. 
KPConv~\cite{thomas2019kpconv} stores convolution weights in Euclidean space by kernel points, which are applied to the input points close in the neighborhood. 
Point Transformer~\cite{zhao2021pointtransformer} is a Transformer-based method that designs self-attention layers for the point cloud.

Different from indoor sense segmentation, outdoor segmentation models are designed for more imbalance point distribution. 
RandLA-Net~\cite{hu2020RandLA} uses random point sampling instead of complicated point selection methods for efficient and lightweight architecture. 
PolarNet~\cite{zhang2020polarnet} uses polar BEV representation to balance the number of points across grids.
Cylinder3D~\cite{zhu2020cylinder3d} introduces a novel cylindrical partition voxelization method and asymmetrical 3D convolution networks to ease the problem of imbalance point distribution. 
(AF)$^2$-S3Net~\cite{cheng2021af2} fuses the voxel-based and point-based learning methods into a unified framework by the proposed multi-branch attentive feature fusion module and adaptive feature selection module.
TornadoNet~\cite{gerdzhev2021tornado} incorporates bird-eye and range view feature with a novel diamond context block. 
AMVNet~\cite{liong2020amvnet} aggregates the semantic features of individual projection-based networks.
DRINet~\cite{ye2021drinet} designs architecture for dual-representation iterative learning between point and voxel. 
DRINet++~\cite{ye2021drinet++} extends DRINet by enhancing the sparsity and geometric properties of a point cloud. 
Sparse Point-Voxel Convolution (SPVConv)~\cite{tang2020spvcnn} uses an auxiliary point-based branch to preserve high-resolution features from voxelization and aggressive downsampling. Information exchanges are performed in different stages of the network between the point-based and voxel-based branches. 
RPVNet~\cite{xu2021rpv} devises a deep fusion network with multiple and mutual information interactions among voxel-, point-, and range-view. 
2D3DNet~\cite{genova20212d3dnet} uses labeled 2D images to generate trusted 3D labels for network supervision. 
2DPASS~\cite{yan20222dpass} boosts the point clouds representation learning by distilling image features into the point cloud network.

\subsection{Sensor Fusion}
% written by huijie / yulu / tianhao
% 
% Perception is a vital task for autonomous driving, since the following tasks, namely prediction and planning, depend on the correct perception of the surrounding world. Multiple sensors are utilized to perform the perception task, such as camera, LiDAR, and Radar. 
Modern autonomous vehicles are equipped with different sensors such as cameras, LiDAR and Radar. Each sensor has advantages and disadvantages.
Camera data contains dense color and texture information but fails to capture depth information. 
LiDAR provides accurate depth and structure information but suffers from limited range and sparsity. 
Radar is more sparse than LiDAR, but has a longer sensing range and can capture information from moving objects. 
Ideally, sensor fusion will push the performance upper bound of the perception system, but how to fuse the data from different modalities is a challenging problem.
% With its own advantages and disadvantage, each type of sensor provides partially corrected information for perception. However, the differences in data representation for different sensors make it difficult to fuse correct perception from different sensors efficiently and effectively. 

% The projection guided approaches fuse the image feature to LiDAR feature by projecting different point cloud representations to image to obtain image feature with different fusion operations.

MVX-Net~\cite{mvxnet2019} projects the voxel region to the image and applies ROI Pooling to extract the image feature.
MMF~\cite{liang2019mmf} and ContFuse~\cite{liang2019contfuse} project LiDAR points from each BEV feature to the image feature map and apply the continuous convolutions to fuse the feature. 
PointAugmenting~\cite{wang2021pointaugmenting} constructs image feature point clouds by projecting all LiDAR points to image features. The fused feature is obtained by concatenation of two BEV features from LiDAR feature point clouds and image feature point clouds after the 3D backbone. 
AutoAlignV2~\cite{chen2022autoalignv2} utilizes the deformable attention~\cite{zhu2020deformable} to extract the image feature with LiDAR feature after projecting the LiDAR point to image as reference point. 
DeepFusion~\cite{li2022deepfusion} applies the transformer~\cite{vaswani2017attention} for fusion by using the voxel feature as query and image feature as key and value.  
CenterFusion~\cite{nabati2021centerfusion} uses the frustum association to fuse the radar feature and image feature with preliminary 3D boxes.
FUTR3D~\cite{chen2022futr3d} projects the 3D reference points to feature maps from different modalities and fuses features using deformable attention.  
TransFusion~\cite{bai2022transfusion} uses a spatially modulated cross attention to fuse the image features and the LiDAR BEV features, to involve the locality inductive bias. 
DeepInteraction~\cite{yang2022deepinteraction} preserves modal-specific representations instead of a single hybrid representation to maintain modal-specific information. 
%UVTR~\cite{li2022uvtr} applies the LSS~\cite{philion2020lift} to project the image features to BEV feature. After constructing the LiDAR BEV feature and image BEV feature, two features are simply added for fusion.
%Both BEVFusion~\cite{liu2022bevfusion,liang2022bevfusion} utilize the LSS~\cite{philion2020lift} to project image and  LiDAR features to BEV space. 
%To fuse the two BEV features,~\cite{liang2022bevfusion} applies the dynamic fusion and~\cite{liu2022bevfusion} applies fully convolutional fusion. 

Besides fusing features from different modalities in the middle of neural networks, early fusion methods augment the LiDAR input with the aid of image information. 
PointPainting~\cite{vora2020pointpainting} concatenates the one-hot label from the semantic segmentation to point features according to the projection relationship. Then the augmented point feature is fed to any LiDAR-only 3D detector. F-PointNet~\cite{qi2018frustumpointnets} utilizes the 2D detector to extract the point cloud within the viewing frustum of each 2D bounding box. Then each viewing frustum with the corresponding point cloud is fed to a LiDAR-only 3D detector. 

Late fusion methods fuse the multi-modal information after the generation of object proposals. 
MV3D~\cite{chen2017multi} takes the BEV, front view LiDAR, and image as input. After obtaining the 3D proposals from BEV and projecting 3D proposals to other modalities, features from different modalities of the same 3D proposal are fused to refine proposals. 
AVOD~\cite{ku2018avod} improves the MV3D~\cite{chen2017multi} with high resolution image feature. 
CLOCs~\cite{pang2020clocs} operate on the prediction results from the 2D and 3D detector with a small network to adjust the confidence of the pair-wise 2D and 3D with non-zero IoU. 

\section{Preliminary in 3D vision}\label{Preliminary_3dvision}
%\subsection{Mathematical Formulation in 3D Vision}
% written by Jia Zeng
% HY: two nice figures inside
% https://zhuanlan.zhihu.com/p/68863677
% https://zhuanlan.zhihu.com/p/54139614

% \smch{this is general 3D 2D projection, and this is for view transformation, which we said as the core of BEV perception, this is also background knowledge}

\begin{figure}[h]
\begin{center}
\includegraphics[width=0.45\textwidth]{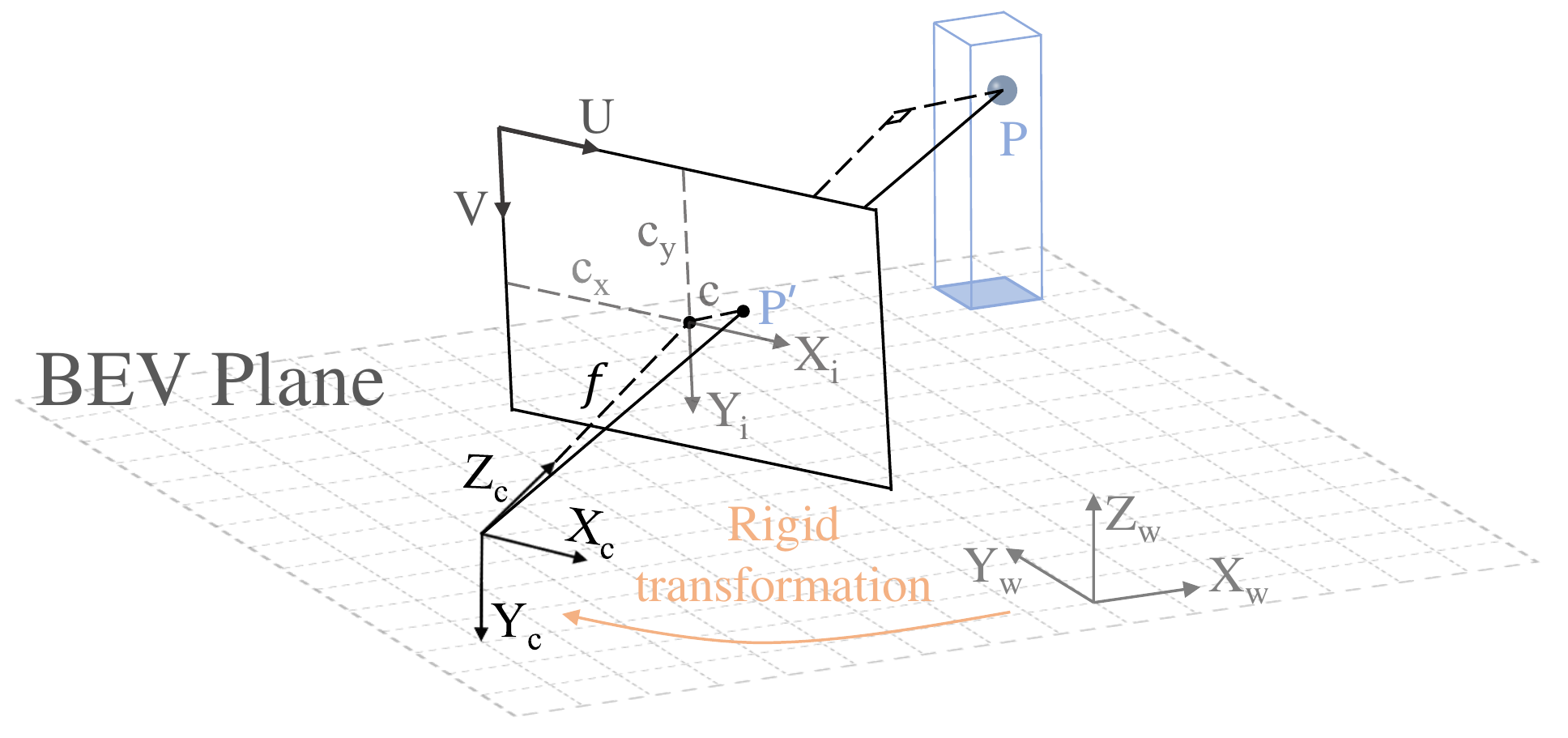}
\end{center}
\caption{View transformation from perspective view to bird's-eye-view (BEV). The $(X_w, Y_w, Z_w)$, $(X_c, Y_c, Z_c)$ represent world coordinate and camera coordinate, and the $(X_i, Y_i)$, $(U, V)$ represent image coordinate and pixel coordinate. A pillar is lifted from the BEV plane. $P$, $P'$ corresponds to the 3D point from the pillar and the projected 2D point from the camera view, respectively. %Knowing 
Given the world coordinates of $P$ and %based on 
the intrinsic and extrinsic parameters of the camera, the pixel coordinates of $P'$ can be obtained.
}
\label{fig:prel_3D_vision}
\end{figure}

Explicit BEV feature construction typically requires indexing local image-view features based on 3D-to-2D projection. Fig.~\ref{fig:prel_3D_vision} depicts the view transformation in BEVFormer~\cite{li2022bevformer}. 
% In Fig.~\ref{fig:prel_3D_vision}, 
A pillar is lifted from the BEV plane, and a 3D point inside the pillar is projected to the camera view. The projection process involves the transformation between the world, camera, image, and pixel coordinate systems.

The transformation from the world coordinate to the camera coordinate is a rigid transformation, requiring translation and rotation only. Let $P_w=[x_w,y_w,z_w,1]$, $P_c=[x_c,y_c,z_c,1]$ be the homogenous representations for a 3D point $P$ in the world coordinate and the camera coordinate respectively. Their relation can be described as follows:
\begin{equation}\label{prel-3dv-E1}
P_c = \begin{bmatrix} x_c \\ y_c \\ z_c \\ 1 \end{bmatrix} = 
\begin{bmatrix} \bm{R} & \bm{T} \\ \bm{0}^T & 1 \end{bmatrix}
\begin{bmatrix} x_w \\ y_w \\ z_w \\ 1 \end{bmatrix},
\end{equation}
where $R$, $T$ refers to a rotation matrix and a translation matrix respectively. 

The image coordinate system is introduced to describe the perspective projection from the camera coordinate system onto the image. When not taking the camera distortion into consideration, the relationship between a 3D point and its projection on image place can be simplified as a pinhole model. The image coordinates $(x_i, y_i)$ can be calculated by Eqn.~\ref{prel-3dv-E2}:
\begin{equation}\label{prel-3dv-E2}
\begin{cases}
\begin{aligned}
x_i &= f \cdot \frac{x_c}{z_c} \\
y_i &= f \cdot \frac{y_c}{z_c}
\end{aligned},
\end{cases}
\end{equation}
where $f$ represents the focal length of camera.

There is a translation and scaling transformation relationship between the image coordinate system and the pixel coordinate system. 
%Suppose the scale factors along the $x$ axis and $y$ axis be $\alpha$ and $\beta$ respectively, and suppose the translation distances between the origins of the two coordinate systems be $c_x$ and $c_y$ respectively. 
Let us denote $\alpha$ and $\beta$ representing the scale factors to the abscissa and the ordinate, and $c_x$, $c_y$ representing the translation value to the coordinate system origin. The pixel coordinates $(u, v)$ is mathematically represented by Eqn.~\ref{prel-3dv-E3}:
\begin{equation}\label{prel-3dv-E3}
\begin{cases}
\begin{aligned}
u &= \alpha x + C_x \\
v &= \beta y + C_y
\end{aligned}.
\end{cases}
\end{equation}

With Eqn.~\ref{prel-3dv-E2} and Eqn.~\ref{prel-3dv-E3}, setting $f_x = \alpha f$, $f_y = \beta f$, we could derive Eqn.~\ref{prel-3dv-E4}:
\begin{equation}\label{prel-3dv-E4}
z_c \begin{bmatrix}
u \\
v \\
1
\end{bmatrix}
= \begin{bmatrix}
f_x & 0 & c_x \\
0 & f_y & c_y \\
0 & 0 & 1
\end{bmatrix}
\begin{bmatrix} x_c \\ y_c \\ z_c \\ \end{bmatrix}.
\end{equation}

In summary, the relationship between the 3D point $P$ in the world coordinate system and its projection $P'$ in the pixel coordinate system can be described as:
\begin{equation}\label{prel-3dv-E5}
\begin{aligned}
z_c \begin{bmatrix} u \\ v \\ 1 \end{bmatrix}
&= \begin{bmatrix}
f_x & 0 & c_x \\
0 & f_y & c_y \\
0 & 0 & 1
\end{bmatrix}
\begin{bmatrix}
\bm{R} & \bm{T}
\end{bmatrix}
\begin{bmatrix}
x_w \\
y_w \\
z_w \\
1
\end{bmatrix}, \\
& = \bm{K} \begin{bmatrix}
\bm{R} & \bm{T}
\end{bmatrix}
\begin{bmatrix}
x_w & y_w & z_w & 1
\end{bmatrix}^T .
\end{aligned}
\end{equation}

The matrix $\bm{K} = \begin{bmatrix} f_x & 0 & c_x \\ 0 & f_y & c_y \\ 0 & 0 & 1 \end{bmatrix}$ is termed as the camera intrinsics, and the matrix $\begin{bmatrix} \bm{R} & \bm{T} \end{bmatrix}$ is called as the camera extrinsics. With the world coordinates of 3D points, the camera intrinsics and extrinsics, the projections on the image view can be obtained by aforementioned transformation.

\section{Dataset and Evaluation Metrics}
\subsection{Argoverse Dataset}
% \noindent\textbf{Argoverse Dataset}
Argoverse~\cite{Argoverse,Argoverse2} is the first self-driving dataset with HD-Map. The sensor set of Argoverse contains 2 LiDARS, 7 ring cameras, and two stereo cameras. The early version, termed Argoverse 1~\cite{Argoverse}, supports two tasks: 3D tracking and motion forecasting. And the new Argoverse 2~\cite{Argoverse2} supports more tasks: 3D object detection, unsupervised learning, motion prediction and changed map perception, which is more diverse and challenging.
% \subssection{Evaluation Metrics}

\subsection{Evaluation Metrics} 

\noindent \textbf{PKL.}
Planning KL-Divergence~(PKL) is a new neural planning metric proposed in CVPR 2020~\cite{philion2020learning}, which is based on the KL divergence of the planner's trajectory and the route of the ground truth. The planner's trajectory is generated by giving the detection results from the trained detectors. The PKL metric is always non-negative. The smaller PKL scores mean the detection performance is better.

\smallskip
The localization affinity of LET-3D-APL is defined as:
\begin{itemize}
    \item If no longitudinal localization error, the localization affinity = 1.0.
    \item If the longitudinal localization error is equal to or exceeds the maximum longitudinal localization error, the  localization affinity = 0.0. 
    \item The localization affinity is linearly interpolated  between 0.0 and 1.0.
\end{itemize}

\section{Industrial View on BEV}

Here we depict in detail the input and network structure in BEV architecture from different companies in Tab.~\ref{tab:bev_arch_summary}.

\begin{table*}
\renewcommand\arraystretch{1.2}
  \setlength\tabcolsep{0.1cm}
  \centering
  \caption{
% %   Important 
Detailed input and network options for BEV architectures.
%   BEV perception architecture in industry summary. The information is collected from public resources. So this table \textbf{only includes publicly shared BEV architectures}. The 
  As we can observe, modality and feature extractor are different; Transformer and ViDAR are the most common choice for BEV transformation in industry. 
  ``-'' indicates unknown information. 
  }
%   \scalebox{0.92}{
  \begin{tabular}{ l |c|c|c|c|c|c }
  \toprule 
  \multirow{2}{*}{\textbf{Company}} & \multicolumn{3}{c|}{\textbf{Modality}}  & \multirow{2}{*}{\textbf{Feature Extractor}}   & \multirow{2}{*}{\textbf{BEV Transformation}}& \multirow{2}{*}{\textbf{Temporal \& Spatial Fusion }}   \\ 
  \cline{2-4} & Camera & LiDAR & IMU/GPS & & & \\
\midrule
Tesla~\cite{tesla_ai_day} & \cmark & \xmark & \cmark  &  RegNet$+$BiFPN & \begin{tabular}[c]{@{}c@{}}Transformer\\ ViDAR\end{tabular} & \cmark \\
% Waymo & MC & -- & ViDAR & -- \\
Mobileye~(SuperVison)~\cite{mobile2020ces} & \cmark & \cmark & - & - & ViDAR & - \\
Hrizon Robotics~\cite{horizon2022bev} & \cmark & \cmark & \cmark & - & Transformer & \cmark \\
HAOMO.AI~\cite{haomo2022aiday} & \cmark& \cmark & \cmark & \begin{tabular}[c]{@{}c@{}}ResNet$+$FPN~\\ Pillar-Feature
Ne\end{tabular} & Transformer & \cmark \\
PhiGent Robotics~\cite{phigent2022vidar} & \cmark & \xmark & \xmark & - & ViDAR & - \\

    \bottomrule
  \end{tabular}
%   }
   
  \label{tab:bev_arch_summary}
\end{table*}

\section{Additional Recipes} \label{additional_recipes_apdx}

\subsection{3D Detection Head in BEVFormer++}
% \subsubsection{3D Detection}\label{sec: cam_3d_head}
%written by hanming
We mainly refer to BEV camera detection task for this trick.
%
% BEVFormer provides high quality BEV features, which allows us to explore the design of different detection heads in both image and LiDAR-based 3D detection.
%
In BEVFormer++, three detection heads are adopted.
% to combine with BEVFormer into three different detection models.
%
Correspondingly, these heads cover three categories of the detector design, including anchor-free, anchor-based and center-based. 
We choose various types of detector heads that differentiate in design as much as possible, so as to fully leverage detection frameworks for the potential ability in different scenarios. The diversity of heads facilitate the ultimate ensemble results.

%DETR-Based Head: Deformable DETR
Original BEVFormer uses a modified Deformable DETR decoder as its 3D detector~\cite{li2022bevformer,zhu2020deformable, carion2020end},  which can detect 3D bounding boxes end-to-end without NMS. 
For this head, we follow the original design but use Smooth L1 loss to replace the origin L1 loss.  
Most of our baseline experiments with tricks are implemented on the DETR decoder, as described in Tab.~\ref{tab:camera_tricks} ID 1-15.

%Dense Head: FCOS, centernet
% \HY{Syntax error here:
BEVFormer++ adopts the FreeAnchor~\cite{zhang2019freeanchor} and CenterPoint~\cite{yin2021center} as alternative 3D detectors, where FreeAnchor is an anchor-based detector that can automatically learns the matching of anchors, and CenterPoint being a center-based anchor-free 3D detector.
Ablation study in Tab.~\ref{tab:camera_tricks} ID 16-20 indicates that these heads perform differently under various settings. This is important for the ensemble part since prediction heads provide various distribution during inference.
More details about the ensemble technique can be found in Sec.~\ref{sec:ensemble_cam}.
It is worth noting that 3D decoder is far from being well developed, as efficient query design~\cite{li2022dn, meng2021conditional} gets it prosperity in 2D perception.
How to transfer those success to 3D perception field would be the next step in this community.
%
% We compute the LET-IoU between prediction results and ground truth. 
% The Centerpoint~\cite{yin2021center} head is the last detector head we utilize, which is a powerful center-based anchor-free 3D detector.

%\subsubsection{Semantic Map Segmentation}
%Dense Head: Deeplabv3+
%DETR-Based Head: mask decoder in panoptic segformer
%\HY{TODO}

\subsection{Test-time Augmentation (TTA)}
\subsubsection{BEV Camera-only Detection}
%written by hanming
Common test-time augmentation for 2D tasks, including multi-scale and flip testing are examined to improve accuracy in 3D case as well.
In  BEVFormer++, this part is simply explored in form of utilizing standard data augmentation such as multi-scale and flip.
The extent of multi-scale augmentation are the same as training, varying from 0.75 to 1.25.
Related experiments are shown in Tab.~\ref{tab:camera_tricks} ID 13.
% \HY{TODO}

\subsubsection{LiDAR Segmentation}
During inference, multiple TTAs are utilized, including rotation, scaling, and flipping. For scaling, the scaling factor is set to $\{0.90, 0.95, 1.00, 1.05, 1.10\}$ for all models, since a larger or smaller scaling factor is harmful to model performance. Flipping remains the same as in the training phase, namely, along $X$ axis, $Y$ axis, and both $X$ and $Y$ axes. Rotation angle is set to $\{-\frac{\pi}{2}, 0, \frac{\pi}{2}, \pi\}$. A more fine-grained scaling factor or rotation angle could be chosen, but with the concern of computation overhead and the strategy of TTA combination, coarse-grained parameters are preferable.

A combination of TTAs would further improve model performance, compared to fine-grained augmentation parameters. However, it is time-consuming due to the multiplication of TTAs. 
The combined model-dependent TTAs
% A combination of TTAs, which is model-dependent, 
with $20$ inference time is employed. Grid search of combination strategy could be conducted. Empirically, a combination of scaling and flipping is preferable. An obvious improvement of $1.5$ mIoU can be obtained (see Tab.~\ref{tab:lidarseg_tricks} ID 4).

\begin{table}[t]
\caption{\textbf{BEV Camera detection track.} Performance of expert models. Abbreviation and experiment IDs are inherited from Tab.~\ref{tab:camera_tricks}. The overall LET-mAPL metrics and that on each category or each time-of-day subset of the models are listed in the table. Expert 1 is class rebalanced and Expert 2 is time rebalanced. D/D is short for dawn/dusk.}
\vspace{-0.4cm}
\label{table:ablation_expert}
\begin{center}
\renewcommand\arraystretch{1.2}
\setlength\tabcolsep{0.15cm}
\begin{tabular}{l|l|c|c|c|c|c|c|c}
\toprule
\multirow{2}{*}{ID}& \multirow{2}{*}{Head}        &\multicolumn{6}{c}{LET-mAPL } \\
\cline{3-9}
& & Overall       &Car & Cyc & Ped & Day & Night & D/D\\
\midrule
19 &DeD       &    48.4 &60.4 &37.0&47.8&49.2 &37.0 &49.5 \\
19.1& expert 1  &48.9&61.3&37.1&48.3& - & - & - \\
19.2& expert 2  &48.7& - & - & - &49.5 &36.2 &49.9 \\
\midrule
20&FrA     &  47.2   &62.7 &36.7 & 42.1&48.4 &33.4 &46.0 \\
20.1& expert 1 &47.2&62.8&36.6&42.3& - & - & - \\
20.2& expert 2 &47.0& - & - & - &48.5 &33.3 &46.3 \\
\midrule
22&CeP     &    41.9     &56.8 &29.7 &39.3&42.8 &31.9 &42.2 \\
22.1& expert 1&42.4&57.2&30.8&39.2& - & - & - \\
22.2& expert 2&42.3& - & - & - &43.2 &31.8 & 42.1\\
\bottomrule
\end{tabular}
\end{center}
\vspace{-0.4cm}
\end{table}

\subsection{Ensemble}

\subsubsection{BEV Camera-only Detection}\label{sec:ensemble_cam}
%written by hanming/yulu
The ensemble technique usually varies among datasets to be tested on; the general practice used in 2D/3D object detection could be applied to BEV perception with some modification.
Take 
% the first place solution in Waymo Open Challenge 2022 
BEVFormer++
for example, in the ensemble phase, we introduce the improved version of the weighted box fusion (WBF)~\cite{solovyev2019weighted}. 
Inspired by Adj-NMS~\cite{liu20201st}, matrix NMS~\cite{wang2020solov2} is adopted following the original WBF to filter out redundant boxes.
In order to generate multi-scale and flip results, a two-stage ensemble strategy is adopted. 
In the first stage, we utilize improved WBF to integrate predictions from the multi-scale pipeline to generate flip and non-flip results for each model.
Related experiments of expert model performance are listed in Tab.~\ref{table:ablation_expert}.
In the second stage, results from all expert models are gathered. 
The improved WBF is used to get the final results. 
This two-stage ensemble strategy improves the model performance by $0.7$ LET-mAPL as shown in Tab.~\ref{tbl:ablation_ensemble}.

Considering the diversity of performance in each model, we argue that the parameter adjustment is much more complex. Hence the evolution algorithm is adopted to search WBF parameters and model weights.
We utilize evolution in NNI~\cite{NNI} to automatically search parameters, where the population size is 100.
The search process is based on the performance of the 3000 validation images; different classes are searched separately.

\begin{table}[t]
\caption{\textbf{BEV Camera detection track.} Ensemble strategies on \texttt{val} set in BEVFormer++. FrA (FreeAnchor head~\cite{zhang2019freeanchor}). DeD (Deformable DETR head~\cite{zhu2020deformable}). CeP (Centerpoint head~\cite{yin2021center}). ``20 models'' denotes expert models under 20 different settings.} 
\vspace{-0.4cm}
\label{tbl:ablation_ensemble}
\begin{center}
    \renewcommand\arraystretch{1.0}
\begin{tabular}{l|c|c|c}
\toprule
Head        &LET-mAPL           &LET-mAP        &LET-mAPH \\
\midrule
FrA     &47.6           &61.4           &57.0\\
+DeD    &49.1           &65.2           &60.2\\
+CeP    &50.6           &66.5          &61.2\\
\midrule
20 models    &53.2   &69.3   &64.9\\
+Two-Stage    &53.9   &69.2   &64.5\\
\bottomrule
\end{tabular}
\end{center}
\vspace{-0.4cm}
% \vspace{-0.3cm}
\end{table}

\subsubsection{LiDAR Segmentation}

%\begin{figure}
%\centering
%\includegraphics[width=0.4\textwidth]{figures/ensemble.png}
%\caption{hierarchical ensemble}
%\label{fig:hierarchical_ensemble}
%\end{figure}

As a per-point classification task, the task of segmentation ensembles per-point probabilities from different models in an average manner. Specifically, probabilities predicted by different models are simply summed, and then per-point classification results are determined with an $argmax$ operation. To improve the diversity of our models, models are trained using a different data resampling strategy called the export model. According to context information about scenes and weather conditions, multiple context-specific models are finetuned on the model trained on all data. As shown in Tab.~\ref{tab:lidarseg_tricks} ID 8 \& 9, the usage of model ensemble and expert model brings an  improvement of $0.7$ and $0.3$ mIoU respectively.

The probabilities of models are aggregated in a hierarchical manner after model-specific TTA. Considering the diversity of models, the model ensemble is processed in two stages. In the first stage, probabilities of homogeneous models, such as models with different hyper-parameters, are averaged with different weights. Then, probabilities of heterogeneous models, namely models with different architectures, are averaged with different weights in the second stage. Annealing algorithm with the maximum trial number $160$ in NNI~\cite{NNI} is utilized to search weights on validation set in both stages.

\subsection{Post-Processing}

\subsubsection{BEV Camera-only Detection}
%written by hanming
While BEV detection removes the burden of multi-camera object level fusion, we observe distinguished fact that can benefit from further post-processing. By nature of BEV transformation, duplicate features are likely to be sampled on different BEV locations along a light ray to camera center. This results in duplicate false detection on one foreground objects, where every false detection have different depth but can all be projected back to the same foreground objects in the image space. To alleviate this issue, it is beneficial to leverage 2D detection results for duplicate removal on 3D detection results, where 2D bounding boxes and 3D bounding boxes are bipartite matched. In our experiments, the 3D detection performance can be improved when using ground truth 2D bounding boxes as the filter. However, when using predicted 2D bounding boxes from the 2D detection head trained with auxiliary supervision as mentioned in Sec.~\ref{sec:camera-loss}, we observe that improvements can be barely obtained. This could be caused by the insufficient training of the 2D detection. Therefore, further investigation of the joint 2D/3D redundant detection removal is in demand.

NMS could be applied depending on whether the detection head design behaves NMS-free property. Usually, for one-to-many assignment, NMS is necessary. Notably, replacing commonly used IoU metric in NMS with the newly proposed LET-IoU~\cite{hung2022let3dap} to remove redundant results can improve detection results. The improvements can be witnessed in Tab.~\ref{tab:camera_tricks} ID 12 \& 17 \& 21. This design is more suitable for BEV camera-only 3D detectors. Since the 3D IoU of two mutually redundant results is numerically small, this often leads to failure in removing  false positive results. With LET-IoU, redundant results tend to obsess higher IoU, thus being removed to a great extent.

\subsubsection{LiDAR Segmentation}
By analyzing the confusion matrix, we observe that most misclassification occurs within similar classes. Thus, semantic classes can be divided into groups, in which classes are intensively confused compared to classes outside the group. Post-processing techniques are conducted on foreground semantic groups separately, and bring an improvement of $0.9$ mIoU in Tab.~\ref{tab:lidarseg_tricks} ID 10.

Existing segmentation methods perform per-point classification, without considering the consistency of a single object. For instance, some points labeled as foreground objects would be predicted as background. Object-level refinement is conducted to further improve object-level integrity based on the aforementioned hierarchical classification. By masking points in the same semantic group based on prediction and performing Euclidean clustering, points could be grouped into instances. Then the prediction of each instance is determined by majority voting. Besides, for each object, the justification of object-level classification  is performed by a lightweight classification network to determine the ultimate predicted class of the object.

As object-level prediction is obtained, the time consistency of the prediction is further refined by tracking. Tracking is performed to find the corresponding object from all previous frames. The predicted class of an object in the current frame is refined by considering all previous predictions.

\section{Challenges and Future Trends} \label{challenge_apdx}

% \textcolor{blue}{@WANG, Wenhai / LI, Hongyang}

Despite the popularity of BEV representation for perception algorithms in autonomous driving, there are still many grand challenges facing the community to resolve. In this section, we list some future research directions.

% large models with moe for multi-modality input and multi auto driving perception tasks
% multi-camera / multi-modality backbone designed and pretrained

\subsection{Depth Estimation}
% lidar info used in train; camera-only during inference

As discussed in main paper, the core issue for vision-based BEV perception lies in an accurate depth estimation, as the task is performed in 3D context. Current approaches to resolve depth prediction is (a) pseudo-LiDAR generation; (b) lifting features from 2D to 3D correspondence; (c) LiDAR-camera distillation; and (d) stereo disparity or temporal motion. 
Either one or a combination of these directions are promising. To guarantee better performance, the large-scale amount of supervision data is of vital importance as well~\cite{park2021pseudo}.

Another interesting and important direction is how to utilize LiDAR information during training (e.g. as depth supervision), whilst only vision inputs are fed during inference. This is desperately favorable for OEMs as often we have convenient amount of training data from multiple sources and yet for the deployment consideration, only camera inputs are available on shipping products.

\subsection{Fusion Mechanism}

Most fusion approaches up to date can be classified as one of the early-fusion, middle-fusion or late-fusion groups, depending on where the fusion module lies in the pipeline. The most straightforward design of sensor fusion algorithms is to concatenate two set of features from camera and LiDAR respectively. However, as previous sections articulate, how to ``align'' features from different sources is of vital importance. This means: (a) the feature representation from camera is appropriately depicted in 3D geometry space rather than 2D context; (b) the point clouds in 3D space have accurate correspondence with those counterparts in 2D domain, implying that the soft and/or hard synchronization between LiDAR and camera is exquisitely guaranteed. 

Built on top of the prerequisites aforementioned, how to devise an elegant fusion scheme needs much more focus from the community. Future endeavors on this part could be, (a) utilizing self and/or cross attention to integrate feature representations from various modalities in Transformer spirit~\cite{vaswani2017_transformer};
(b) knowledge from the general multi-modality literature could be favorable as well, e.g., the philosophy of text-image pairs in CLIP formulation~\cite{radford2021_clip} could inspire the information integration of different sensors in the autonomous driving domain.

\subsection{Parameter-free Design to Improve Generalization}

One of the biggest challenges in BEV perception is the domain adaptation. How well the trained model in one dataset would behave and generalize in another dataset. One cannot afford the heavy cost (training, data, annotation, etc.) of the algorithm in each and every dataset. Since BEV perception is essentially a 3D reconstruction to the physical world, we argue that a good detector must bundle tight connection to camera parameters, esp. the extrinsic matrix. Different benchmarks have different camera/sensor settings, corresponding to physical position, overlap area, FOV (field-of-view), distortion parameters, etc. These factors would all contribute to the (extreme) difficulty of transferring good performance from one scenario to another domain. 

To this end, it urges us to decouple the network from camera parameters, aka, making the feature learning independent of the extrinsic and/or intrinsic matrix. There are some interesting work in this direction both from academia (extrinsic free,~\cite{Zhou2021_parameterfree}) and industry (rectify module,~\cite{tesla_ai_day}).
This is non-trival nonetheless and it would be better to investigate more from the community as future work. The parameter-free design is robust to resolve detection inaccuracy due to road bumpiness and camera unsteadiness in realistic applications.

\subsection{Foundation Models to Facilitate BEV Perception}

There is blossom in recent years from the general vision community where large or foundation models~\cite{bommasani2021_foundationmodel,radford2021_clip,vaswani2017_transformer,dosovitskiy2020_vit,brown2020_gpt} achieve impressive performance and override state-of-the-arts in many domains and tasks. At least two aspects are worth investigating for BEV perception. 

One is to apply the affluent knowledge residing in the large pre-trained models and to provide better initial checkpoints to finetune. However, the direct adaptation of some 2D foundation model might not work well in 3D BEV sense, as previous section implies. How to design and choose foundation models to better adapt autonomous driving tasks is a long standing research issue we can embrace.

Another unfinished endeavor is how to develop the idea of multi-task learning as in the foundation models (generalist) for BEV perception. There are interesting work in the general vision literature where OFA~\cite{wang2022ofa}, Uni-perceiver-MoE~\cite{zhu2022_uniperceiver-moe}, GATO~\cite{scott2022_gato}, etc. would perform multiple complicated tasks and achieve satisfying results. Can we apply the similar philosophy into BEV perception and unify multiple tasks in one single framework? This is meaningful as the perception and cognition domains in autonomous driving need collaborating to handle complicated scenarios towards the ultimate L5 goal.

\end{document}